\documentclass[11pt, oneside]{article}

\usepackage[english]{babel}
\usepackage[T1]{fontenc}
\usepackage[utf8]{inputenc}
\usepackage{lmodern}
\usepackage{microtype}
\usepackage{fontawesome}
\usepackage{marvosym}
\usepackage{pifont}
\usepackage{xspace}
\usepackage{lipsum}

\PassOptionsToPackage{dvipsnames,table,xcdraw}{xcolor}
\PassOptionsToPackage{hyphens}{url}

\usepackage{geometry}
\usepackage{authblk}
\usepackage{sectsty}

\usepackage{amssymb}
\usepackage{amsmath}
\usepackage[version=4]{mhchem}
\usepackage{siunitx}

\usepackage{graphicx}
\usepackage{svg}
\usepackage[most]{tcolorbox}
\tcbuselibrary{listings, breakable, skins}
\usepackage{tikz}
\usetikzlibrary{arrows.meta, positioning, shapes.geometric, calc}
\usepackage{annotate-equations}

\usepackage{array}
\usepackage{booktabs}
\usepackage{longtable}
\usepackage{makecell}
\usepackage{multirow}
\usepackage{tabularx}
\usepackage{float}

\usepackage{subcaption}
\usepackage[labelfont=bf]{caption}

\usepackage{enumitem}
\usepackage{listings}
\usepackage{fvextra}
\usepackage{csquotes}

\usepackage[backend=biber,style=nature, doi=false,isbn=false, date=year, autopunct=true]{biblatex}
\usepackage{url}
\usepackage[hidelinks]{hyperref}
\hypersetup{breaklinks=true}
\usepackage{cleveref}

\IfFileExists{./fonts/Heliotrope-Bold.otf}{
    \usepackage{fontspec}
    \setmainfont{Heliotrope}[Extension = .otf, UprightFont = *-Regular, ItalicFont = *-Italic, BoldFont=*-Bold, Path = ./fonts/]
}{
    \usepackage{biolinum}
    
}

\definecolor{tumblack}{rgb}{0,0,0}
\definecolor{linkcolor}{RGB}{0, 0, 0}
\definecolor{PositiveColor}{RGB}{46, 139, 87}     
\definecolor{NegativeColor}{RGB}{178, 34, 34}     
\definecolor{TemperatureColor}{RGB}{255, 140, 0}  
\definecolor{SimilarityColor}{RGB}{70, 130, 180}  
\definecolor{OverallColor}{RGB}{75, 0, 130}       
\definecolor{examplered}{RGB}{180, 60, 60}
\definecolor{lightgray}{gray}{0.93}

\sectionfont{\color{tumblack}\selectfont\sffamily}
\subsectionfont{\color{tumblack}\selectfont\sffamily}
\subsubsectionfont{\color{tumblack}\selectfont\sffamily}
\paragraphfont{\selectfont\sffamily}

\makeatletter
\renewbibmacro*{textcite}{%
  \iffieldequals{namehash}{\cbx@lasthash}
    {\usebibmacro{cite:comp}}
    {\usebibmacro{cite:dump}%
     \ifbool{cbx:parens}
       {\bibclosebracket\global\boolfalse{cbx:parens}}
       {}%
     \iffirstcitekey
       {}
       {\textcitedelim}%
     \usebibmacro{cite:init}%
     \ifnameundef{labelname}
       {\printfield[citetitle]{labeltitle}}
       {\printnames{labelname}}%
     \addspace
     \ifnumequal{\value{citecount}}{1}
       {\usebibmacro{prenote}}
       {}%
     \mkbibsuperscript{\usebibmacro{cite:comp}}%
     \stepcounter{textcitecount}%
     \savefield{namehash}{\cbx@lasthash}}}
\makeatother

\lstdefinelanguage{json}{
  morestring=[b]",
  morestring=[s]{"}{"},
  morecomment=[l]{//},
  morecomment=[s]{/*}{*/},
  morekeywords={true,false,null},
  sensitive=false
}

\lstdefinestyle{json}{
    basicstyle=\ttfamily\small,
    breaklines=true,
    frame=single,
    backgroundcolor=\color{gray!10},
    keywordstyle=\color{blue},
    stringstyle=\color{red!70!black}
}

\lstdefinestyle{plaintext}{
    basicstyle=\ttfamily\small,
    breaklines=true,
    postbreak=\mbox{\textcolor{red}{$\hookrightarrow$}\space},
}

\lstdefinestyle{python_example}{
    language=Python,
    basicstyle=\ttfamily\small,
    keywordstyle=\color{blue},
    stringstyle=\color{red!70!black},
    commentstyle=\color{green!50!black},
    morecomment=[l]{\#},
    breaklines=true,
    showstringspaces=false,
}

\lstdefinestyle{python}{
    language=Python,
    basicstyle=\ttfamily\small,
    keywordstyle=\color{blue}\bfseries,
    commentstyle=\color{green!60!black}\itshape,
    stringstyle=\color{red},
    breaklines=true,
    showstringspaces=false,
    frame=none,
    tabsize=4,
    morekeywords={True,False,None,and,or,not,is,in,def,class,if,elif,else,for,while,try,except,finally,with,as,import,from,return,yield,lambda,global,nonlocal}
}

\lstnewenvironment{pythoncode}[1]
{
    \renewcommand{\lstlistingname}{#1}
    \lstset{
        style=python,
        title=#1,
        frame=single,
        frameround=tttt,
        backgroundcolor=\color{blue!5!white},
        rulecolor=\color{blue!60!black}
    }
}
{}

\newcolumntype{Y}{>{\centering\arraybackslash}X}
\newcolumntype{L}[1]{>{\raggedright\arraybackslash}p{#1}}

\newtcolorbox{briefbox}[1][]{
    title=#1,
    breakable,
    enhanced,
    colback=gray!5!white,
    colframe=gray!60!black,
    fonttitle=\bfseries
}
\newtcolorbox{workflowbox}[1][]{
    title=#1,
    breakable,
    enhanced,
    colback=gray!5!white,
    colframe=gray!80!black,
    fonttitle=\bfseries
}
\newtcolorbox{comprehensivebox}[1][]{
    title=#1,
    breakable,
    enhanced,
    colback=gray!5!white,
    colframe=gray!75!black,
    fonttitle=\bfseries
}

\newtcolorbox[auto counter]{examplebox}[2][]{
    title = {Example~\thetcbcounter: #2},
    breakable,
    enhanced,
    colback=white,
    colframe=examplered,
    colbacktitle=examplered,
    fonttitle=\bfseries,
    boxrule=0.5mm,
    left=2mm,
    right=2mm,
    top=4mm,
    bottom=4mm,
    middle=10mm,
    #1
}

\newtcolorbox{hypobox}[1]{%
    breakable,
    colback=lightgray,
    colframe=black!15,
    boxrule=0.5pt,
    arc=2mm,
    left=3mm, right=3mm, top=2mm, bottom=2mm,
    fonttitle=\bfseries,
    coltitle=black,
    title={#1}
}

\newfloat{example}{htbp}{loe}
\floatname{example}{Example}

\newfloat{extdatafig}{htbp}{loedf}
\floatname{extdatafig}{Extended Data Fig.}
\crefname{extdatafig}{Extended Data Fig.}{Extended Data Figs.}
\Crefname{extdatafig}{Extended Data Figure}{Extended Data Figures}

\newfloat{extdatatab}{htbp}{loedt}
\floatname{extdatatab}{Extended Data Table}
\crefname{extdatatab}{Extended Data Table}{Extended Data Tables}
\Crefname{extdatatab}{Extended Data Table}{Extended Data Tables}

\crefname{tcb@cnt@examplebox}{example}{examples}
\Crefname{tcb@cnt@examplebox}{Example}{Examples}
\crefname{example}{example}{examples}
\Crefname{example}{Example}{Examples}

\newcommand{\toolname}[1]{\texttt{\url{#1}}}
\newcommand{\corrallogo}{%
  \raisebox{-0.219\height}{\includegraphics[height=2.25em]{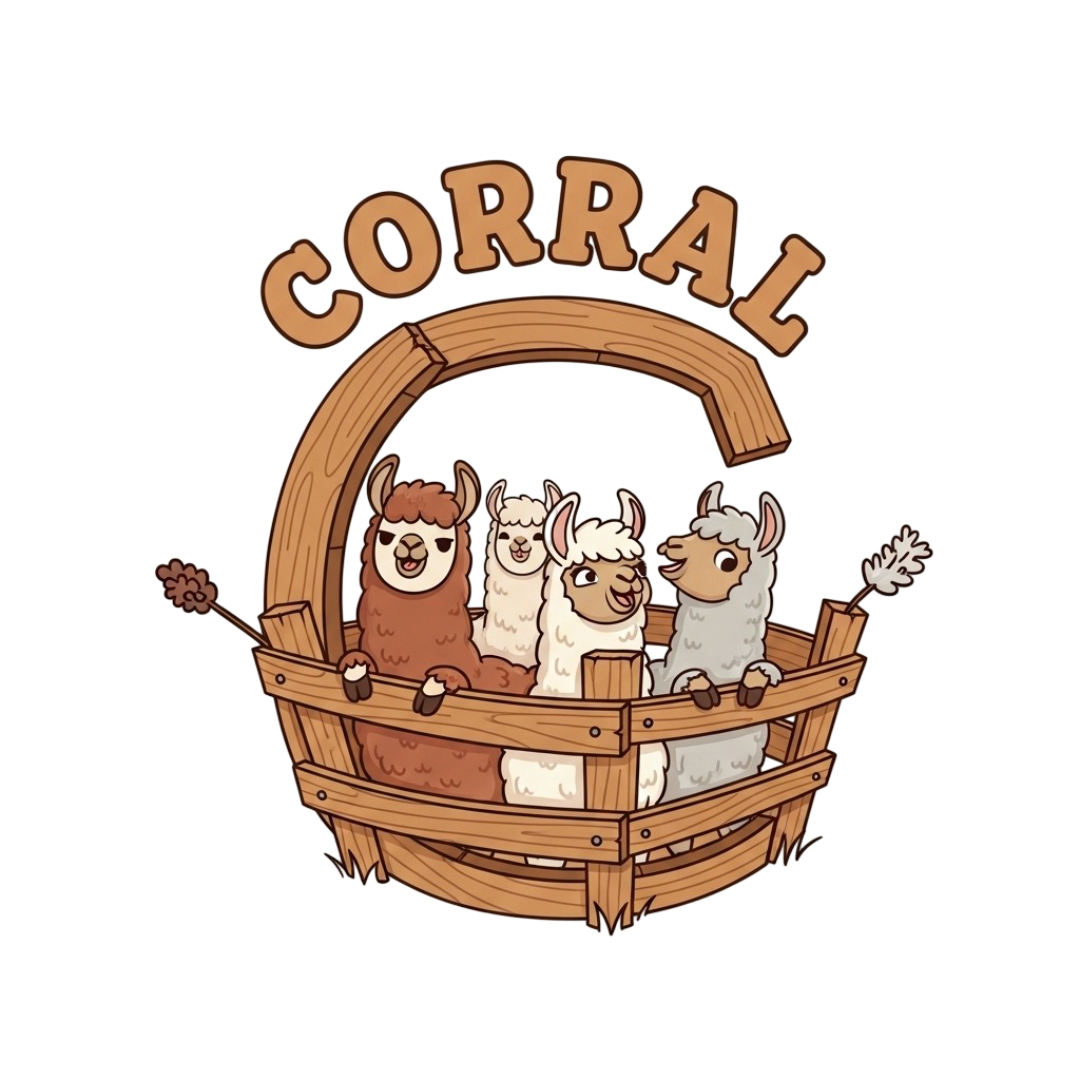}}%
}

\newcommand{\gptfour}{GPT-4\xspace}

\newcommand{\gptoss}{GPT-OSS-120B \xspace}

\newcommand{\claudesonnet}{Claude Sonnet 4.5 \xspace}

\newcommand{\gptfourO}{GPT-4o \xspace}
\newcommand{\gptfourOfull}{GPT-4o (\texttt{gpt-4o-2024-08-06})\xspace}
\newcommand{\claudesonnetfull}{Claude Sonnet 4.5 (\texttt{claude-sonnet-4-5-20250929})\xspace}
\newcommand{\gptossfull}{GPT-OSS (\texttt{gpt-oss-120b})\xspace}

\newcommand{\corral}{\textit{Corral}\xspace}

\newcommand{\afm}{\emph{AFM Experiment Execution}\xspace}
\newcommand{\catalyst}{\emph{Adsorption Surface Construction}\xspace}
\newcommand{\md}{\emph{Molecular Simulation}\xspace}
\newcommand{\ml}{\emph{ML-based Property Prediction}\xspace}
\newcommand{\resistor}{\emph{Circuit Inference}\xspace}
\newcommand{\retro}{\emph{Retrosynthetic Planning}\xspace}
\newcommand{\spectra}{\emph{Spectroscopic Structure Elucidation}\xspace}
\newcommand{\wetlab}{\emph{Inorganic Qualitative Analysis}\xspace}

\newcommand{\passat}{\emph{$\text{Pass}@k$}\xspace}
\newcommand{\passhat}{\emph{$\text{Pass}^\wedge k$}\xspace}

\usepackage{credits}
\usepackage{orcidlink}
\usepackage{hyperref}
\usepackage{multirow}
\usepackage{threeparttable}
\usepackage{arydshln}
\usepackage{xcolor}
\usepackage{tcolorbox}
\usepackage{svg}
\usepackage{algorithm}
\usepackage{algpseudocode}
\usepackage{etoc}
\usepackage{listings}
\lstdefinestyle{corralpy}{
    language=Python,
    basicstyle=\ttfamily\footnotesize,
    keywordstyle=\color{blue!70!black}\bfseries,
    commentstyle=\color{gray!70!black}\itshape,
    stringstyle=\color{teal!70!black},
    showstringspaces=false,
    breaklines=true,
    frame=single,
    rulecolor=\color{gray!40},
    backgroundcolor=\color{gray!5},
    numbers=left,
    numberstyle=\tiny\color{gray},
    xleftmargin=1.5em,
    aboveskip=0.5em,
    belowskip=0.5em,
}
\title{\textsf{AI scientists produce results without reasoning scientifically}}

\author[1, $\star$]{Martiño Ríos-García~\orcidlink{0000-0003-1507-4048}}
\author[1, $\star$]{Nawaf Alampara~\orcidlink{0009-0001-7697-7315}}

\author[2]{Chandan Gupta~\orcidlink{0000-0002-2371-3586}}
\author[3]{Indrajeet Mandal~\orcidlink{0000-0002-8808-4602}}
\author[2]{Sajid Mannan~\orcidlink{0000-0002-7887-2250}}
\author[1]{Ali~Asghar~Aghajani~\orcidlink{0000-0002-1318-9311}}

\author[2,3,4,  \Letter]{N. M.~Anoop~Krishnan~\orcidlink{0000-0003-1500-4947}}
\author[1,5,6,  \Letter]{Kevin~Maik~Jablonka~\orcidlink{0000-0003-4894-4660}}

\affil[1]{Laboratory of Organic and Macromolecular Chemistry (IOMC), Friedrich Schiller University Jena, Humboldstr. 10, 07743 Jena, Germany}
\affil[2]{Department of Civil Engineering, Indian Institute of Technology Delhi, Hauz Khas, New Delhi 110016, India}

\affil[3]{School of Interdisciplinary Research, Indian Institute of Technology Delhi, Hauz Khas, New Delhi 110016, India}
\affil[4]{Yardi School of Artificial Intelligence, Indian Institute of Technology Delhi, Hauz Khas, New Delhi 110016, India}

\affil[5]{Center for Energy and Environmental Chemistry Jena, Friedrich Schiller University Jena, Philosophenweg 7, 07743 Jena, Germany}
\affil[6]{Helmholtz Institute for Polymers in Energy Applications Jena (HIPOLE Jena), Lessingstraße 12--14, 07743 Jena, Germany}
\affil[$\star$]{These authors had an equivalent impact on this work. Name order was decided at random, and the first position is interchangeable, reflecting their shared contribution.}
\affil[\Letter]{Correspondence to \texttt{krishnan@iitd.ac.in} and \texttt{mail@kjablonka.com}}

\credit{Martiño~Ríos-García}{1,1,1,0,1,1,0,0,1,0,1,1,1,1}
\credit{Nawaf~Alampara}{1,1,1,0,1,1,0,0,1,0,1,1,1,1}
\credit{Chandan~Gupta}{1,1,1,0,1,1,0,0,1,0,1,1,1,1}
\credit{Indrajeet~Mandal}{1,1,1,0,1,1,0,0,1,0,1,0,0,1}
\credit{Sajid~Mannan}{1,1,1,0,1,1,0,0,1,0,1,0,0,1}
\credit{Ali~Asghar~Aghajani}{1,1,1,0,1,1,0,0,1,0,1,0,0,1}
\credit{N.~M.~Anoop~Krishnan}{1,0,1,1,0,1,1,1,0,1,1,0,0,1}
\credit{Kevin~Maik~Jablonka}{1,0,1,1,0,1,1,1,0,1,1,0,1,1}

\begin{document}
\etocdepthtag.toc{main}
\maketitle
\vspace*{-1em}
\begin{abstract}
\noindent Large language model (LLM)-based systems are increasingly deployed to conduct scientific research autonomously, yet whether their reasoning adheres to the epistemic norms that make scientific inquiry self-correcting is poorly understood. Here, we evaluate LLM-based scientific agents across eight domains, spanning workflow execution to hypothesis-driven inquiry, through more than 25,000 agent runs and two complementary lenses: (i) a systematic performance analysis that decomposes the contributions of the base model and the agent scaffold, and (ii) a behavioral analysis of the epistemological structure of agent reasoning. We observe that the base model is the primary determinant of both performance and behavior, accounting for 41.4\% of explained variance versus 1.5\% for the scaffold. Across all configurations, evidence is ignored in 68\% of traces, refutation-driven belief revision occurs in 26\%, and convergent multi-test evidence is rare. The same reasoning pattern appears whether the agent executes a computational workflow or conducts hypothesis-driven inquiry. They persist even when agents receive near-complete successful reasoning trajectories as context, and the resulting unreliability compounds across repeated trials in epistemically demanding domains. Thus, current LLM-based agents execute scientific workflows but do not exhibit the epistemic patterns that characterize scientific reasoning. Outcome-based evaluation cannot detect these failures, and scaffold engineering alone cannot repair them. Until reasoning itself becomes a training target, the scientific knowledge produced by such agents cannot be justified by the process that generated it.
\end{abstract}

\pagebreak

\begin{refsection}

\section{Introduction}

DENDRAL \autocite{lindsay1993dendral}, one of the first artificial intelligence (AI) systems to produce a scientific result, showed its reasoning at every step. Developed by Feigenbaum, Lederberg, and Djerassi in 1965, it inferred molecular structures from mass spectra by generating candidate structures, matching them against spectral evidence, and eliminating those that failed. DENDRAL established a template for AI in science: epistemic machinery encoded in explicit rules, every inference step inspectable. Expert systems built on this template remain in use in laboratories today. Large language models (LLMs) have renewed this ambition. A growing number of LLM-based agents are now designed to conduct scientific research autonomously, from hypothesis generation through experimentation to manuscript writing (see \Cref{app:rise_of_ai_scientists} for the rising interest in those systems) \autocite{alampara2026general,lu2026towards,Boiko2023autonomous,MBran2024augmenting,ghafarollahi2025sciagents,novikov2025alphaevolve0,miret2025enabling,mandal2025evaluating}.

Yet the nature of these systems differs fundamentally from their predecessors. Where DENDRAL showed its reasoning at every step, LLM-based agents do not. Their behavior derives from statistical regularities learned during self-supervised training. The epistemic process by which they arrive at scientific conclusions is largely inaccessible to scrutiny.

This opacity matters for science itself. The tools scientists use shape the science they produce \autocite{culkin1967schoolman}. Evans and colleagues recently confirmed this for AI: the adoption of machine learning (ML) is narrowing the hypotheses scientists pursue \autocite{hao2026artificial}. LLM-based agents go further, directing the inquiry itself. How these systems reason, and whether their reasoning adheres to the norms that make scientific inquiry self-correcting, spans computer science, philosophy of science, and metascience \autocite{ioannidis2015meta,pittphilsci28744}. It bears on whether AI systems can produce scientific understanding beyond correct predictions \autocite{krenn2022understanding}.

Current evaluations do not address this. Benchmarks for scientific agents measure task completion: whether a system answers domain-specific questions \autocite{mirza2025framework,alampara2025macbench,laurent2024lab}, executes a data-analysis or discovery workflow \autocite{chen2024scienceagentbench,shen2026sciagentgym,cui2025curie,mandal2025evaluating}, reimplements a published result, or produces a manuscript that passes peer review \autocite{alampara2025lessons,chen2025mlr,bragg2025astabench,rabanser2026towards,mitchener2025bixbench0}. These metrics are task-specific. They reveal whether an agent succeeded but not how or why, and they do not predict behavior on new problems \autocite{Zhou2026}.
The epistemic process underlying a result determines whether the knowledge produced is justified in the philosophical sense of justified true belief \autocite{sep-knowledge-analysis}, whether failures are predictable, and whether the system can be trusted beyond the conditions under which it was tested.

Here, we study LLM-based scientific agents, also referred to as AI scientists, through two complementary lenses. 
The first is a systematic performance evaluation across controlled environments with standardized tools, graded task scopes, and fine-grained subtask diagnostics, designed to separate the contributions of the base model from the agent scaffold \autocite{meirelesinfluence}. The second is a behavioral analysis of the epistemological structure of agent reasoning: what cognitive operations an agent performs, how they relate to one another, and whether they constitute disciplined inquiry. 
We evaluate three frontier models paired with two scaffold architectures across eight scientific domains, from molecular simulation and surface construction to spectroscopic structure elucidation and qualitative chemical analysis (see \Cref{fig:overview}). We find that the base model is the primary determinant of both performance and behavior, with the scaffold contributing little. Agents routinely ignore evidence they have gathered, commit to hypotheses without testing them, and fail to revise beliefs when confronted with contradictory data. These patterns do not adapt to the demands of the problem. The same undisciplined reasoning appears whether the task requires executing a computational workflow or conducting a hypothesis-driven inquiry under uncertainty. These findings suggest that evaluating LLM-based agents requires direct assessment of their reasoning process, and that progress will require changes at the base-model level rather than further scaffold engineering.

\section{Results}
\begin{figure}
    \centering
    \includegraphics[width=1\linewidth]{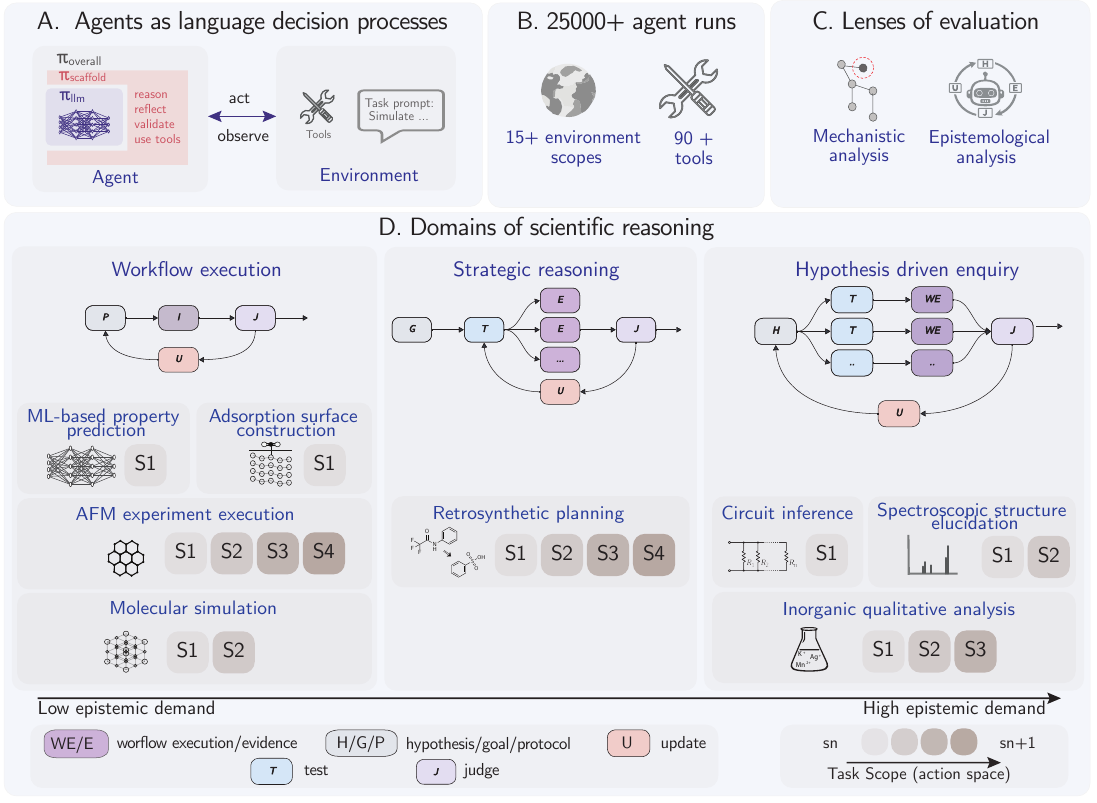}
    \caption{\textbf{Benchmarking scientific reasoning across epistemic demand and problem scope.}
(A) Agents as language decision processes. A base LLM policy is augmented with an agent scaffold (e.g., reasoning, reflection, tool use) to form an agent that interacts with scientific environments via iterative act-observe loops.
(B) Large-scale evaluation grounds our findings. The benchmark comprises 25{,}000+ agent runs across 15+ environment scopes and 90+ tools, enabling statistically grounded evaluation of model and scaffold contributions.
(C) Mechanistic and epistemological analyses reveal failure modes. We analyze agent behavior through mechanistic analysis of trajectories and epistemological analysis of reasoning processes.
(D) Environments are organized along an axis of increasing epistemic demand (left to right). Workflow execution tasks (e.g., \md, \catalyst, \ml, \afm) emphasize procedural correctness; strategic reasoning tasks (e.g., \retro) require multi-step planning under constraints; hypothesis-driven inquiry tasks (e.g., \spectra, \wetlab, \resistor) require generating, testing, and revising competing hypotheses. Within each domain, task scope increases through graded configurations (sn $\rightarrow$ sn+1) that expand the search space. Axis labels WE and E mark workflow-execution and evidence anchors; H, G, and P mark hypothesis, goal, and protocol anchors; T, J, and U denote test, judge, and update operations.}
    \label{fig:overview}
\end{figure}

We model LLM-based agents as language decision processes \autocite{narayanan2024aviary0,carta2023grounding}, in which a composite policy factors into a base-model component governing token-level choices and a scaffold component governing prompting strategy, tool routing, and orchestration logic (\Cref{fig:overview}A, Methods \Cref{sec:formalism}). 
In practice, a scaffold wraps the base model in a loop: it formats prompts, routes tool calls, appends observations to the conversation history, and repeats until a termination criterion is reached. Common implementations include ReAct \autocite{yao2022react} and structured tool-calling interfaces. 
This decomposition allows us to attribute observed behavior to the base model or to the scaffold. We evaluate agents through two lenses: a systematic performance analysis supported by manually curated diagnostic assessments, and a behavioral analysis of the epistemological structure of reasoning traces.

We constructed environments in eight scientific domains that vary along two dimensions: epistemic demand and task scope (\Cref{fig:overview}D, \Cref{sec:domains_and_environments}). 
Epistemic demand captures the kind of reasoning a problem requires, ranging from workflow to hypothesis-driven tasks. Workflow-construction tasks include molecular simulation, surface construction, machine learning, and atomic force microscopy, which are procedurally complex but epistemically straightforward---the solution path is well-defined and the agent must execute it correctly. Hypothesis-driven tasks in spectroscopic structure elucidation, qualitative chemical analysis, and circuit inference are different. The agent must first understand the problem space itself, then form hypotheses about hidden structure, design discriminating experiments, and revise beliefs as evidence accumulates. Retrosynthetic planning occupies an intermediate position, requiring strategic search through large combinatorial spaces under cost and availability constraints. Problem scope varies within each domain through graded configurations that expand problem scope by modulating the search space: from a few to many interacting ions, from simple to complex resistor networks, from spectral confirmation to full elucidation, from measuring friction to (re)discovering power-law behavior. 
Each environment further decomposes into subtasks isolating specific cognitive operations---retrieval, execution, reasoning, and validation (more details in \Cref{sec:task_subtask}).

Within each group, specific environments were designed to isolate aspects of scientific reasoning that are difficult to probe with existing benchmarks. The qualitative analysis environment simulates chemical equilibria from thermodynamic data, requiring iterative experimental design under limited sample volume. The circuit inference environment isolates logical reasoning from domain knowledge. The spectroscopic structure elucidation environment poses the same class of problem that motivated DENDRAL---inferring molecular identity from spectral evidence---with access to multiple analytical techniques and a cost-information tradeoff. 
All environments share a common implementation framework with standardized tool interfaces and can be explored interactively at \url{https://lamalab-org.github.io/corral/#environments}.

\subsection*{The base model determines performance}

\begin{figure}
    \centering
    \includegraphics[width=\linewidth]{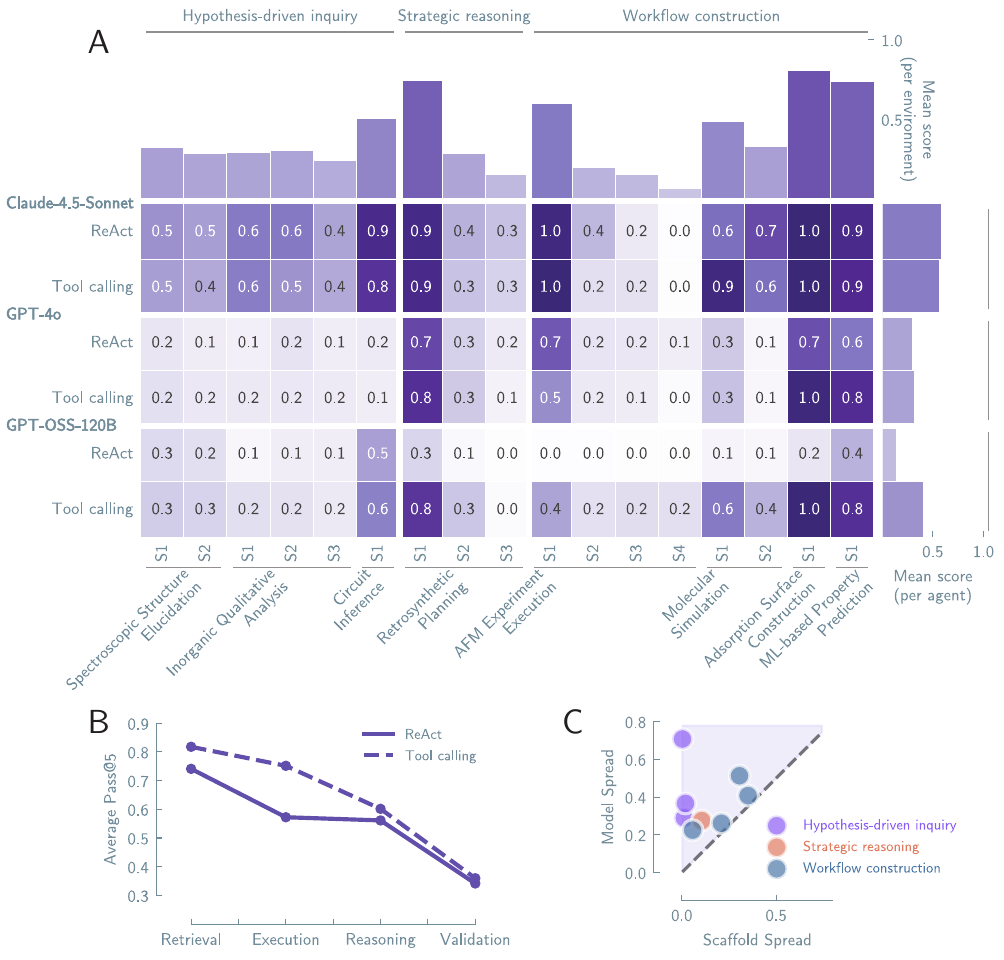}
    \caption{\textbf{Performance is primarily driven by model choice and degrades with epistemic demand.}
    (A) Heatmap of average scores for all model-scaffold configurations across environment-scope combinations, grouped by epistemic demand category.
    (B) Average $\text{Pass}@5$ across subtask categories ordered by increasing cognitive complexity. Both scaffolds exhibit the same decline from retrieval to validation.
    (C) Model spread (performance range across models) versus scaffold spread (performance range across scaffolds) for each environment. Nearly all environments fall above the diagonal, indicating that model choice, not scaffold design, is the primary source of performance variation.}
    \label{fig:performance}
\end{figure}

We evaluated three frontier (two commercial closed-source and one open-source) models---\gptfourOfull, \claudesonnetfull, and \gptossfull---each paired with two scaffold architectures (ReAct and structured tool-calling), across all eight domains and their task scopes (\Cref{fig:performance}; per-configuration scores in \Cref{tab:aggregate-scores,tab:aggregate-scores-subtask}).

Agents solve workflow-construction tasks at guided scopes reliably: in molecular simulation and surface construction, leading configurations approach the performance ceiling (see \Cref{fig:performance}). 
Performance declines as the problem scope broadens within a domain and declines more steeply as epistemic demand increases. 
In \wetlab and \spectra at the broadest scope, even the strongest configuration falls below 60\%. Across nearly all environments, the performance spread attributable to model choice exceeds the spread attributable to scaffold choice (\Cref{fig:performance}C). Both scaffolds yield the same decline in $\text{Pass}@5$ from retrieval through execution and reasoning to validation (\Cref{fig:performance}B). Manual annotation of reasoning traces by two expert annotators, using a custom tool and a taxonomy developed for this study (\Cref{app:manual_annotation_description}), independently confirms this pattern (\Cref{fig:annotation-main-fig}).

\begin{figure}[htb]
    \centering
    \includegraphics[width=\linewidth]{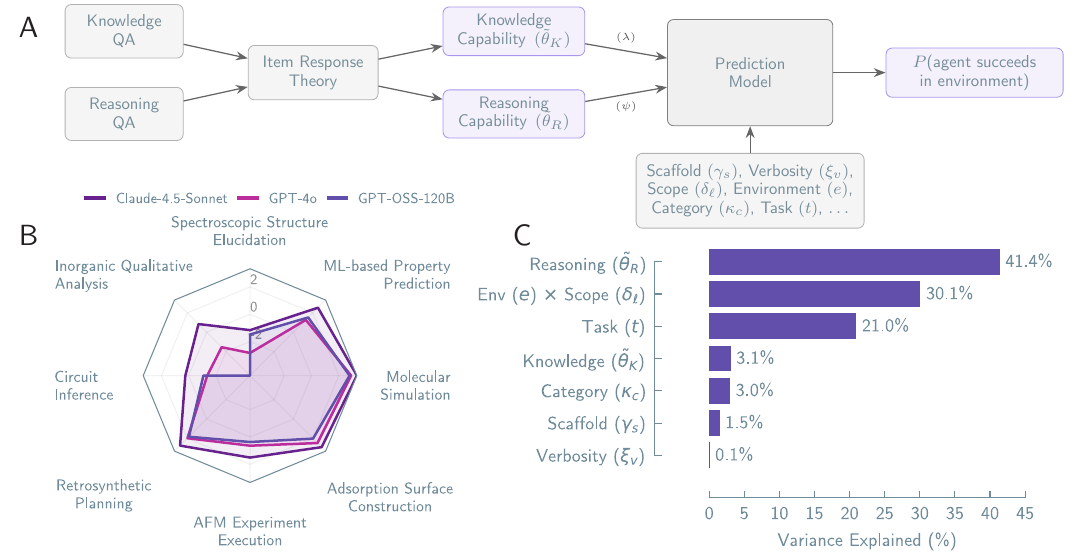}
    \caption{\textbf{Reasoning ability is the dominant predictor of task success.} (A) A two-stage latent-factor framework separates knowledge and reasoning contributions to task success. Stage~1: item response theory estimates latent knowledge and reasoning abilities from domain-specific diagnostic assessments. Stage~2: a prediction model combines these ability estimates with environment, scope, scaffold, and verbosity to predict task success probability. (B) Models diverge most in hypothesis-driven domains. Reasoning capability profiles show the largest separation in \wetlab and \spectra. (C) Reasoning ability dominates; scaffold choice is negligible. Variance decomposition of the prediction model: reasoning ability (41.4\%) and environment--scope interaction (30.1\%) together account for over 70\% of explained variance. Scaffold (1.5\%) and verbosity (0.1\%) contribute minimally.}
    \label{fig:irt-main-fig}
\end{figure}

To move beyond task-level performance and identify the underlying capabilities that drive it, we constructed a diagnostic battery of question-answer pairs for each domain, separately targeting the knowledge and reasoning operations associated with each environment (\Cref{sec:qa}). 
We applied item response theory (IRT) to these diagnostic items \autocite{hori2022item,zhou2026lost}. IRT jointly estimates item difficulty and respondent ability from patterns of correct and incorrect answers, yielding latent ability parameters for each model-environment combination (\Cref{fig:irt-main-fig}A, \Cref{sec:app_irt}). The resulting capability profiles differ across models (\Cref{fig:irt-main-fig}B). The separation is most pronounced for reasoning ability in hypothesis-driven domains, where the gap between the strongest and weakest model exceeds 5.3 standard units in \wetlab (see \Cref{fig:capability_heatmaps} for all environments). In \retro and \spectra, two of the three models effectively lack the domain knowledge required. 
The \ml environment shows the smallest knowledge gap (0.62 standard units, see \Cref{fig:capability_heatmaps} for knowledge capabilities), consistent with the broad coverage of machine-learning topics in LLM training data.

We combined these IRT-derived abilities with environment, scope, scaffold, and verbosity-level (see Methods for details) as parameters in a latent factor model predicting task success (\Cref{fig:irt-main-fig}A). 
We evaluated eight candidate model specifications of increasing structural complexity (M1--M8; \Cref{sec:models_irt}) and selected the best-fitting model (M7) by PSIS-LOO cross-validation (\Cref{app:binomial-subsec}, \Cref{fig:elpd_vs_complexity}). The variance decomposition is clear (\Cref{fig:irt-main-fig}C). Reasoning ability accounts for 41.4\% of the explained variance, environment scope for 30.1\%, scaffold for 1.5\% and verbosity for 0.1\%. 
All three independent analyses (raw performance comparisons, manual trace annotation, and latent factor modeling) converge on the same conclusion: the base model determines the performance of the agent, and the scaffold modulates the interaction only marginally.

\subsection*{Epistemological structure of agent reasoning}

\begin{figure}
    \centering
    \includegraphics[width=\textwidth]{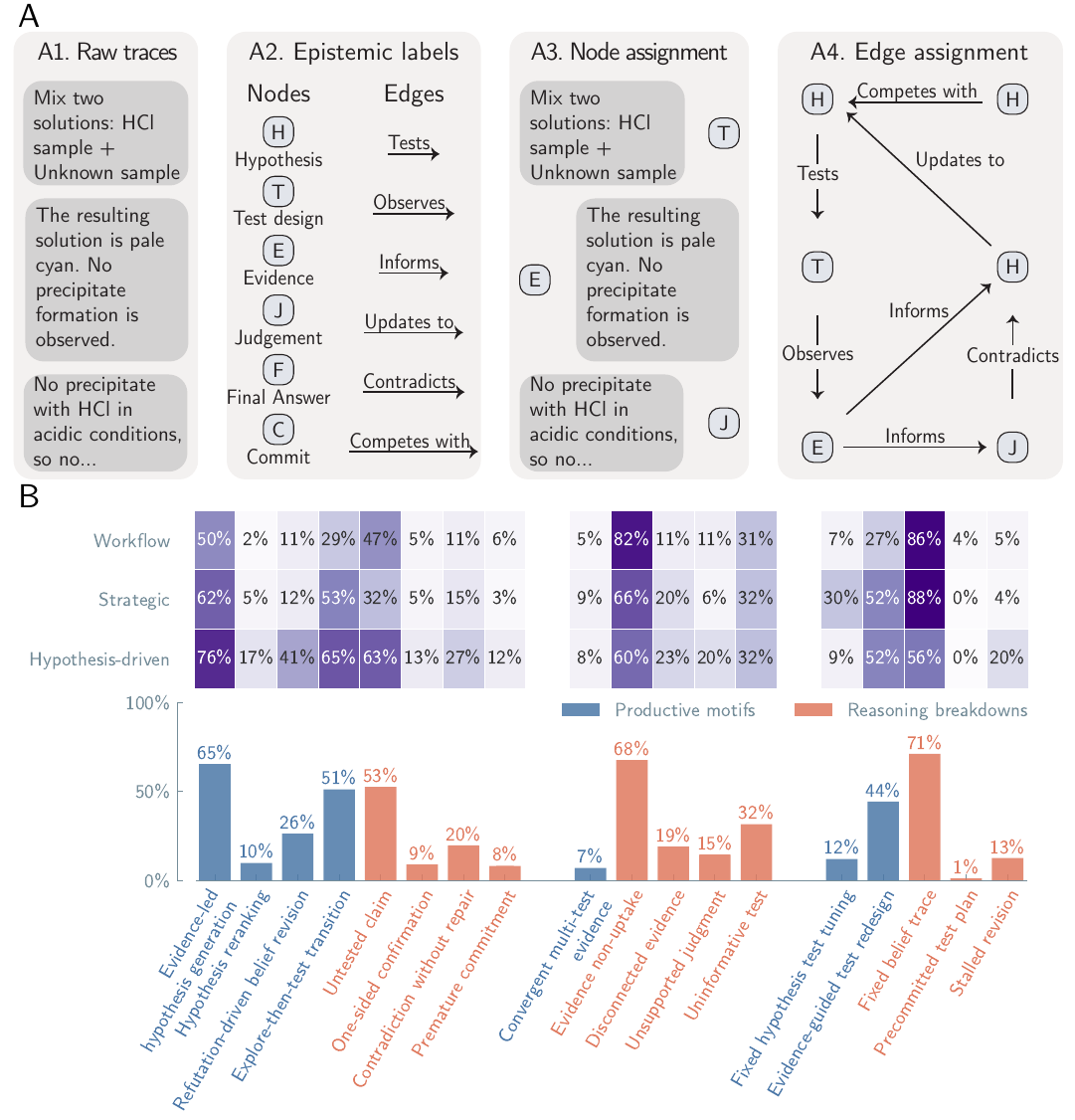}
    \caption{\textbf{Reasoning breakdowns dominate across all domain groups.} (A) Epistemological annotation process. Agent traces are annotated in two stages using the model \claudesonnet: first, each step is labeled with an epistemic operation (hypothesis, test, evidence, judgment, update, commitment); second, directed edges capture dependencies between operations. The annotation process is described in detail in \Cref{app:annotation-pipeline}. (B) Frequency of each pattern across domain groups. Bar plots show that reasoning-breakdown rates exceed productive-motif rates in every domain group; detailed scores are in \Cref{tab:epist-full-scores}.}
    \label{fig:epistemology}
\end{figure}

To examine how agents reason, we analyzed the full agent traces.
We classified each step in an agent's trajectory according to the epistemic operation it performs: forming a hypothesis~(H), gathering evidence~(E), testing a prediction~(T), justifying a conclusion~(J), updating a belief~(U), or committing to a hypothesis~(C). We then represented the dependencies among these operations as a directed graph in which each node is an epistemic operation and each edge a dependency or reference between operations (\Cref{fig:epistemology}A). We observe that different reasoning modes produce different structural signatures. A Popperian cycle of conjecture and refutation generates loops between H, T, and U nodes. Abductive inference produces converging structures where multiple E nodes feed into H. A linear chain of tool calls with no back-references produces a path graph with minimal branching (see Methods \Cref{sec:epistemological-graphs} for more details and \Cref{app:annotation-pipeline} for the full annotation protocol).

\paragraph{Reasoning breakdowns predominate.}
We compiled a catalog of productive reasoning patterns and anti-patterns, defined as structural templates over the annotated graphs (see \Cref{tab:prod-motifs-def} and \Cref{tab:reas-break-def} for definitions and illustration of graphs), and measured their prevalence across traces (\Cref{fig:epistemology}B). 
Untested claims, hypotheses stated without experiments designed to test them, appear in 53\% of traces overall and 63\% in hypothesis-driven domains. Evidence non-uptake, gathering a result and proceeding without incorporating it, occurs in 68\% of traces. In 71\% of traces, beliefs are never updated. Only 26\% of traces exhibit refutation-driven belief revision. Convergent multi-test evidence, multiple independent lines of evidence brought to bear on a single hypothesis, occurs rarely (7\%). Evidence-led hypothesis generation occurs in 65\% of traces but coexists with the high rates of evidence non-uptake and untested claims. While abductive reasoning occurs, downstream testing and revision, in most traces, do not. Representative trace excerpts for each breakdown category are shown in \Cref{fig:reasoning-breakdown-examples}; the annotated traces are browsable at \url{https://lamalab-org.github.io/corral/#explainers} (see \Cref{app:illustrative-traces}).

\paragraph{Reasoning does not adapt to epistemic demand.}
Agents do not adapt their reasoning to the task, as a human practitioner would. Ion identification rewards hypothesis-test cycles, retrosynthesis rewards combinatorial tree search, and molecular simulation rewards a simulate-validate-refine loop; yet the topology of agent reasoning is the same across all three. Evidence non-uptake rates are 82\% in workflow domains, 66\% in strategic reasoning, and 60\% in hypothesis-driven domains (\Cref{fig:epistemology}C). Convergent multi-test evidence remains at or below 10\% in every domain group. Increasing problem scope within a domain produces no corresponding shift in the agent's epistemic operations (\Cref{fig:scope_invariance}). Agents apply a single reasoning mode across the full epistemic range. This mode is adequate for workflow execution but not for the disciplined treatment of evidence that hypothesis-driven inquiry requires (\Cref{fig:performance}).

This invariance extends to the model level. Models differ in the raw frequency of individual operations, with stronger models producing more H and E nodes per trace, but the topology of their reasoning graphs is similar. Higher-performing models retrieve more relevant knowledge and execute procedures more accurately. The epistemic structure of their reasoning does not change correspondingly.

\subsection*{Trace interventions and reliability}

\begin{figure}
    \centering
    \includegraphics[width=\linewidth]{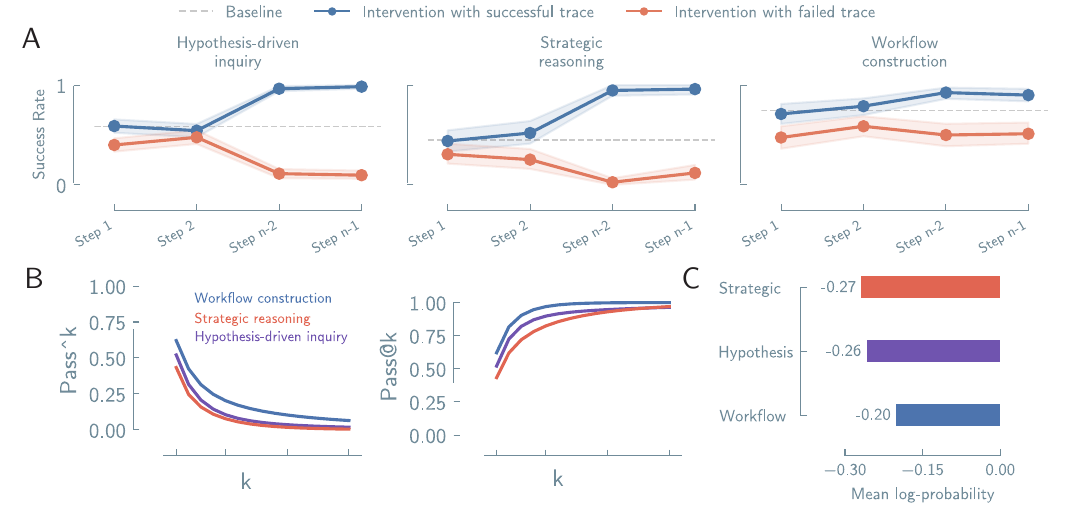}
    \caption{\textbf{Scaffold interventions rescue workflow execution but not hypothesis-driven reasoning; failures in the latter recur across repeated trials.}
    (A) Success rates under trace interventions. Agents receive a partial trajectory from a prior successful run (success trace; blue) or prior failed run (failed trace; red) before attempting a new trial. In workflow-construction environments (mean log-probability $-$0.20), one or two injected successful steps suffice to exceed baseline. In hypothesis-driven and strategic-reasoning environments (\spectra, \wetlab, \retro; mean log-probability $\approx -0.27$), gains emerge only at near-complete trajectories (steps $n{-}2$ or $n{-}1$). Shaded regions show $\pm 1$ standard deviation.
    (B) \passhat{} (all $k$ trials succeed) and \passat{} curves by environment group. In hypothesis-driven domains, \passhat{} falls below 0.1 by $k=4$ under baseline conditions, and below 0.05 by $k=4$--$6$ even with success-trace interventions at early steps.
    (C) Mean token-level log-probability per domain group, ordered by epistemic demand. Domains where partial interventions fail show the most negative log-probabilities, linking intervention unrecoverability to low token-level confidence.}
    \label{fig:combined}
\end{figure}

The reasoning breakdowns documented above raise the question of whether they can be mitigated by providing agents with better context. 
To test this, we designed a trace intervention experiment that provides an extreme form of scaffolding: injecting partial reasoning trajectories from prior runs into the agent's conversation history before it begins a new trial (methodology in \Cref{sec:method_intervention}). These interventions inject actual sequences of tool calls, observations, and reasoning steps: the content a successful agent would have produced on its own. We varied both the source (successful or failed prior runs) and the extent (early steps only, or near-complete trajectories) of the injected traces (\Cref{fig:combined}A).

The effectiveness of these interventions tracks epistemic demand. We compare each intervention to a no-injection baseline (\Cref{sec:method_intervention}) and relate the results to token-level \gptoss log-probabilities (\Cref{sec:logprob-analysis}). In workflow construction domains such as \ml, where models exhibit high token-level confidence (mean log-probability $-$0.2 across the group; per-environment values in \Cref{fig:logprob_breakdown}, \Cref{fig:combined}C), injecting one or two successful steps suffices to exceed baseline performance. Failed-trace interventions have little effect: success rates remain near baseline. In \spectra, \wetlab, and \retro, where token-level confidence is lower (mean log-probability $-$0.27), partial interventions do not improve performance. Gains emerge only when agents receive near-complete successful trajectories (steps $n{-}2$ or $n{-}1$).

This has a direct consequence for reliability. \passhat, the probability that all $k$ independent trials succeed, decays rapidly in hypothesis-driven domains (\Cref{fig:combined}B). Under baseline conditions, \passhat drops below 0.05 by $k = 4$--$6$ for \spectra and \wetlab. Even under success-trace interventions at early steps, the decay is similar. Repeated attempts do not yield consistent success.

This pattern holds across both scaffold architectures. ReAct and tool-calling agents exhibit similar dependence on late-stage successful steps and similar degradation under failed-trace interventions (\Cref{fig:pass_hat_per_agent}). 
Across all four analyses (performance decomposition, latent factor modeling, epistemological trace analysis, and trace interventions), the same conclusion holds: the base model determines what LLM-based agents can do in scientific settings, and current base models do not exhibit the epistemic patterns characteristic of scientific reasoning.
\section{Discussion}

The question of whether machines can conduct scientific inquiry has accompanied AI since DENDRAL built explicit rules for generating and eliminating hypotheses from spectral evidence in 1965 \autocite{lindsay1993dendral}. Our data provide a specific answer for current LLM-based agents. 
In workflow-construction domains, agents approach ceiling performance. In hypothesis-driven domains, they do not adapt their reasoning to the demands of the problem: evidence is ignored in 68\% of traces, refutation-driven belief revision occurs in only 26\%, and convergent multi-test evidence is rare. These patterns persist across scaffold architectures, across domains, and under interventions that provide near-complete successful trajectories. They are properties of the base model.

The epistemic process by which a result is reached determines whether the resulting knowledge is justified \autocite{pittphilsci28744,krenn2022understanding, sep-knowledge-analysis}. 
An agent that reaches a correct answer while ignoring contradictory evidence and leaving hypotheses untested produces a result whose reliability on new problems is unknown. Much current engineering effort focuses on scaffolding (prompting strategies, orchestration logic, tool interfaces), but the reasoning patterns we document persist across all scaffold conditions we tested, including extreme interventions. Addressing them will likely require changes to how base models are trained: a slower process than scaffold improvement, and one currently not guided by epistemic criteria. The framework we introduce can supply such guidance: each environment provides a reproducible task, tools, and a scoring function over agent trajectories, sufficient to define training signals on the reasoning process itself.

Human scientists operate within institutional structures that enforce epistemic norms: peer review, replication, reputational consequences. LLM-based agents are subject to no such constraints. Evidence that reliance on AI assistance can erode independent problem-solving \autocite{liu2026persistence} compounds this gap. Characterizing the epistemic behavior of these systems requires explicit measurement. The evaluation framework we introduce does this across tasks and configurations. New environments, models, scaffolds, and metrics can be contributed, allowing the community to track how the epistemic behavior of these systems changes as models evolve. The methods and findings extend beyond scientific agents to any setting where the reasoning process determines the value of the result.

The tools scientists use shape the science they produce \autocite{culkin1967schoolman,hao2026artificial}. LLM-based agents are now among those tools. Our analysis shows that their epistemic behavior does not exhibit the patterns that characterize scientific reasoning. As long as they are evaluated only by the answers they produce, this difference will remain invisible, and it will shape the knowledge they help produce.
\newpage
\section{Methods}

\subsection{Formalism}
\label{sec:formalism}

We adopt the language decision process (LDP) framework \autocite{narayanan2024aviary0,carta2023grounding}, a formulation of a partially observable Markov decision process in which observations and actions are represented in text. 
An LLM-based agent implements a composite policy $\pi_{\text{overall}}$ that factors into two components:
\begin{equation}
    \pi_{\text{overall}}(a_t \mid s_t) = f\bigl(\pi_{\text{LLM}}(a_t \mid s_t),\; \pi_{\text{scaffold}}(a_t \mid s_t)\bigr),
\end{equation}
where $\pi_{\text{LLM}}$ is the base-model policy governing token-level choices and $\pi_{\text{scaffold}}$ is the scaffold policy governing prompting strategy, tool routing, memory management, and orchestration logic. The state $s_t = (P, h_t)$ comprises the task prompt $P$ and the conversation history $h_t$ at step $t$, which accumulates prior actions, tool outputs, and observations. At each step, the agent selects an action $a_t$ (a tool call, a reasoning step, or a final answer) according to $\pi_{\text{overall}}$. The environment returns an observation, the history grows, and the cycle repeats.

Tools are part of the environment, not the agent. Each environment defines a set of callable functions with standardized docstrings; the agent receives no information beyond what the tool descriptions and returned observations provide. This separation is mirrored directly in our code implementation (\Cref{app:framework-details}): the environment serves tools, task descriptions, and scoring functions through a common interface, while the agent interacts solely through text-based requests. Because all agent configurations face identical tool interfaces and environment dynamics, observed differences in performance and behavior can be attributed to the base model or the scaffold.

\subsection{Task and tool formalism}
\label{sec:task_subtask}

  \paragraph{Tasks and subtasks.}
  A benchmark item is either a single task or a task group $G = \{\tau_1, \ldots, \tau_K\}$ whose dependencies form a directed acyclic graph: each subtask consumes the outputs of its predecessors. The framework resolves this graph and presents subtasks to the agent sequentially, injecting predecessor outputs into the prompt at each step. An \ml task, for example, decomposes into data retrieval, feature engineering, model training, and evaluation, with each subtask consuming the artifacts produced by the previous one.

\paragraph{Task prompt.}
\label{sec:task-prompt}

The message sequence presented to the base model comprises a system prompt, an environment-specific task prompt (goal description, submission format, and any input data from predecessor subtasks), a scaffold-specific user template that adds response-format instructions, and tool descriptions filtered by the verbosity axis defined below (see \Cref{sec:tool-verbosity}). Only the task prompt varies across environments; all other components are held constant. Full construction details are given in \Cref{app:prompt-construction}.

\paragraph{Tool verbosity.}
\label{sec:tool-verbosity}

Tool docstrings are authored with tagged sections (such as \texttt{[BRIEF]}, \texttt{[DETAILED]}, \texttt{[EXAMPLES]}; full tag definitions in \Cref{sec:verobisty_examples}). Three levels expose cumulative subsets to the agent: brief includes only the \texttt{[BRIEF]} tag; workflow extends through \texttt{[WORKFLOW\_INTEGRATION]}; comprehensive adds examples, error conditions, and limitations. Tool descriptions and task prompts are constructed independently, so verbosity varies without altering the task, enabling a controlled ablation over the information available to $\pi_{\text{LLM}}$.

\subsection{Agent implementations} \label{sec:agents-implementations}

All experiments use Corral, the evaluation framework we developed for this study (\Cref{app:framework-details}). Only the model backend and agent scaffold vary across runs. We evaluated three models: GPT-4o (\texttt{gpt-4o-2024-08-06}) via the OpenAI API, Claude Sonnet 4.5 (\texttt{claude-sonnet-4-5-20250929}) via the Anthropic API, and GPT-OSS-120B via the Blablador API. Temperature was set to 0.0 for all runs, and each environment enforces a fixed iteration limit. Tool verbosity was set to brief for main-text results; workflow and comprehensive settings appear in the appendix.

We used two agent scaffolds. The first is ReAct \autocite{yao2022react}: the agent produces reasoning traces and tool calls in a single text stream. The second is structured tool calling: the API parses tool names and arguments from the model output, the agent executes the call, appends the result to the conversation history, and repeats until a termination criterion is met. The two scaffolds differ in how actions are expressed, but the surrounding benchmark logic is identical. Scope and measurement caveats, including compute budget and malformed-response rates, are detailed in \Cref{app:limitations}.

\subsection{Domains and environments} \label{sec:domains_and_environments}
 
Each domain is realized as one or more environments providing tools, a task description, and a scoring function. Observations combine structured data (numerical values, tables, spectra) with natural-language summaries at the selected verbosity level. Scoring functions evaluate both the final answer and, where applicable, intermediate subtask completions. Each environment is backed by a domain-specific engine, instrument, or dataset, ranging from LAMMPS simulations and a live atomic force microscope to a custom thermodynamic equilibrium engine and manually curated spectra (\Cref{tab:env-summary-table}). The inputs each environment receives and the artifacts each produces are shown in \Cref{fig:envs-input-output}. Below, we summarize each domain. Full tool lists, scoring rubrics, and task specifications for the Inorganic Qualitative Analysis environment appear in \Cref{app:wetlab-section}; details for every environment are at \url{https://lamalab-org.github.io/corral/#environments}.
 
\paragraph{Spectroscopic structure elucidation.}
The agent determines the molecular structure of an unknown compound by requesting and interpreting spectroscopic data\autocite{Banfi2008nmr, Jablonka2022making, Binev2007prediction}. Available tools include mass spectrometry, $^{1}$H and $^{13}$C NMR, HSQC, and infrared spectroscopy, alongside reference databases for chemical shift ranges, isotope distributions, and degree-of-unsaturation calculations. The agent can validate proposed structures and simulate spectra for comparison. Each task comprises 10 sequential subtasks that progress from molecular formula determination through functional group identification to complete SMILES assignment. Two scopes vary the molecular complexity: the first isolates individual analytical steps; the second requires integrating ambiguous, multi-modal evidence. HSQC acquisition is marked as expensive, introducing a cost-information tradeoff.
 
\paragraph{Inorganic qualitative analysis.}
In a custom-built simulation environment, the agent identifies unknown cations in solution using the systematic wet-laboratory procedure taught in introductory chemistry.
Tools include adding reagents, performing flame tests, measuring pH, centrifuging, decanting, heating, and observing color and precipitate changes. The environment simulates real chemical equilibria; observations are computed from thermodynamic data (rather than scripted).
Three scopes increase the number of candidate ions from 3--5 common species with distinct chemistry to 15 or more, including ions with overlapping reactivity (e.g.\ \ce{Ag+}, \ce{Hg^{2+}}), progressively requiring the agent to design discriminating experimental sequences rather than follow a decision tree. Sample volume is limited, penalizing redundant experiments.
 
\paragraph{Circuit inference.}
The agent recovers the topology and component values of a hidden resistor network from pairwise resistance measurements between nodes. Tools provide series and parallel resistance calculations, delta-wye and wye-delta transforms, resistance measurement between specified node pairs, and circuit validation. Six tasks progress from simple parallel-series networks (6--7 resistors) through linear chains (10 resistors) to complex topologies (13+ resistors with multiple interaction points). This domain isolates logical and mathematical reasoning from domain-specific scientific knowledge.
 
\paragraph{Retrosynthetic planning.}
The agent designs multi-step synthetic routes to a target molecule under cost, step-count, and commercial-availability constraints. Tools include a template catalog searchable by functional group or bond type, template application and step verification, commercial-availability lookup with CAS number conversion, and functional- and protecting-group detection. Three scopes increase target complexity from simple molecules with one-step solutions to complex natural products requiring up to five steps within a \$9{,}999 budget, where all leaf nodes must be commercially available.
 
\paragraph{AFM experiment execution.}
The agent operates an atomic force microscope, executes the experimental procedure, and analyzes the resulting nanoscale surface characterization data. The four scopes vary in the degree of guidance: the first specifies all scan parameters for identical measurements, while the last provides only the objective and expects the agent to choose scan areas and derive a mathematical relationship. The Image Optimizer tool is removed at higher scopes, introducing a tradeoff between available assistance and independent reasoning about scan modes and parameters. 
 
\paragraph{Molecular simulation.}
The agent designs and executes molecular dynamics simulations using the LAMMPS package \autocite{LAMMPS} to predict materials properties. Tools cover the full workflow: retrieving crystal structures from the Materials Project \autocite{jain2013commentary, Horton2025}, converting to LAMMPS data format, querying force-field metadata, running simulations, and analyzing output logs.  Two scopes vary the degree of procedural guidance: at the first, the simulation protocol is prescribed; at the second, the agent must select simulation parameters, design the heating or cooling schedule, and choose appropriate analysis methods. Tasks span quenching, surface energy computation, and melting simulations, with quantitative targets and specified tolerances (e.g.\ self-diffusivity of molten \ce{SiO2} within 20\%).
 
\paragraph{Adsorption surface construction.}
The agent builds an adsorbate-slab configuration from bulk crystal structures and molecules for heterogeneous catalysis studies. Tools integrate with the Materials Project to retrieve structures and polymorph data, generate surface slabs for specified Miller indices, vacuum, and enumerate adsorption sites. The agent must select an appropriate bulk polymorph, generate a surface termination, identify relevant adsorption configurations, and place the adsorbate molecule on the adsorption site. 
 
\paragraph{ML-based property prediction.}
The agent assembles a complete machine-learning pipeline to predict formation energies of material polymorphs. Tools cover data retrieval from the Materials Project (batch and single-composition queries), dataset filtering and consolidation, tabular feature engineering, XGBoost  \autocite{chen2016xgboost} training with configurable hyperparameters, evaluation, and cross-validation. Tasks span three material classes (oxides, nitrides, sulfides). Models are scored against specified $R^{2}$ thresholds and cross-validation targets.
 
\subsection{Manual trace annotation}

We built a custom web application to support the manual annotation described in \Cref{fig:marker_annotation_app} (see the annotation app at \url{https://lamalab-org.github.io/corral/#annotate}). The tool displays reconstructed agent traces as interactive graphs. Each node represents a single message-like event, such as user instructions, reasoning steps, or tool calls. Annotators process traces in order. They label only behavior-relevant nodes; by design, system messages, tool outputs, and the initial task description are not annotatable. For each eligible node, annotators select one or more markers from a taxonomy we developed for this study. The taxonomy distinguishes positive behaviors (e.g., planning statements, validation attempts), failure modes (e.g., hallucinations, unnecessary tool use), and neutral states (detailed in \Cref{app:manual_annotation_description}). Annotators can also add free-text observations to each node and trace. The interface requires left-to-right progression and checks that all annotatable nodes receive at least one marker before submission.

For downstream analysis, two expert annotators completed the full annotation process on 773 selected traces. Each selected trace received annotations from one domain expert. The two annotators first annotated and discussed several calibration traces and edge cases to align their understanding. This process produced one final annotation record per trace, which informed all later marker-level analyses.

\subsection{Diagnostic question-answer pairs} \label{sec:qa}
For each domain, we designed two types of diagnostic questions. Knowledge questions target the specific facts needed to complete tasks; reasoning questions target the specific inference operations the environment requires. The designer of each environment compiled the initial set; submissions were made via pull requests and accepted only after two independent expert reviews. Knowledge question counts range from approximately twenty to over one hundred per environment; reasoning question counts are more uniform, with twenty to twenty-five per environment. All three models were evaluated on the full item bank at a temperature of 0.0. Full results are detailed in \Cref{tab:qa-scores}; the item format, scoring rubric, and review protocol are described in \Cref{app:capability-items}.

\subsection{IRT and phenomenological model}

We adopt a two-stage modeling strategy to (i)~estimate the latent cognitive capabilities that each model brings to each scientific domain, and (ii)~quantify how those capabilities, together with agent-design choices, predict benchmark success as illustrated in \Cref{fig:irt-main-fig} A.

\paragraph{Extracting model capabilities.} A two-parameter logistic (2PL) IRT model estimates latent abilities from binary QA responses. The diagnostic questions are described in \Cref{sec:qa}.
For model~$j$ answering item~$i$:

  \begin{equation}\label{eq:irt}
    P(Y_{ij}=1) = \sigma\!\bigl(a_i\,(\theta_j - b_i)\bigr),
  \end{equation}

where $\theta_j$ is the latent ability of model~$j$, $a_i > 0$ is item discrimination, and $b_i$ is item difficulty. We fit two separate 2PL models, one to knowledge items and one to reasoning items, yielding $\theta_j^{(K)}$ and $\theta_j^{(R)}$ per model-environment pair. Both are standardized to zero mean and unit variance before entering Stage~2.

\paragraph{Phenomenological latent factor models.}
The IRT-derived abilities serve as covariates in a second modeling stage that predicts agent success on the full benchmark. Raw trial outcomes are first balanced (equal task counts per environment) and aggregated into Binomial counts $(k_{\text{success}}, n_{\text{trials}})$ per unique combination of model, environment, scaffold, scope, tool-description verbosity, aggregation grain (task-level vs.\ subtask-level outcome), and task identifier (the specific task-scope combination). The standardized IRT abilities $\tilde{\theta}_K$ and $\tilde{\theta}_R$ are then merged as continuous predictors.

We specify a family of Bayesian generalised linear models \autocite{edition2013bayesian} (M1--M8; \Cref{sec:models_irt}) of increasing structural complexity, all sharing the Binomial likelihood
  \begin{equation}
    k \sim \mathrm{Binomial}(n,\; p), \qquad
    \operatorname{logit}(p) = \eta,
    \label{eq:lfm_likelihood}
  \end{equation}
and a common set of sum-to-zero categorical effects for scaffold ($\gamma$), verbosity ($\xi$), and scope ($\kappa$). Models differ in how they partition the remaining variance. Candidate specifications are compared using Pareto-smoothed importance sampling leave-one-out cross-validation (PSIS-LOO; \Cref{fig:elpd_vs_complexity}). The best-fitting model is M7, which allows knowledge and reasoning slopes to vary by environment $\times$ scope. All main-text variance decompositions and predicted success probabilities are derived from M7 (see \Cref{sec:models_irt}).

\subsection{Epistemological graphs} \label{sec:epistemological-graphs}

We sampled traces balanced across environments, models, and problem scope levels, prioritizing task diversity. Only ReAct traces were used, as they contain explicit reasoning steps. Tool description verbosity was set to brief; differences between verbosity levels are minimal (see \Cref{fig:fig-5_app}).

Each trace was annotated using Claude 4.5 Sonnet in two stages. In the first stage, the model labels nodes from short, overlapping windows of messages. Node labels correspond to epistemic operations: hypothesis, test, evidence, judgment, update, and commitment. In the second stage, the model adds directed edges between related nodes. Edge types include testing, observing, using, contradicting, competing, and updating. In both stages, the temperature was set to 0.7. Every node and edge includes a supporting quote and the index of the source message, grounding annotations in the trace. After processing the full trace, repeated items are normalized and duplicates removed. \Cref{app:annotation-pipeline} details the annotation process.

Annotations were validated in two ways. Automated checks verify that each supporting quote appears in the cited message and that each edge connects valid nodes. Expert review guided the annotation design: an expert iteratively revised prompts, labels, and markers until annotations matched their own interpretation of the traces. Inter-annotator agreement analysis on a representative sample of 25 traces showed substantial human–human agreement (overall 92.6\%, mean PABAK 0.853) and even higher human–LLM agreement (95.7\%), confirming that the automated pipeline performs within the range of human expert judgment. \Cref{app:annotation-agreement} provides a deeper analysis of the annotation results.

Productive patterns and anti-patterns are defined as explicit structural templates over the annotated graphs (\Cref{tab:prod-motifs-def}, \Cref{tab:reas-break-def}). Productive patterns include Popperian falsification \autocite{popper2005logic}, exploratory-to-confirmatory transitions \autocite{tukey1977exploratory}, and active learning \autocite{lindley1956measure, fedorov2013theory}. Anti-patterns are drawn from established epistemic failures in scientific reasoning and from behaviors observed in the traces, including untested hypotheses, ignored evidence, and unresolved contradictions. Both are grouped into higher-level families for aggregated analysis (\Cref{tab:epist-full-scores}).

\subsection{Trace intervention experiment} \label{sec:method_intervention}
The trace intervention experiment probes whether agent performance is driven by latent model capability or by accumulated conversational context. We first ran 15 trials per task across all environments at temperature 0.7 to populate a registry of successful and failed execution traces. A moderate temperature follows standard practice for balancing sample diversity against coherence \autocite{chen2021evaluating}. These runs are distinct from the main evaluation, which uses temperature 0.0 for deterministic scores. For each new trial, we inject a partial trajectory from this registry into the agent's conversation history before the trial begins. The injected traces are replayed, tool calls and environment interactions included, so that the agent resumes from a consistent intermediate state. We vary two factors: the source of the injected trace (successful or failed prior run) and its extent (early steps 1--2 versus near-complete trajectories at steps $n{-}2$ or $n{-}1$). The intervention supplies the actual reasoning and actions a successful agent would have produced rather than instructions or prompting strategies. Intervention trials also use temperature 0.7. Protocol details and pseudocode are in \Cref{app:interventions}.

\subsection{Token-level log-probability analysis}
  \label{sec:logprob-analysis}

  To quantify the model's token-level confidence across benchmark environments, we analyze the log-probabilities returned by the language model for each
  generated token. For every assistant message produced during a trial, we select and save the log-probability of the top-1 token at each
  decoding step. Formally, for a response consisting of $T$ tokens
  $\{w_1, w_2, \ldots, w_T\}$,
  \begin{equation}
    \ell_t \;=\; \log p_\theta(w_t \mid w_{<t}, \mathbf{x}),
    \qquad t = 1, \ldots, T,
    \label{eq:token-logprob}
  \end{equation}
  where $p_\theta$ denotes the language model, $w_{<t}$ is the preceding token
  sequence, and $\mathbf{x}$ is the input context (which includes system prompt, task prompt, tool
  definitions, and conversation history).

  For each environment~$e$, we pool all non-special token log-probabilities
  across every assistant message from every trial in that environment. Tokens
  whose log-probability is exactly zero and that correspond to
  special control tokens (e.g., \texttt{<|endoftext|>},
  \texttt{<|im\_end|>}) are excluded. The
  per-environment mean log-probability is then,
  \begin{equation}
    \bar{\ell}_e \;=\; \frac{1}{N_e}\sum_{t=1}^{N_e} \ell_t^{(e)},
    \label{eq:mean-logprob}
  \end{equation}
  where $N_e$ is the total number of retained (non-zero, finite) tokens across the message or trial it belongs to, giving a vocabulary-level confidence estimate. More negative values of $\bar{\ell}_e$ signal that the model distributes probability more diffusely, reflecting greater uncertainty. Per-environment values and interpretation are reported in \Cref{app:logprob-results}.

\section*{Acknowledgements}

The research of M.R.G., N.A., and K.M.J.\ was supported by the Carl-Zeiss Foundation, Open Philanthropy, as well as Intel and Merck via the AWASES research center. K.M.J.\ is part of the NFDI consortium FAIRmat funded by the Deutsche Forschungsgemeinschaft (DFG, German Research Foundation) -- project 460197019. K.M.J.'s work was also supported by a Google Research Scholar Award. N.M.A.K, S.M, C.G, and I.M. thank the High Performance Computing Facility at IIT Delhi for computational and storage resources. S.M. acknowledges the partial funding support from the Prime Minister's Research Fellowship. N.M.A.K. acknowledges the funding support received by Google Research Scholar Award, Anusandhan National Research Foundation (ANRF/ARG/2025/007405/ENS), and Alexander von Humboldt fellowship. 
We thank Nitya Nand Gosvami for providing the support with the Atomic Force Microscopy and valuable feedback associated with it. We thank Santiago Miret, Rob Jordan, and Vijay Narasimhan for valuable discussion. We thank Mara Schilling-Wilhelmi for feedback on a draft of this article and K. Sidharth Nambudiripad for support in human annotation.

\section*{Data availability}

The data supporting this study, including environment tasks\autocite{nawaf_alampara_2026_environment_tasks}, evaluation reports \autocite{lab_of_kevin_jablonka_at_uni_jena_2026_reports}, agent traces \autocite{lab_of_kevin_jablonka_at_uni_jena_2026_traces}, question-answer pairs\autocite{lab_of_kevin_jablonka_at_uni_jena_2026_qas, lab_of_kevin_jablonka_at_uni_jena_2026_qa_topic_reports, lab_of_kevin_jablonka_at_uni_jena_2026_qa_reports}, token-level log probabilities \autocite{lab_of_kevin_jablonka_at_uni_jena_2026_logprobs}, and other data related to the study \autocite{martino_rios-garcia_2026_rise, martino_rios-garcia_2026_annotation, nawaf_alampara_2026_reasoning_annotation, nawaf_alampara_2026_irt, nawaf_alampara_2026_intervention_traces, nawaf_alampara_2026_intervention_reports}, are publicly available in a Hugging Face collection at \url{https://huggingface.co/collections/jablonkagroup/corral} under permissive license.

\section*{Code availability}

The package built in this study is publicly available at GitHub (\url{https://github.com/lamalab-org/corral}). A version of the code corresponding to this manuscript is archived on Zenodo \url{https://doi.org/10.5281/zenodo.19659851}.
\section*{Competing interests}
The authors declare no competing interests. 
\section*{Declaration of AI-assisted technologies in the research and writing process}

The authors used Claude Opus (Anthropic) for assistance with language and readability of the manuscript. During code development, the authors made use of several large language model assistants (Claude Code, GitHub Copilot). After using these services, the authors reviewed and edited the content as needed and take full responsibility for the content of the publication.

\section*{Author contributions}
\footnotesize
\insertcredits
\normalsize

\noindent M.R.G., N.A., and K.M.J.\ jointly designed the Corral framework and the analysis pipeline; M.R.G. and N.A.\ led the implementation. Environments were developed as follows: spectroscopic structure elucidation and retrosynthetic planning by M.R.G.; catalyst surface construction, machine learning, and circuit inference by N.A.; atomic force microscopy by I.M. and S.M.; molecular dynamics by S.M. and C.G.; and inorganic qualitative analysis by A.A.\@ The reasoning-trace analysis was led by M.R.G.; the intervention analysis by N.A.; and the performance-based analysis, enabled by the Corral core, was led jointly by C.G., N.A., and M.R.G.\@ 
N.M.A.K.\ and K.M.J.\ acquired funding, administered the project, provided resources, and supervised the work. K.M.J.\ drafted the manuscript with substantial contributions from M.R.G.\ and N.A.; C.G.\ drafted the related-work section. All authors reviewed, edited, and approved the final version of the manuscript.

\normalsize 

\clearpage
\printbibliography
\end{refsection}
\clearpage

\setcounter{figure}{0}
\setcounter{table}{0}

\begin{extdatafig}[H]
    \centering
    \includegraphics[width=1\linewidth]{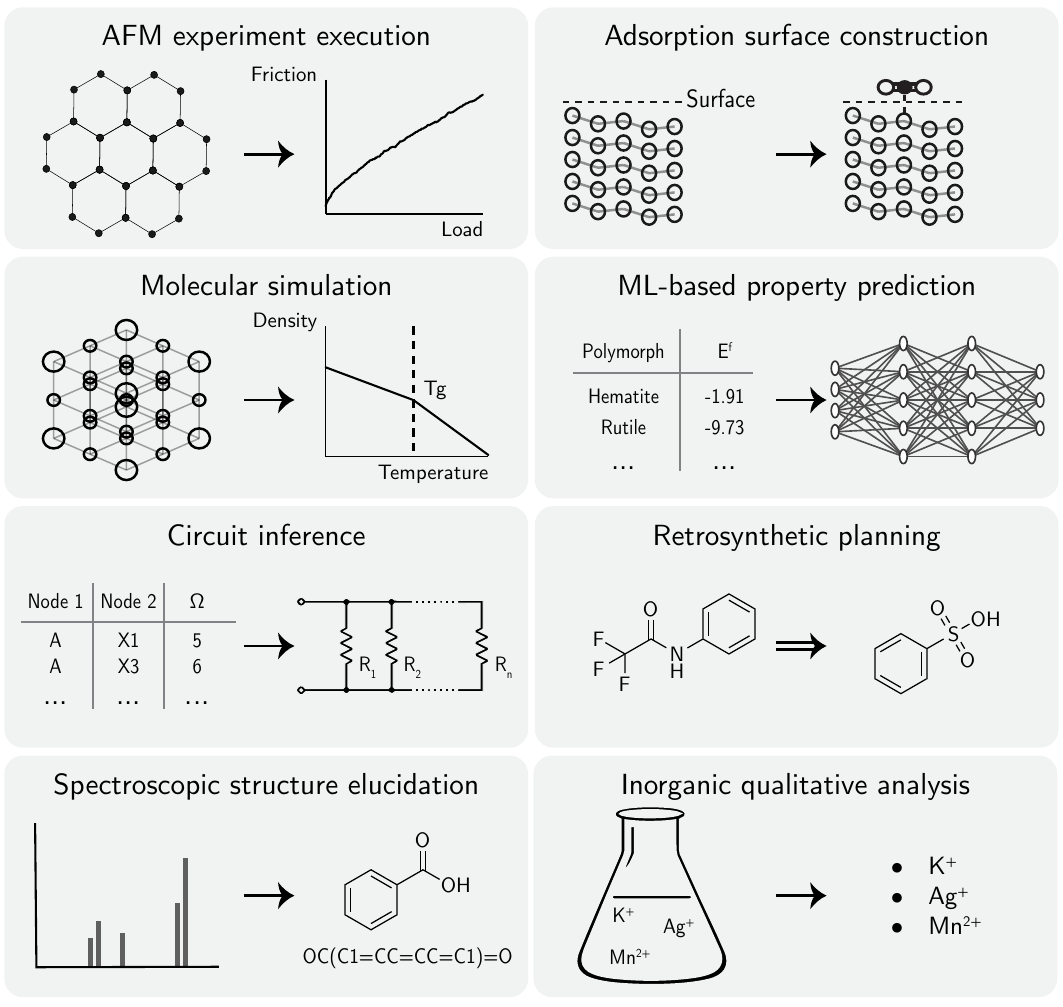}
    \caption{\textbf{Overview of the inputs and outputs of the environments available in \corral.} In \afm, the agent operates an atomic force microscope to capture material images and compute properties such as surface roughness. In \catalyst, the agent builds catalytic surfaces for simulation, producing surface structure files. In \md, the agent runs simulations to obtain material properties. In \ml, the agent trains an XGBoost model to predict a property. In \resistor, the agent infers the complete system topology from resistor measurements. In \retro, the agent constructs a synthesis path for a target molecule under economic constraints. In \spectra, the agent elucidates the molecule from spectral data. In \wetlab, the agent identifies the ions present in a provided sample.}
    \label{fig:envs-input-output}
\end{extdatafig}

\begin{extdatafig}[H]
    \centering
    \includegraphics[width=\linewidth]{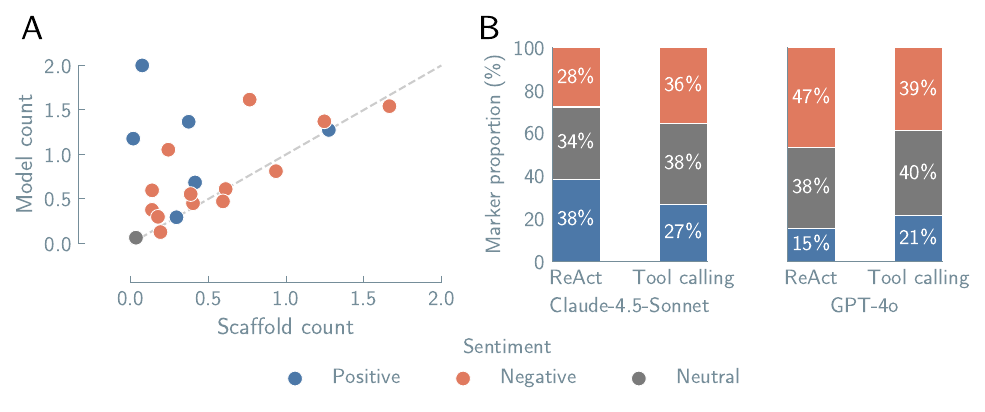}
    \caption{\textbf{Trace quality markers are more strongly influenced by the model than by the scaffold.}
    (A) Normalized effect sizes for each annotated marker. Most points lie above the diagonal, indicating that model differences exceed scaffold differences for the majority of markers.
    (B) Proportion of positive, negative, and neutral markers across scaffold-model combinations. The annotation process and the marker taxonomy are described in \Cref{app:manual_annotation_description}. 773 traces were manually annotated across all environments; detailed results per marker are in \Cref{app:marker_annotation_results}.}
    \label{fig:annotation-main-fig}
\end{extdatafig}

\begin{extdatafig}[H]
    \centering
    \includegraphics[width=1\linewidth]{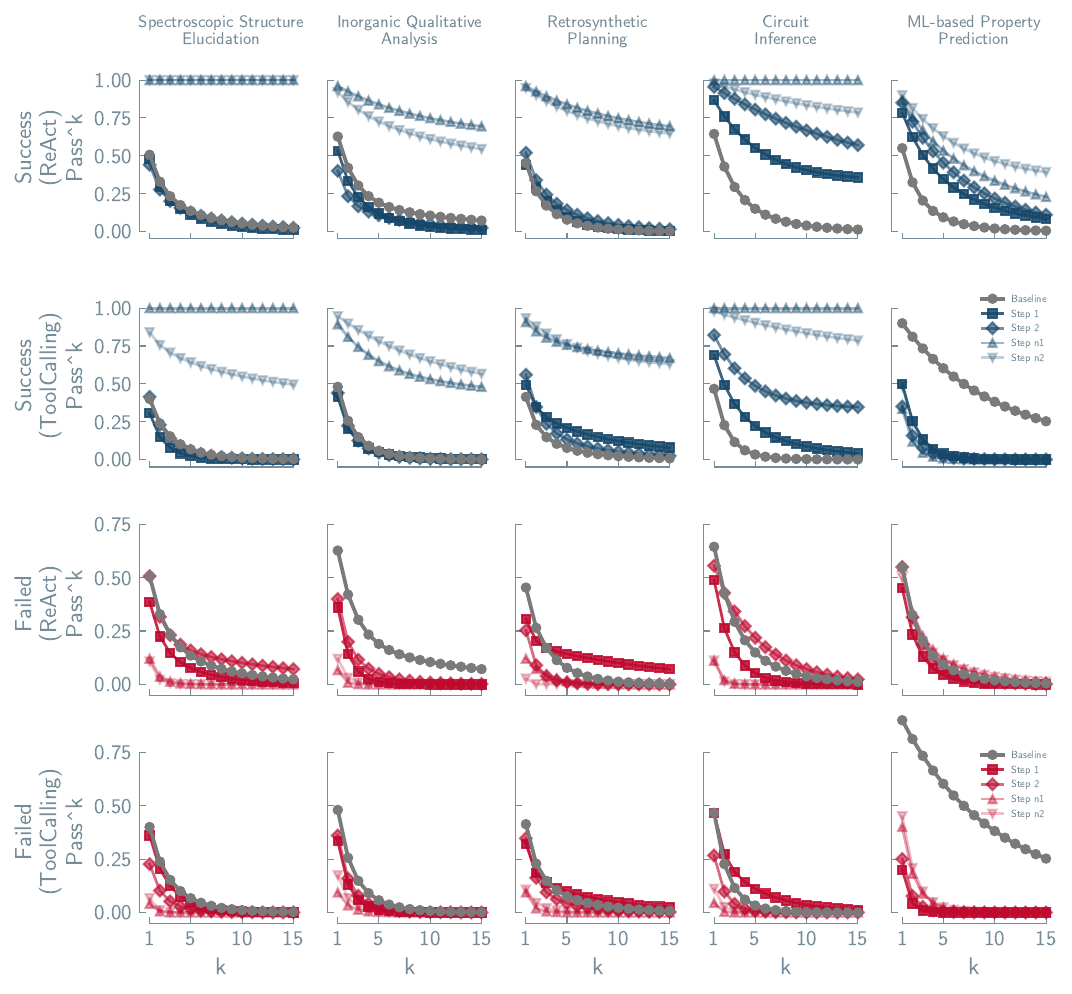}
    \caption{\textbf{\passhat decay under trace interventions is consistent across scaffold architectures.} \passhat as a function of $k$ under trace interventions, shown separately for ReAct and tool-calling agents.
      Rows correspond to interventions with traces from successful baseline runs (top two) and traces from failed baseline runs (bottom two), with columns representing each environment.
      Unlike \passat, \passhat requires all $k$ trials to succeed.}
      \label{fig:pass_hat_per_agent}
\end{extdatafig}

\begin{extdatafig}[H]
    \centering
    \includegraphics[width=1\linewidth]{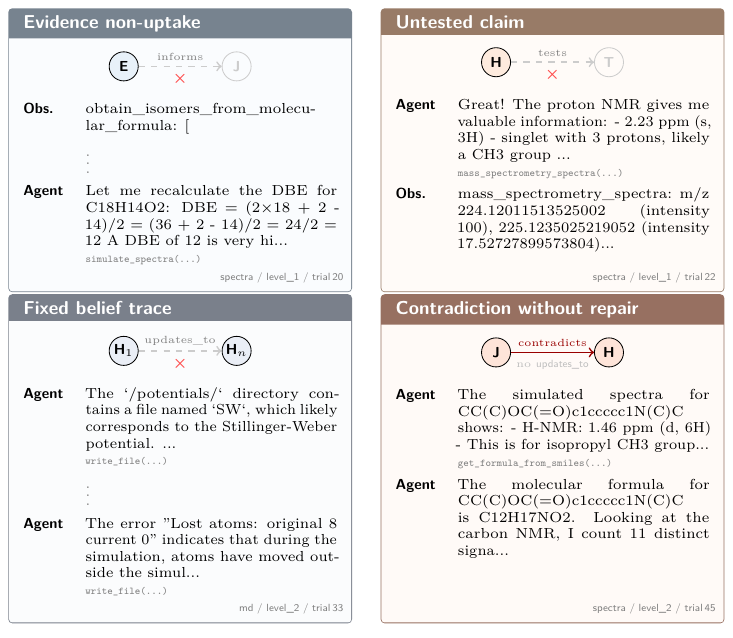}
    \caption{\textbf{Autonomous agents repeatedly fail to correct flawed reasoning: they ignore accessible evidence, leave hypotheses untested, cling to false beliefs, and accept contradictions.} Each panel lists model, environment, scope, and trial index. Evidence non-uptake (top left): Agent retrieves 20 isomers (including the correct one) but never consults the list, guessing instead. Untested claim (top right): From proton NMR, the agent infers methylene groups, acetyl, and benzene ring, but designs no test, skipping to mass spectrometry. Fixed belief trace (bottom left): LAMMPS simulation: agent mistakes \texttt{/potentials/SW} (a directory) for a file, blames \enquote{lost atoms} on heating rate, and never revises any of six hypotheses across three errors. Contradiction without repair (bottom right): Agent's proposed isopropyl ester predicts 6H doublet at 1.46 ppm (NMR) vs. experimental 3H at 1.43 ppm; agent notes the discrepancy but keeps structure as final answer. Full annotated traces are browsable at \url{https://lamalab-org.github.io/corral/\#explainers} (see also \Cref{app:illustrative-traces}).}
    \label{fig:reasoning-breakdown-examples}
\end{extdatafig}

\clearpage

\renewcommand\thefigure{\thesection.\arabic{figure}}

\renewcommand\thetable{\thesection.\arabic{table}}
\renewcommand\theequation{\thesection\arabic{equation}}
\setcounter{figure}{0}
\setcounter{table}{0}
\setcounter{equation}{0}

\begin{refsection}
\appendix
\clearpage

\begin{center}
{\LARGE\bfseries Appendix\par}
\end{center}
\vspace{1em}

\etocdepthtag.toc{app}
\begingroup
\etocsettagdepth{main}{none}
\etocsettagdepth{app}{subsection}
\etocsettocstyle{\subsection*{Contents}}{}
\tableofcontents
\endgroup

\section{Related work} \label{app:related-work}

\corral sits at the intersection of five lines of research: benchmarks for scientific agents, end-to-end AI scientist systems, the attribution of agent capability to the base model versus the scaffold, psychometric measurement of model ability, and process-level epistemic evaluation of AI-generated science.

\subsection{Benchmarks for scientific agents}

Existing benchmarks for scientific agents primarily score outcomes. Domain-specific evaluations such as ChemBench, MaCBench, and Lab-Bench probe knowledge and reasoning in chemistry, materials science, and biology through curated question sets \autocite{mirza2025framework, alampara2025macbench, laurent2024lab}. Workflow-oriented benchmarks including ScienceAgentBench, SciAgentGym, and Curie evaluate end-to-end execution of data analyses and discovery pipelines drawn from published work \autocite{chen2024scienceagentbench, shen2026sciagentgym, cui2025curie}. Replication-oriented evaluations such as Lessons-Learned, MLR-Bench, ResearchBench, and DeepResearch test whether agents can reproduce or extend specific results \autocite{alampara2025lessons, chen2025mlr, liu2025researchbench, du2025deepresearch}. AstaBench and BixBench evaluate agents across the full research pipeline, including manuscript-style outputs \autocite{bragg2025astabench, mitchener2025bixbench0}. Generalist agent benchmarks such as AgentBench, GAIA, Gaia2, MultiAgentBench, and ToolSandbox \autocite{liu2023agentbench, mialon2023gaia, froger2026gaia2, zhu2025multiagentbench, lu2025toolsandbox} measure capabilities that scientific agents draw on but were not designed for science.

These evaluations score what an agent produced, not how it produced it. A correct answer reached by lookup is scored identically to the same answer reached by hypothesis testing; an incorrect answer reached by careful inquiry is scored identically to one reached by confabulation. Concerns that benchmarks measure narrow skill rather than the underlying capability they purport to assess are long-standing \autocite{chollet2019measure}. \corral preserves the outcome metrics these benchmarks provide and adds standardized environments with graded task scopes and fine-grained subtask diagnostics.

\subsection{AI scientists}

A growing body of work develops LLM-based systems intended to conduct end-to-end research: hypothesis generation, experimental design, execution, and write-up. Sakana's AI Scientist \autocite{lu2026towards} produces submission-ready manuscripts; Kosmos \autocite{mitchener2025kosmos} reports cross-domain discoveries; ChemCrow \autocite{MBran2024augmenting}, Coscientist \autocite{Boiko2023autonomous}, the El Agente series \autocite{zou2025agente}, Organa \autocite{Darvish2025organa}, and AlphaEvolve \autocite{novikov2025alphaevolve0} target specific scientific or engineering domains. Multi-agent architectures such as SciAgents, ScienceClaw, and virtual laboratories structure the workflow with role specialization or knowledge-graph guidance \autocite{ghafarollahi2025sciagents, wang2026autonomous, swanson2025virtual}, building on earlier instantiations for protein design \autocite{ghafarollahi2024protagents} and finite-element mechanics \autocite{ni2024mechagents}; recent work also adapts the objective each sub-agent optimizes to mitigate reward hacking \autocite{du2025accelerating}. Complementary efforts target reasoning at the base-model level, training models to produce reflection and recursive refinement steps explicitly \autocite{buehler2025preflexor}. Alongside this systems-building effort, risk-focused perspectives argue that AI scientists introduce novel vulnerabilities requiring explicit safeguards across user intent, agent alignment, and environmental feedback \autocite{Tang2025}. These systems are deployed and reported on without a shared instrument for assessing the quality of the reasoning that produced their outputs. \corral is designed to fill this role: its environments, tools, scoring functions, and runner can be reused as building blocks for new agents and scaffolds, and the diagnostic and epistemic instruments introduced here can be applied to any system built on top. The environments can also serve as training substrates: each supplies reproducible tasks and scoring over agent trajectories, sufficient to optimize agents against criteria defined on the reasoning process \autocite{ma2026skillclaw0}.

\subsection{Scaffolds and the base-model/scaffold attribution}

The capability of an LLM-based agent reflects two contributions: the base model's token-level competence and the scaffold's orchestration logic. Prompt-based scaffolds such as Chain-of-Thought, ReAct, and self-consistency \autocite{wei2022chain, yao2022react, wang2022self}, and agentic scaffolds with tool access such as Toolformer, Voyager, and Reflexion \autocite{schick2023toolformer, wang2023voyager, shinn2023reflexion0}, are now standard, and gains over base models are routinely reported. The question of how much of any gain is the scaffold and how much is the base model is rarely answered directly. \textcite{meirelesinfluence} are among the few to address it, showing that effective scaffolding preserves coordination scaling laws that deteriorate without it. \corral varies model and scaffold orthogonally on a fixed task set, allowing each contribution to be read directly from the results.

\subsection{Item response theory for model evaluation}

Item response theory (IRT), developed for educational assessment \autocite{hori2022item, toland2014practical, lovelace2013best}, models benchmark performance as the interaction between latent ability and item characteristics such as difficulty and discrimination. Early NLP applications used IRT to construct evaluation datasets and demonstrated that item difficulty predicts model behavior more reliably than aggregate accuracy \autocite{lalor2016building, lalor2018understanding}. Bayesian IRT-based leaderboards introduced uncertainty-aware ability estimates \autocite{rodriguez2021evaluation}, and subsequent work showed that hidden implementation choices distort raw rankings in ways that IRT can correct \autocite{schilling2025lifting, zhou2026lost}. For LLM agents, IRT-inspired task selection has been used to preserve ranking fidelity across scaffolds and environments \autocite{ndzomga2026efficient}. We apply IRT to our diagnostic subtask items to estimate latent capability profiles per model--environment pair, separating ability from item difficulty in a way that aggregate Pass@$k$ does not.

\subsection{Epistemic evaluation of AI in science}

Process-level epistemic evaluation of AI-generated science remains underdeveloped. Field-level critiques are well established: \textcite{ioannidis2015meta} on irreproducibility in the empirical sciences, and \textcite{hao2026artificial} on machine learning, narrowing the hypothesis space scientists pursue. More directly on large language models, \textcite{Mitchell2023} survey the debate over whether such systems exhibit genuine understanding, noting that surface-level competence on benchmarks does not settle the underlying epistemic question. \textcite{pittphilsci28744} extend these concerns to LLM-based scientific systems specifically, arguing that current AI-for-science work conflates predictive accuracy with epistemic warrant and lacks operational instruments for assessing whether claims are justified in the philosophical sense \autocite{sep-knowledge-analysis}. \corral provides such an instrument. The behavioral analysis introduced here annotates reasoning traces as graphs of operations and detects structural patterns of productive inquiry, such as Popperian falsification, exploratory-to-confirmatory transitions, and active learning \autocite{popper2005logic, tukey1977exploratory, lindley1956measure, fedorov2013theory}, as well as patterns of epistemic failure, such as untested hypotheses, ignored evidence, and unresolved contradictions. This shifts the evaluation from whether an agent reached the right answer to the process by which it arrived there.

\subsection{Rise of AI Scientists} \label{app:rise_of_ai_scientists}

\begin{figure}[H]
    \centering
    \includegraphics[width=1\linewidth]{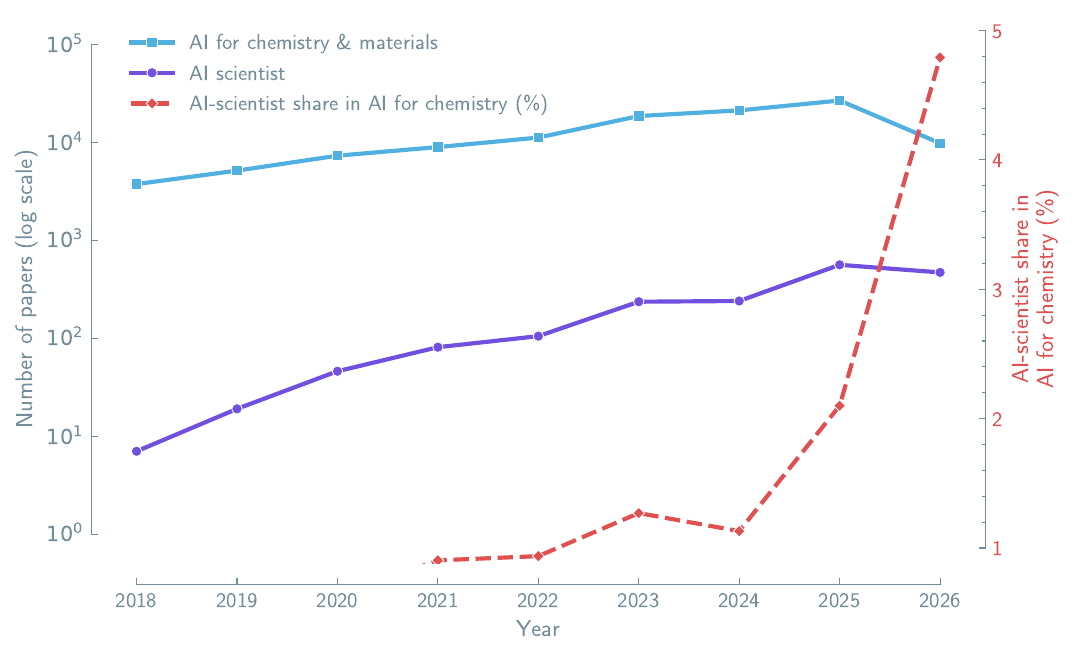}
    \caption{\textbf{Sharp rise of AI-scientist publications within chemistry and materials AI literature.} Annual publication counts (log scale) for the broader AI-for-chemistry-and-materials corpus versus a strict AI-scientist subset. The right axis shows the share of AI-scientist papers within the broader corpus, revealing a steep recent increase.}
    \label{fig:raise-scientists}
\end{figure}

Following \textcite{hao2026artificial}, we quantified the emergence of AI-scientist research within chemistry and materials science and contextualized its growth against the broader AI/ML-for-chemistry-and-materials literature. We constructed a narrow corpus targeting AI scientists, LLM agents, autonomous discovery systems, and closely related agentic workflows. Using OpenAlex \autocite{priem2022openalex}, we first retrieved the broader set of AI- and ML-related papers in chemistry and materials science, then applied a stricter keyword-based filter to isolate publications matching our AI-scientist definition. \Cref{fig:raise-scientists} reports yearly publication counts for both corpora on a log scale, along with the share of AI-scientist papers within the broader AI-for-chemistry literature.

Two patterns are immediately apparent. First, the broader AI-for-chemistry literature has reached roughly $10^{4}$ papers per year, while the strict AI-scientist subset has grown from about $10^{1}$ papers in 2018 to the upper $10^{2}$ range in the most recent partial year—already exceeding earlier full-year totals. Thus, although AI-scientist work remains much smaller than AI for chemistry as a whole, it is no longer a marginal subfield but a recognizable and rapidly growing research stream.

Second, the relative share of AI-scientist papers within the AI-for-chemistry literature is rising even faster than absolute counts. This strict subset accounted for only about 1\% of AI-for-chemistry output through 2024, increased to roughly 2\% in 2025, and approaches 5\% in the current partial year. Consequently, while AI-scientist research still represents a very small fraction of all chemistry and materials science publishing---approximately 0.01\% through 2024 and around 0.06\% in 2025---its footprint within the AI segment of the field is expanding rapidly.

\section{Glossary}
\label{app:glossary}

\begin{description}
    \item[Domain.] One of the eight scientific areas covered by the benchmark (e.g.,\ retrosynthetic planning, atomic force microscopy); see \Cref{sec:domains_and_environments}.
    \item[Scope.] A difficulty or ablation variant within a domain, varying task complexity, tool availability, or degree of agent guidance. Scopes are introduced per domain in \Cref{sec:domains_and_environments}.
    \item[Environment.] A concrete instance of a domain at a specific scope, formalized as a language decision process in \Cref{sec:formalism}. Each environment exposes tools, a task description, and a scoring function through the \corral HTTP interface.
    \item[Task.] A single problem an agent is asked to solve within an environment; may decompose into dependency-ordered subtasks (\Cref{sec:formalism}).
    \item[Subtask.] A component of a task whose output is forwarded as input to downstream subtasks in the dependency graph (\Cref{sec:formalism}).
    \item[Scaffold.] The orchestration layer wraps a base LLM: prompt formatting, tool routing, history management, and termination logic. The two scaffolds evaluated here are ReAct and structured tool calling (\Cref{sec:agents-implementations}).
    \item[Tool-documentation verbosity.] The level of detail retained in tool docstrings presented to the agent varied as an experimental axis (\Cref{sec:tool-verbosity}).
    \item[Configuration.] A particular combination of base model, scaffold, tool-documentation verbosity, and environment.
    \item[Trial.] A single end-to-end execution of an agent configuration on a task.
    \item[Trace.] The conversation history produced during a trial is used downstream for epistemological annotation (\Cref{sec:epistemological-graphs}).
    \item[Pass@$k$.] Probability that at least one of $k$ independent trials on a task succeeds. Estimated from $n \ge k$ trials by the unbiased estimator $1 - \binom{n-c}{k} / \binom{n}{k}$, where $c$ is the number of successful trials.
    \item[Pass$^k$.] Probability that all $k$ independent trials on a task succeed. A reliability measure, unlike Pass@$k$, requires every trial to succeed rather than at least one.
    \item[Language decision process (LDP).] The formal model of agent--environment interaction adopted here: a partially observable Markov decision process whose observations and actions are represented in text (\Cref{sec:formalism}).
    \item[ELPD-LOO.] Expected log predictive density estimated by leave-one-out cross-validation; used to compare the candidate hierarchical models in \Cref{app:binomial-subsec}.
    \item[Inter-annotator agreement ($\kappa$).] Cohen's $\kappa$ between two independent annotators on the same set of traces, reported in \Cref{app:reasoning-process}, to quantify the reliability of the manual review.
    \item[Epistemic operation.] A node type in the annotated reasoning graph, drawn from the fixed vocabulary \{hypothesis, test, evidence, judgment, update, commitment\}. Each operation corresponds to an identifiable epistemic act in the trace (\Cref{sec:epistemological-graphs}).
    \item[Item response theory (IRT).] A psychometric modeling framework that jointly estimates respondent ability and item difficulty from a matrix of correct/incorrect outcomes. We fit the two-parameter logistic (2PL) form, in which each item has a difficulty $b$ and a discrimination $a$, and each respondent a latent ability $\theta$ (\Cref{sec:app_irt}).
    \item[Latent ability ($\theta$).] The IRT parameter representing a model's capability on a set of items, separated from the items' own difficulty. Higher $\theta$ corresponds to a higher probability of success on items of a given difficulty.
    \item[Posterior predictive check (PPC).] A fitted Bayesian model's predictions for held-out observations are compared against the observed outcomes to assess calibration and generalization. Reported for the selected hierarchical model in \Cref{app:binomial-subsec}.
    \item[Capability item.] A question--answer item used to estimate latent model capability through IRT (\Cref{sec:qa}, \Cref{app:capability-items}).
    \item[Productive motif.] A graph template over the annotated trace that matches a recognized practice of disciplined inquiry: Popperian falsification, exploratory-to-confirmatory transition, active-learning-style test redesign (\Cref{app:pattern-taxonomies}).
    \item[Reasoning breakdown.] A graph template over the annotated trace that matches an established epistemic failure mode: untested claim, evidence non-uptake, fixed belief trace, contradiction without repair (\Cref{app:pattern-taxonomies}).
\end{description}

\section[Corral: Framework]{%
    \texorpdfstring{\corral \corrallogo: Framework}{Corral: Framework}%
} \label{sec:corral-framework}
\label{app:framework-details}
\begin{figure}[H]
    \centering
    \includegraphics[width=1\linewidth]{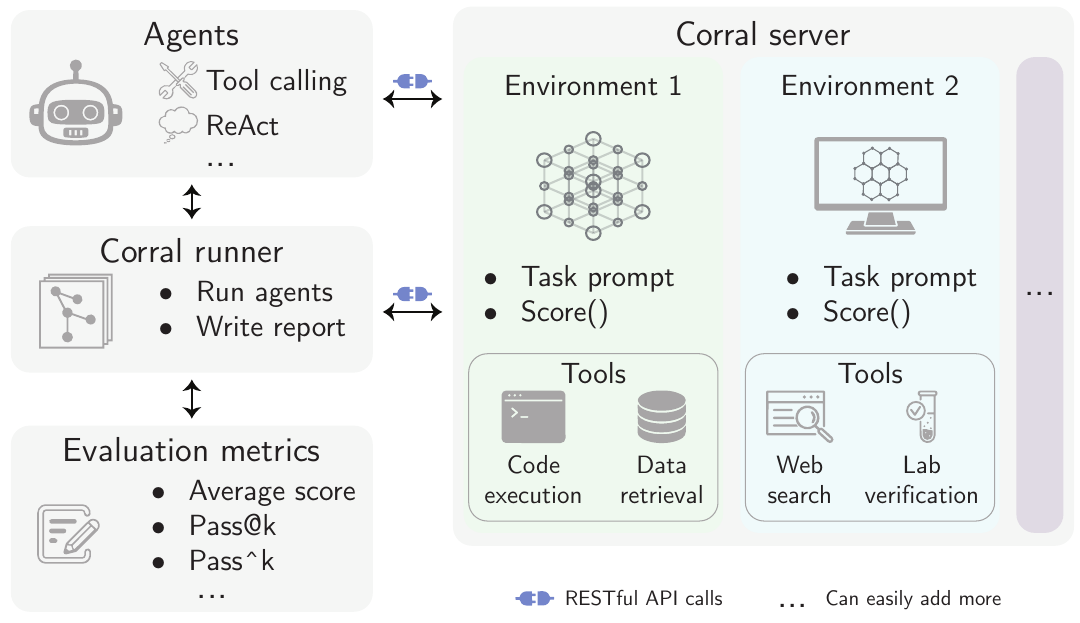}
    \caption{\textbf{\corral framework overview.} Environments are hosted in a FastAPI server; each exposes a task description, tools, and a scoring function. Agents interact with an environment by issuing HTTP requests to query information, check state, or submit answers. The \corral runner drives evaluation, handling checkpointing, aggregation, and metric computation. New agents, environments, and metrics extend the framework through the three subclass interfaces described in this section.}
    \label{fig:corral-framework-app}
\end{figure}

\corral is a composable evaluation framework that separates environments from agents through a standard HTTP interface. Environments are hosted by a FastAPI server that exposes task descriptions, tools, and scoring through a small set of RESTful endpoints (\Cref{tab:api-endpoints}, \Cref{fig:corral-framework-app}). Agents interact with an environment solely by issuing HTTP requests. The \corral runner drives evaluation by iterating trials over registered agent--environment pairs, checkpointing intermediate state, and aggregating results into per-configuration scores and diagnostics. The snippets below illustrate the three main extension points; the canonical API reference is maintained at \url{https://lamalab-org.github.io/corral}.

\paragraph{Tools.} Tools are formally subclasses of the \texttt{Tool} base class. The \texttt{@tool} decorator is a convenience shortcut that wraps a plain Python function into such a subclass. In both cases, the docstring of the tool is part of the contract between environment and agent: \corral exposes it to the agent as the tool's natural-language description, and its tagged sections are filtered by the tool-documentation verbosity axis (\Cref{sec:tool-verbosity}).

\begin{lstlisting}[caption={Defining a tool via the decorator shortcut.}]
from corral.backend.tool import tool

@tool
def my_custom_tool(input_param: str) -> str:
    """[BRIEF] One-sentence summary.

    Args:
        input_param: description of the parameter.
    Returns:
        description of the return value.
    """
    return f"Processed: {input_param}"
\end{lstlisting}

\paragraph{Environments.} An environment subclasses \texttt{Environment}, attaches its tools in the constructor, and overrides \texttt{get\_task\_prompt} and \texttt{score}. A dictionary of environment instances is then served over HTTP by \texttt{create\_benchmark\_server}.

\begin{lstlisting}[caption={Defining and serving an environment.}]
from corral.backend import Environment
from corral.backend.server import create_benchmark_server
import os, uvicorn

BASE_WORK_DIR = os.getenv("CORRAL_WORK_DIR")

class MyEnvironment(Environment):
    def __init__(self, task_id, problem, answer,
                 base_workdir=BASE_WORK_DIR):
        self.problem = problem
        self.correct_answer = answer
        super().__init__(task_id, base_workdir)
        self.add_tool(my_custom_tool)

    def get_task_prompt(self):
        return f"Solve this problem: {self.problem}"

    def score(self):
        if self.state.submitted_answer is None:
            return 0.0
        return 1.0 if self.state.submitted_answer == self.correct_answer else 0.0

environments = {
    "task_1": MyEnvironment("task_1", "Problem 1", "Answer 1"),
    "task_2": MyEnvironment("task_2", "Problem 2", "Answer 2"),
}

if __name__ == "__main__":
    app = create_benchmark_server(environments)
    uvicorn.run(app, host="0.0.0.0", port=8000)
\end{lstlisting}

\paragraph{Agents.} An agent subclasses \texttt{BaseAgent} and implements \texttt{run}. The agent drives inquiry through the router interface: \texttt{get\_task\_guide} returns the task prompt, \texttt{execute\_tool} invokes a tool, and \texttt{get\_llm\_response} routes a message list through the configured base model.

\begin{lstlisting}[caption={Defining a custom scaffold.}]
from corral.agents import BaseAgent
from corral.agents.prompt_utils import create_prompt
from corral.agents.utils import LiteLLMMessage

class MyAgent(BaseAgent):
    def __init__(self, model: str, **kwargs):
        super().__init__(model, **kwargs)

    def run(self, interface, task_id: str) -> str:
        task_guide = interface.get_task_guide(task_id)
        messages = create_prompt(task_guide)
        # loop: call base LLM, parse action, execute tool,
        # append observation, repeat until termination
        response = self.get_llm_response(messages)
        return response
\end{lstlisting}

\paragraph{Runner.} The \corral runner iterates over tasks and trials, dispatches agent calls through the router, persists intermediate state for fault tolerance, and returns a unified result summary. Experiment execution and environment definition are kept separate.

  \begin{lstlisting}[caption={Runner invocation with custom diagnostic metric.}]
  from corral import CorralRouter, CorralRunner
  from corral.agents import MyAgent


  # Connect to the environment server
  router = CorralRouter()

  # Configure the agent (model + scaffold)
  agent = MyAgent(model="gpt-4o", temperature=0.7, max_iterations=100)

  # Create runner and register custom metrics
  runner = CorralRunner(router, agent)

  # Run evaluation
  result = runner.bench(
    ["task_1", "task_2"],
    trials_per_task=3,
    verbose=True
  )

  # Results include default + custom metrics
  print(f"Average score: {result.metrics['average_score']}")
  \end{lstlisting}

\subsection{Prompt construction}
\label{app:prompt-construction}

The message sequence delivered to the base model is assembled from four independently authored components.

\paragraph{System prompt.} A static instruction, identical across all configurations, that establishes the agent's role and a context management directive.

\paragraph{Task prompt.} Each environment implements a \texttt{get\_task\_prompt()} method that assembles a structured prompt from the task definition (see the endpoints in \Cref{tab:api-endpoints}). All environments follow a common template: (i)~a task name, (ii)~a natural-language description of the scientific goal, (iii)~the required submission format (e.g., a SMILES string, a list of ions), and (iv)~any available input data. For task groups, the input data include the artifacts forwarded from predecessor subtasks in the dependency graph, plus initial parameters such as target molecule or material composition. Some environments prepend a brief role preamble or append workspace instructions when filesystem tools are available.

\paragraph{Scaffold template.} The task prompt is inserted into a scaffold-specific user template. For ReAct, the template embeds the task prompt together with a textual tool-usage guide and instructs the model to respond in XML-tagged thought---action---observation turns; the final answer is submitted via a \texttt{<final\_answer>} tag. For structured tool calling, the template contains only the task prompt; tool schemas are passed separately through the native function-calling API, and the final answer is submitted with a \enquote{Final Answer:} keyword prefix.

\paragraph{Tool descriptions.} Filtered by the verbosity axis (\Cref{sec:tool-verbosity}) and injected either inline in the user prompt (ReAct) or via the function-calling API (tool calling).

\noindent  Only the task prompt varies across environments; the system prompt, scaffold template, and tool-injection mechanism are held constant.

\subsection{Example of different verbosity}
\label{sec:verobisty_examples}
The example boxes below render the docstring of the tool search\_by\_smiles at each of the three verbosity levels. \Cref{sec:tool-verbosity} defines the basis on which information is controlled for different verbosity levels. \Cref{tab:verbosity-tags} lists the tagged sections and their content.

  \begin{table}[H]
  \centering
  \caption{\textbf{Tagged sections used in tool docstrings.} Each tool's documentation is authored with the sections below. At each verbosity level, a cumulative subset of these sections is retained (\Cref{sec:tool-verbosity}).}
  \label{tab:verbosity-tags}
  \renewcommand{\arraystretch}{1.15}
  \begin{tabular}{@{}lp{0.62\linewidth}@{}}
    \toprule
    \textbf{Tag} & \textbf{Content} \\
    \midrule
    \texttt{[BRIEF]}               & One-sentence functional summary \\
    \texttt{[DETAILED]}            & Extended description of the function \\
    \texttt{[PROCEDURAL]}          & Step-by-step usage instructions \\
    \texttt{[CONTEXTUAL]}          & When and why to use this tool \\
    \texttt{[WORKFLOW\_INTEGRATION]}& How this tool fits into multi-step workflows \\
    \texttt{[SYNTACTICAL]}         & Argument formatting conventions \\
    \texttt{[EXAMPLES]}            & Concrete input/output examples \\
    \texttt{[RAISES]},\;\texttt{[LIMITATIONS]} & Error conditions and known constraints \\
    \bottomrule
  \end{tabular}
  \end{table}

\begin{hypobox}{Verbosity: Brief}
  Search the NMRShift database for entries matching or chemically similar to the given SMILES.

  \medskip
\textbf{  \textsc{Arguments}}

  \textbf{smiles}( \texttt{str})\\
  \quad The SMILES representation of the compound to search for in the NMRShift database

  \textbf{top\_k} (\texttt{int})\\
  \quad The maximum number of results to return. Defaults to 10

  \medskip
\textbf{  \textsc{Returns}}

  \quad A list of dictionaries containing the most relevant entries from the NMRShift database
  \end{hypobox}

 \begin{hypobox}{Verbosity: Workflow}
    Search the NMRShift database for entries matching or chemically similar to the given SMILES.

    This function searches the NMRShift database for entries that match or are chemically similar to the provided SMILES string. It uses a vector database search to find the top
  \texttt{top\_k} results based on chemical similarity. The function returns a list of matching entries, each containing relevant information about the compound.

    \textbf{When to use this tool:}
    \begin{itemize}[nosep]
      \item Use it to validate your proposed molecule against the NMRShift database.
      \item Use it when you have a SMILES string and want to find related compounds in the NMRShift database.
      \item When you need to retrieve chemical shifts or other NMR-related information for a specific compound.
    \end{itemize}

    \textbf{How this tool works:}
    \begin{itemize}[nosep]
      \item It uses a vector database search in the \texttt{"nmrshiftdb2"} collection.
      \item The search uses chemical embeddings generated by the \texttt{"ibm-research/MoLFormer-XL-both-10pct"} model.
      \item The function retrieves the top \texttt{top\_k} results based on chemical similarity.
      \item Each result contains relevant information about the compound, such as its SMILES, chemical shifts, and other properties.
    \end{itemize}

    \textbf{Typical workflow integration:}
    \begin{enumerate}[nosep]
      \item Have a SMILES representation of the molecule you want to search for, or the SMILES of a structure that you think can be valid.
      \item Apply this tool with the SMILES string to search for matching or chemically similar entries in the NMRShift database.
      \item Use the retrieved entries to validate your proposed molecule, compare chemical shifts, or gather additional information. You can also use \texttt{get\_formula\_from\_smiles} to
   validate the molecular formula.
    \end{enumerate}

    \medskip
   \textbf{ \textsc{Arguments}}

    \textbf{smiles} (\texttt{str})\\
    \quad The SMILES representation of the compound to search for in the NMRShift database\\
    \quad \textit{It should be a valid SMILES notation that can be processed by the vector database search.}

    \textbf{top\_k} (\texttt{int})\\
    \quad The maximum number of results to return. Defaults to 10\\
    \quad \textit{It should be a positive integer.}

    \medskip
    \textbf{\textsc{Returns}}

    \quad A list of dictionaries containing the most relevant entries from the NMRShift database. Each dictionary contains relevant information about the compound. Results are sorted by
  similarity score in descending order.
  \end{hypobox}

\begin{hypobox}{Verbosity: Comprehensive}
    Search the NMRShift database for entries matching or chemically similar to the given SMILES.

    This function searches the NMRShift database for entries that match or are chemically similar to the provided SMILES string. It uses a vector database search to find the top
  \texttt{top\_k} results based on chemical similarity. The function returns a list of matching entries, each containing relevant information about the compound.

    \textbf{When to use this tool:}
    \begin{itemize}[nosep]
      \item Use it to validate your proposed molecule against the NMRShift database.
      \item Use it when you have a SMILES string and want to find related compounds in the NMRShift database.
      \item When you need to retrieve chemical shifts or other NMR-related information for a specific compound.
    \end{itemize}

    \textbf{How this tool works:}
    \begin{itemize}[nosep]
      \item It uses a vector database search in the \texttt{"nmrshiftdb2"} collection.
      \item The search uses chemical embeddings generated by the \texttt{"ibm-research/MoLFormer-XL-both-10pct"} model.
      \item The function retrieves the top \texttt{top\_k} results based on chemical similarity.
      \item Each result contains relevant information about the compound, such as its SMILES, chemical shifts, and other properties.
      \item The function returns a list of matching entries, each represented as a dictionary containing the relevant information.
    \end{itemize}

    \textbf{Typical workflow integration:}
    \begin{enumerate}[nosep]
      \item Have a SMILES representation of the molecule you want to search for, or the SMILES of a structure that you think can be valid.
      \item Apply this tool with the SMILES string to search for matching or chemically similar entries in the NMRShift database.
      \item Use the retrieved entries to validate your proposed molecule, compare chemical shifts, or gather additional information. You can also use \texttt{get\_formula\_from\_smiles} to
   validate the molecular formula.
    \end{enumerate}

    \textbf{Usage examples:}
    \begin{lstlisting}[language=Python, basicstyle=\ttfamily\small]
  search_by_smiles("CCO")
  search_by_smiles("C1=CC=CC=C1")
  search_by_smiles("C(C(=O)O)N")
  search_by_smiles("C1=CC=C(C=C1)C(=O)O")
    \end{lstlisting}

    \textbf{Exceptions:} If an error occurs during the search process, such as errors in the embedding of the query. This can also occur if the \texttt{top\_k} parameter is not a positive
  integer. Try another tool.

    \textbf{Known Limitations:}
    \begin{itemize}[nosep]
      \item The function requires a valid SMILES string that can be processed by the vector database search.
      \item The search results are limited to the \texttt{top\_k} parameter, which defaults to 10.
      \item The search is based on chemical embeddings, which may not capture all chemical similarities perfectly.
    \end{itemize}

    \medskip
    \textbf{\textsc{Arguments}}

    \textbf{smiles} (\texttt{str})\\
    \quad The SMILES representation of the compound to search for in the NMRShift database\\
    \quad \textit{The SMILES string representing the chemical structure of the molecule to search for in the NMRShift database. It should be a valid SMILES notation that can be processed
  by the vector database search.}\\
    \quad Valid SMILES string\\
    \quad Examples: \texttt{"CCO"}, \texttt{"C1=CC=CC=C1"}, \texttt{"C(C(=O)O)N"}, \texttt{"C1=CC=C(C=C1)C(=O)O"}

    \textbf{top\_k} (\texttt{int})\\
    \quad The maximum number of results to return. Defaults to 10\\
    \quad \textit{The maximum number of search results to return from the NMRShift database. It should be a positive integer.}\\
    \quad Any positive integer (e.g., 10, 20, 50)\\
    \quad Examples: \texttt{10}, \texttt{20}, \texttt{50}

    \medskip
    \textbf{\textsc{Returns}}

    \quad A list of dictionaries containing the most relevant entries from the NMRShift database. Each dictionary contains relevant information about the compound, such as its SMILES,
  chemical shifts, and other properties. The results are sorted by similarity score in descending order.\\
    \quad Example: \texttt{[\{"entry\_id": "nmrshiftdb2:234", "compound\_name": "Benzene", "smiles": "c1ccccc1", ...\}]}
  \end{hypobox}

\include{tables/framework_endpoint}

\section{Domains and environments}
\label{app:environment-details}

Each of the eight domains described in Methods exposes multiple task scopes; each scope is realized as a separate environment. The eight domains are summarised in \Cref{fig:envs-input-output}, which shows the input the agent receives and the artifact it produces in each case. An overview of the characteristics of each domain is provided in \Cref{tab:env-summary-table}. Full per-scope tool lists, scoring rubrics, and task specifications are maintained at \url{https://lamalab-org.github.io/corral/#environments}. One environment, the inorganic qualitative analysis (\textit{wetlab}), is used as a worked example below (\Cref{app:wetlab-section}).

\begin{table}[H]
    \centering
    \caption{\textbf{Overview of the \corral environments.}For each of the eight scientific domains, the table reports: (i) the number of difficulty levels, (ii) the number of distinct tasks per level, (iii) the total number of domain‑specific tools available to the agent, (iv) the typical trace length, and (v) the backing: the engine, instrument, dataset, or library on which the environment is built. The trace length is defined as the mean number of agent‑generated messages (excluding system and initial user prompts), averaged across all models, agent architectures, and verbosity settings.}
    \label{tab:env-summary-table}
    
\small

\newcolumntype{D}{>{\raggedright\arraybackslash\hsize=.75\hsize}X}
\newcolumntype{B}{>{\centering\arraybackslash\hsize=1.25\hsize}X}
\begin{tabularx}{\linewidth}{DccccB} 
\toprule
  Domain & Scopes & \makecell{Tasks\\per scope} & Tools & \makecell{Trace\\length} & Backing \\
  \midrule
  AFM experiment execution & 4 & 1 & 6 & 26.3 & Live atomic force microscope in the loop (DriveAFM (Nanosurf)\autocite{nanosurf_driveafm})\\
  Adsorption surface construction & 1 & 3 & 15 & 19.6 & Surface generation via \texttt{pymatgen} \\
  Molecular simulation & 2 & 2-3 & 8 & 30.4 & LAMMPS simulation engine \\
  ML-based property prediction & 1 & 3 & 14 & 16.6 & Materials Project API dataset\autocite{jain2013commentary} \\
  Circuit inference & 1 & 6 & 9 & 15.0 & Custom circuit generation engine \\
  Retrosynthetic planning & 3 & 8 & 15 & 25.5 & Novel tool suite built on an in-house reaction-template library assembled from scratch\autocite{Kannas2022rxnutils}  \\
  Spectroscopic structure elucidation & 2 & 20 & 16 & 15.1 & Manually curated spectral dataset\autocite{Binev2007prediction, Banfi2008nmr, Jablonka2022making} \\
  Inorganic qualitative analysis & 3 & 10 & 14 & 39.4 & Thermodynamic equilibrium engine\autocite{leal2015reaktoro} coupled to a custom color-mixing engine \\
  \bottomrule
\end{tabularx}
\end{table}

\subsection{Inorganic Qualitative Analysis Environment Example} \label{app:wetlab-section}

\begin{figure}[H]
    \centering
    \includegraphics[width=1\linewidth]{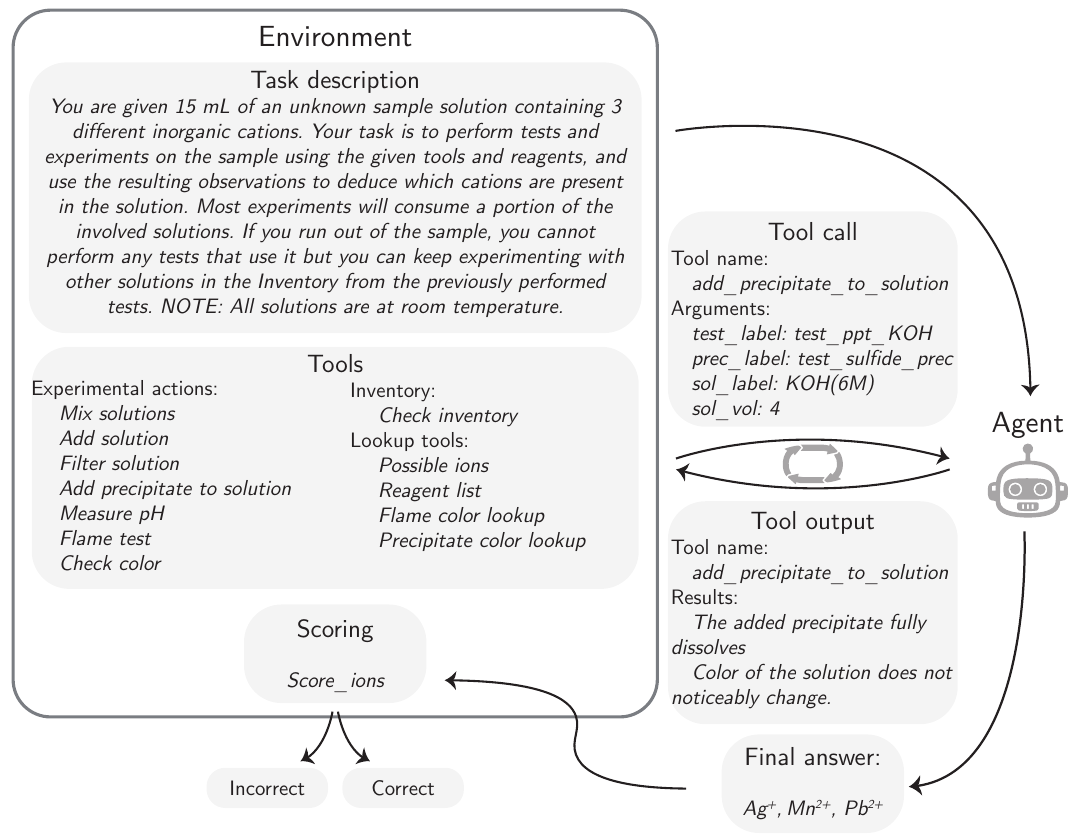}
    \caption{\textbf{Overview of the Inorganic Qualitative Analysis environment.} The agent's workflow consists of three main phases: (1) Initialization: The prompt is constructed with the task description, a list of available tools (including their arguments), and the required response format. (2) Iterative Interaction: The agent calls tools to perform experiments (e.g., pH measurement, flame tests; a detailed view of the tools is provided in \Cref{tab:wetlab-tools}) and can reference external information, such as lists of possible ions, reagent properties, color indicators, and the current inventory. (3) Final Scoring: The agent outputs a list of ions present in the sample. The environment scores this answer by comparing it to the ground truth, returning a binary score based on its correctness.}
    \label{fig:wetlab-app}
\end{figure}

\Cref{fig:wetlab-app} overviews the agent workflow in this environment.

\section{Capability assessment items}
\label{app:capability-items}

The latent capability profiles in \Cref{sec:app_irt} are estimated from a bank of question-answer items curated for each environment. For every environment, items are partitioned into a knowledge set that probes domain-specific facts required to operate the environment and a reasoning set that requires multi-step inference over domain concepts. \Cref{tab:qa-know-examples} and \Cref{tab:qa-reas-examples} show one representative item of each type per environment. Per-model scores on the full bank are in \Cref{tab:qa-scores}.

\subsection{Item generation protocol}

Benchmark items follow a structured review protocol.
Items are authored by the domain expert responsible for the corresponding
environment.
Two categories of items are distinguished:
\emph{knowledge} items, which test factual and procedural understanding of the
domain (e.g.,\ instrument settings, simulation parameters),
and \emph{reasoning} items, which require multi-step logical inference, scenario
analysis, or quantitative problem-solving within the domain. Each collection of items is submitted as a pull request. Two independent expert reviewers then assess the items against a predefined rubric (see \Cref{par:qa-rubric}). Items that do not pass receive change requests and return to the review stage for reassessment. Once approved, items undergo a final review and are merged into the benchmark.
\Cref{fig:item-generation-protocol} illustrates at a high level the protocol followed. After consolidation of the items, the dataset is pushed and maintained on Hugging Face  \autocite{lab_of_kevin_jablonka_at_uni_jena_2026_qas} for versioning.

\resizebox{\textwidth}{!}{\definecolor{boxfill}{HTML}{F1EFE8}
\definecolor{boxstroke}{HTML}{888780}
\definecolor{phasetext}{HTML}{A3A29C}
\definecolor{subtitletext}{HTML}{73726C}

\tikzset{
  process/.style={
    rectangle, rounded corners=4pt,
    draw=boxstroke, fill=boxfill,
    minimum width=4.8cm, minimum height=1.1cm,
    align=center, font=\sffamily\small,
    line width=0.4pt,
  },
  smallprocess/.style={
    rectangle, rounded corners=4pt,
    draw=boxstroke, fill=boxfill,
    minimum width=3.4cm, minimum height=0.8cm,
    align=center, font=\sffamily\small,
    line width=0.4pt,
  },
  decision/.style={
    diamond, aspect=2.2,
    draw=boxstroke, fill=boxfill,
    minimum width=2.6cm, minimum height=1.2cm,
    align=center, font=\sffamily\small\bfseries,
    line width=0.4pt, inner sep=1pt,
  },
  terminal/.style={
    rectangle, rounded corners=10pt,
    draw=boxstroke, fill=boxfill,
    minimum width=3.4cm, minimum height=0.8cm,
    align=center, font=\sffamily\small\bfseries,
    line width=0.4pt,
  },
  arr/.style={
    -{Stealth[length=5pt, width=4pt]},
    draw=boxstroke, line width=0.6pt,
  },
  phaselabel/.style={
    font=\sffamily\scriptsize, text=phasetext,
  },
  subtitle/.style={
    font=\sffamily\scriptsize, text=subtitletext,
  },
  dashedsep/.style={
    draw=phasetext, line width=0.3pt, dashed, dash pattern=on 3pt off 3pt,
  },
}

\begin{tikzpicture}[node distance=0.9cm]
\label{fig:item-generation-protocol}

\node[process] (draft)
  {\textbf{Expert drafts QA items}\\[1pt]
   {\scriptsize\color{subtitletext} Knowledge + reasoning questions}};

\node[smallprocess, below=of draft] (pr)
  {\textbf{PR created}};

\node[process, below=of pr] (review)
  {\textbf{Two expert reviewers}\\[1pt]
   {\scriptsize\color{subtitletext} Assess against rubric}};

\node[decision, below=1.0cm of review] (decide)
  {Approved?};

\node[process, below=1.0cm of decide] (merge)
  {\textbf{Final review + merge}\\[1pt]
   {\scriptsize\color{subtitletext} Items added to benchmark}};

\node[terminal, below=0.8cm of merge] (done)
  {Benchmark updated};

\node[process, left=2.8cm of decide] (changes)
  {\textbf{Changes requested}\\[1pt]
   {\scriptsize\color{subtitletext} Author revises items}};

\draw[arr] (draft) -- (pr);
\draw[arr] (pr) -- (review);
\draw[arr] (review) -- (decide);
\draw[arr] (decide) -- node[right, subtitle, xshift=2pt] {Yes} (merge);
\draw[arr] (merge) -- (done);

\draw[arr] (decide.west) -- node[above, subtitle] {No} (changes.east);
\draw[arr] (changes.north) |- (review.west);

\coordinate (leftedge) at ([xshift=-0.6cm]changes.west);
\coordinate (rightedge) at ([xshift=0.6cm]draft.east);

\node[phaselabel, anchor=east] at (leftedge |- draft) {Item generation};

\coordinate (sep1) at ($(pr.south)!0.5!(review.north)$);
\draw[dashedsep] (leftedge |- sep1) -- (rightedge |- sep1);
\node[phaselabel, anchor=east] at (leftedge |- review) {Quality assessment};

\coordinate (sep2) at ($(decide.south)!0.5!(merge.north)$);
\draw[dashedsep] (leftedge |- sep2) -- (rightedge |- sep2);
\node[phaselabel, anchor=east] at (leftedge |- merge) {Acceptance};

\end{tikzpicture}}

\paragraph{Item format and scoring rubric}

Every benchmark item is a four- to six-option multiple-choice question stored in JSON.
Each item carries a unique \texttt{uuid}, a human-readable \texttt{name}, a short
\texttt{description}, and a list of \texttt{keywords} whose first entry
identifies the environment, and whose second entry is either
\texttt{requires\_knowledge} or \texttt{requires\_reasoning}.
The question text and answer options are stored in an \texttt{examples} list, where
each entry contains an \texttt{input} string (the question) and a
\texttt{target\_scores} dictionary mapping every option to~$1$ (correct) or~$0$
(incorrect). Exactly one option is marked correct.
All items use the metric \texttt{multiple\_choice\_grade}: a model receives a score
of~$1$ if it selects all and only the correct option and~$0$ otherwise.
A canary string is prepended to every item to enable the detection of the benchmark
contamination in training corpora.
Listing \Cref{lst:qa-example} shows an example item.

\begin{hypobox}{Example format of a benchmark item in JSON format.}
  \begin{lstlisting}[
  language={},
  basicstyle=\ttfamily\small,
  caption={},
  label={lst:qa-example},
  frame=single,
  xleftmargin=1em,
  xrightmargin=1em,
  breaklines=true
]
{
  "name": "AFM_QA_25",
  "uuid": "5f17549c-7bf4-4eb6-a547-7274c7dddcb3",
  "description": "Reasoning-based",
  "examples": [
    {
      "input": "Which scanning mode best balances tip safety?",
      "target_scores": {
        "Contact
mode": 0,
        "Tapping mode": 1,
        "Lateral force mode": 0,
        "Conductive AFM" : 0
      }
    }
  ],
  "keywords": ["requires_knowledge", "AFM", 
               "microscopy", "scanning"],
  "metrics": ["multiple_choice_grade"],
  "preferred_score": "multiple_choice_grade"
}
\end{lstlisting}
\end{hypobox}

\paragraph{Quality-assessment procedure}
\label{par:qa-rubric}

Each pull request is reviewed by two independent expert reviewers who evaluate every item against the following rubric:
 
\begin{enumerate}
  \item \textit{Correctness.}
    The marked answer must be verifiably correct and all distractors genuinely
    incorrect; reviewers independently re-derive calculations where applicable.
    Items containing factual errors or misconceptions are rejected. Incorrect options must be plausible enough to require domain knowledge to
    eliminate.

  \item \textit{Clarity and ambiguity.}
    Questions must be unambiguous, use consistent notation (e.g.\
    \texttt{siunitx} units, \verb|\ce{}| formulae, SMILES), expand
    domain-specific acronyms, and be compatible with the evaluation harness.
 
  \item \textit{Originality.}
    Questions must not be readily findable online or in standard textbooks;
    items that can be answered by retrieving a verbatim source are rejected.
 
  \item \textit{Scope and relevance.}
    Each item must be tightly linked to its environment's domain and must
    measure domain expertise rather than general instruction-following ability.
    Out-of-scope questions are rejected.
 
  \item \textit{Difficulty and diversity.}
    Items should pose a genuine challenge to a domain expert, and the item set. Each environment should cover a broad range based on the tasks in the environment.
 
  \item \textit{Format and metadata.}
    Keyword ordering, the \texttt{requires\_knowledge}/\texttt{requires\_reasoning} tag, and the format of the JSON are verified for consistency.
\end{enumerate}

\begin{table}[H]
    \centering
    \caption{\textbf{Example knowledge-based questions.} For each environment, one representative factual question is shown, testing domain-specific knowledge required to operate the environment. The correct answer is indicated in the last column.}
    \label{tab:qa-know-examples}
    \resizebox{\textwidth}{!}{
    \begin{tabularx}{\textwidth}{p{2.5cm} X c}
\toprule
\textbf{Environment} & \textbf{Question} & \textbf{Answer} \\
\midrule
AFM experiment execution & Which scanning mode best balances tip safety? A. Contact mode  B. Tapping mode  C. Lateral force mode  D. Conductive AFM & B \\
\addlinespace
Adsorption surface construction & What is the coordination number of copper in Cu2O? A. 2  B. 4  C. 6  D. 8 & A \\
\addlinespace
Molecular simulation & You are simulating a bulk crystalline material (no surfaces). What boundary setting should you use? A. ppp  B. ppf  C. fff  D. pps & A \\
\addlinespace
ML-based property prediction & Which ensemble method combines predictions by averaging? A. Bagging  B. Boosting  C. Stacking  D. Random Forest & A \\
\addlinespace
Circuit inference & What is the equivalent resistance of n resistors each of resistance R connected in parallel? A. nR  B. R/n  C. R  D. n/R & B \\
\addlinespace
Retrosynthetic planning & What products are commonly formed from sydnone-alkyne cycloadditions? A. Pyrazoles  B. Imidazoles  C. Triazines  D. Oxadiazoles & A \\
\addlinespace
Spectroscopic structure elucidation & Which analysis method is most appropriate for determining if a molecule contains halides (C-X)? A. IR  B. UV-Vis  C. COSY  D. MS & D \\
\addlinespace
Inorganic qualitative analysis & Which one of the oxides below is white and insoluble in water? A. HgO  B. BaO  C. CuO  D. ZnO  E. Ag2O & D \\
\addlinespace
\bottomrule
\end{tabularx}
    }
\end{table}

\begin{table}[H]
    \centering
    \tiny
    \caption{\textbf{Example reasoning-based questions.} For each environment, one representative reasoning question is shown, requiring multi-step inference over observations or domain concepts. Answer options are truncated for brevity. The correct answer is indicated in the last column.}
    \label{tab:qa-reas-examples}
    \resizebox{\textwidth}{!}{
    \begin{tabularx}{\textwidth}{p{2.5cm} X c}
\toprule
\textbf{Environment} & \textbf{Question} & \textbf{Answer} \\
\midrule
AFM experiment execution & The following observation were collected during the task:   Perform three sequential high-quality scans of the same 10x10 µm² area using identical scanning parameters (P gain: 100, I gain: 6000, D gain: 10, Time per line: 0.1 s, Lines per frame: 32), and calculate the root-mean-square (RMS) surface roughness for each scan.  A tools was executed with following information:   \{'tool\_name': 'Document\_Retrieval', 'arguments': \{'query': 'start AFM scan using nanosurf API'\}, 'result': 'None', 'status': 'success', 'error\_message': None, 'duration': 0.8432333999990078,\}   What should the agent do next? A. Try using a different query in the document retrieval tool.  B. Call the same tool again with the same query.  C. Take no further action and stop the process.  D. Terminate the experiment, as it appears to be a dead end. & A \\
\addlinespace
Adsorption surface construction & Which of the following computational parameters or choices directly and fundamentally influence the accuracy of calculated molecular adsorption energies in periodic Density Functional Theory (DFT) simulations? A. The number of atomic layers included in the surface slab model.  B. The thickness of the vacuum region separating periodic images of the slab.  C. The density of the k-point mesh used for Brillouin zone integration.  D. The total simulation time of a molecular dynamics trajectory.  E. The computational power (CPU cores, RAM) of the hardware cluster.  F. The visualization software used to render atomic structures.  G. The convergence criterion for the self-consistent field (SCF) loop.  H. The type of thermostat used in a molecular dynamics simulation.  I. The compiler optimization flags used for the DFT code. & A \\
\addlinespace
Molecular simulation & This is a task from an agent benchmark. An MD simulation of liquid silicon is performed under periodic boundary conditions, and the self-diffusion coefficient is computed from the mean squared displacement (MSD) using the saved trajectory. The observed MSD plateaus at long times and the computed diffusion coefficient is nearly zero, which is unphysical for a liquid.  Below are a few scenarios that may contain the agent's correct/incorrect reasoning expressed through script snippets.  Scenario A: \# Dump wrapped coordinates dump 1 all custom 100 traj.lammpstrj id type x y z  Scenario B: \# Dump unwrapped coordinates dump 1 all custom 100 traj.lammpstrj id type xu yu zu  Scenario C: \# Dump scaled (fractional) wrapped coordinates dump 1 all custom 100 traj.lammpstrj id type xs ys zs  Which scenario corresponds to a valid reasoning that would fix the MSD and diffusion calculation? A. Scenario A  B. Scenario B  C. Scenario C  D. None of the above & B \\
\addlinespace
ML-based property prediction & You trained five different models and evaluated them on your test set. Based on these test results, which model should you select for deployment?  Model A: Test \$R\textasciicircum{}2\$ = 0.79 Model B: Test \$R\textasciicircum{}2\$ = 0.85 Model C: Test \$R\textasciicircum{}2\$ = 0.72 Model D: Test \$R\textasciicircum{}2\$ = 0.83 Model E: Test \$R\textasciicircum{}2\$ = 0.81 A. Model B  B. Model D  C. Model A  D. Not enough context to make a meaning choice & D \\
\addlinespace
Circuit inference & What is the effective resistance between A and C for the circuit with following topology \{"resistors": \{"R1": 10.0/4, "R2": 10.0, "R3": 30.0, "R4": 20.0\}, "connections": [["A", "B", "R1"], ["B", "C", "R2"], ["B", "C", "R3"], ["A", "C", "R4"]]\}\}? A. 20/3  B. 40/7  C. 10/3  D. Zero & A \\
\addlinespace
Retrosynthetic planning & A synthesis requires building a spirocyclic benzofuran with a nitro group, which must be converted to an amine for final sulfonamide formation.  The following reactions are needed: A. Nitro reduction: NO2 -> NH2 B. Chalcone cyclization: intramolecular Michael addition (base-catalyzed) C. SNAr: piperazine displaces Br D. Sulfonamide formation: NH2 + sulfonyl chloride E. Wolff-Kishner: reduce ketone to methylene F. Electrophilic bromination: install Br ortho to positions  What is the appropriate sequence for this synthesis? A. B into E into A into F into C into D  B. B into F into E into C into A into D  C. B into E into F into A into C into D  D. B into E into F into C into A into D & D \\
\addlinespace
Spectroscopic structure elucidation & In the context of a lab environment, in which an agent is trying to solve the task of how many aromatic carbons are in the molecule of the experimental setup, the following is an observation from an agent-environment interaction: Observation: \{'tool\_name': 'carbon\_nmr\_spectra', 'arguments': \{\}, 'result': 'Deltas: 135.80, 117.40, 70.50, 54.00, 43.70', 'status': 'success', 'error\_message': None, 'duration': 0.6971059627830982, 'timestamp': '2026-01-06T13:43:01.986330+00:00'\}  Below are a few scenarios that contain some reasoning statements.    Which option corresponds to a valid reasoning? A. Perfect! Now I can analyze the carbon shifts. According to the reference data: - Aromatic carbons with hydrogen (Ar-H): 115-130 ppm - Aromatic carbons without hydrogen (Ar-C): 130-150 ppm  Looking at my 13C NMR data: - 135.80 ppm: This falls in the aromatic carbon range (130-150...  B. Perfect! Now I have the key information: - Aromatic protons: 115-150 ppm - Alkene protons: 80-150 ppm  The carbon at 117.4 ppm with two protons at \textasciitilde{}5 ppm is clearly a vinyl =CH2 group, not aromatic.  The carbon at 135.8 ppm with a proton at 6.05 ppm is more ambiguous. The proton...  C. Perfect! Now I can analyze the carbon shifts. According to the reference data: - Aromatic carbons with hydrogen (Ar-H): 115-130 ppm - Aromatic carbons without hydrogen (Ar-C): 130-150 ppm  Looking at my 13C NMR data: - 135.80 ppm: This falls in the aromatic carbon range (130-150...  D. Perfect! Now I can analyze the carbon shifts. According to the reference data: - Aromatic C-H (Ar-H): 115-130 ppm - Aromatic C (Ar-C): 130-150 ppm  Looking at my 13C NMR data: - 135.80 ppm: This falls in the aromatic carbon range (130-150 ppm for Ar-C) - 117.40 ppm: This falls... & B \\
\addlinespace
Inorganic qualitative analysis & The following observations were collected during experimentation on a black inorganic salt: - It dissolves in nitric acid, resulting in a green solution (solution X). - Addition of DMG to solution X causes a red precipitate to form.  Assuming that all of the above statements are true, which of the following options is correct? A. a black precipitate will form when Pb(NO3)2 is added to solution X   B. a green precipitate will form when NH3 is added to solution X  C. the original black salt will not dissolve in sulfuric acid  D. the original black salt will also dissolve in concentrated NaOH & A \\
\addlinespace
\bottomrule
\end{tabularx}
    }
\end{table}

\paragraph{Per-model scores.} \Cref{tab:qa-scores} presents the scores for each model on knowledge and reasoning questions. \claudesonnet achieves the highest scores in most environments, winning or tying on 13 of 16 environment–question-type pairs. \gptfourO performs competitively on several environments (e.g., AFM experiment execution and adsorption surface construction knowledge questions, where all models tie at 0.88–0.90), but lags substantially behind on spectroscopic structure elucidation (0.46 vs. 0.81 for \claudesonnet) and inorganic qualitative analysis (0.25 vs. 0.70). \gptoss shows its strongest relative performance in molecular simulation, where it attains the top knowledge score (0.95) and ties for the best reasoning score (0.95). Reasoning questions are generally more difficult. Spectroscopic structure elucidation is the hardest domain for every model: \claudesonnet, \gptfourO, and \gptoss score 0.31, 0.14, and 0.31, respectively.

\begin{table}[H]
    \centering
    \caption{\textbf{Q\&A Scores for each of the analyzed models across the different tasks.} Bold text indicates the best scores for each of the topics. \textit{Knowledge questions} evaluate the factual knowledge needed to solve the agentic tasks, while the \textit{Reasoning Questions} aim to simulate the reasoning-related processes that agents face during problem-solving.}
    \label{tab:qa-scores}
    \resizebox{\textwidth}{!}{
      \begin{tabular}{l c S[table-format=1.2] S[table-format=1.2] S[table-format=1.2]}
    \toprule
    Environment & \# Questions & {Claude-4.5-Sonnet} & {GPT-4o} & {GPT-OSS-120B} \\
    \midrule
    \multicolumn{5}{l}{\textit{Knowledge questions}} \\
    \midrule
    AFM experiment execution & 50 & {\textbf{0.88}} & {\textbf{0.88}} & {\textbf{0.88}} \\
    Adsorption surface construction & 50 & {\textbf{0.88}} & 0.86 & 0.84 \\
    Molecular simulation & 20 & 0.90 & 0.90 & {\textbf{0.95}} \\
    ML-based property prediction & 50 & {\textbf{0.74}} & {\textbf{0.74}} & 0.66 \\
    Circuit inference & 31 & {\textbf{0.90}} & 0.74 & 0.84 \\
    Retrosynthetic planning & 90 & {\textbf{0.84}} & 0.69 & 0.73 \\
    Spectroscopic structure elucidation & 106 & {\textbf{0.81}} & 0.46 & 0.67 \\
    Inorganic qualitative analysis & 20 & {\textbf{0.70}} & 0.25 & 0.60 \\
    \midrule
    \multicolumn{5}{l}{\textit{Reasoning questions}} \\
    \midrule
    AFM experiment execution & 20 & {\textbf{0.80}} & 0.70 & 0.65 \\
    Adsorption surface construction & 21 & {\textbf{0.90}} & {\textbf{0.90}} & 0.86 \\
    Molecular simulation & 20 & {\textbf{0.95}} & {\textbf{0.95}} & {\textbf{0.95}} \\
    ML-based property prediction & 25 & {\textbf{0.88}} & 0.76 & 0.80 \\
    Circuit inference & 20 & {\textbf{0.70}} & 0.40 & 0.45 \\
    Retrosynthetic planning & 20 & {\textbf{0.90}} & 0.85 & 0.85 \\
    Spectroscopic structure elucidation & 29 & {\textbf{0.31}} & 0.14 & {\textbf{0.31}} \\
    Inorganic qualitative analysis & 20 & {\textbf{0.70}} & 0.30 & 0.60 \\
    \bottomrule
  \end{tabular}

    }
\end{table}

\section{Marker manual annotation} \label{app:manual_annotation_description}

\begin{figure}
    \centering
    \includegraphics[width=1\linewidth]{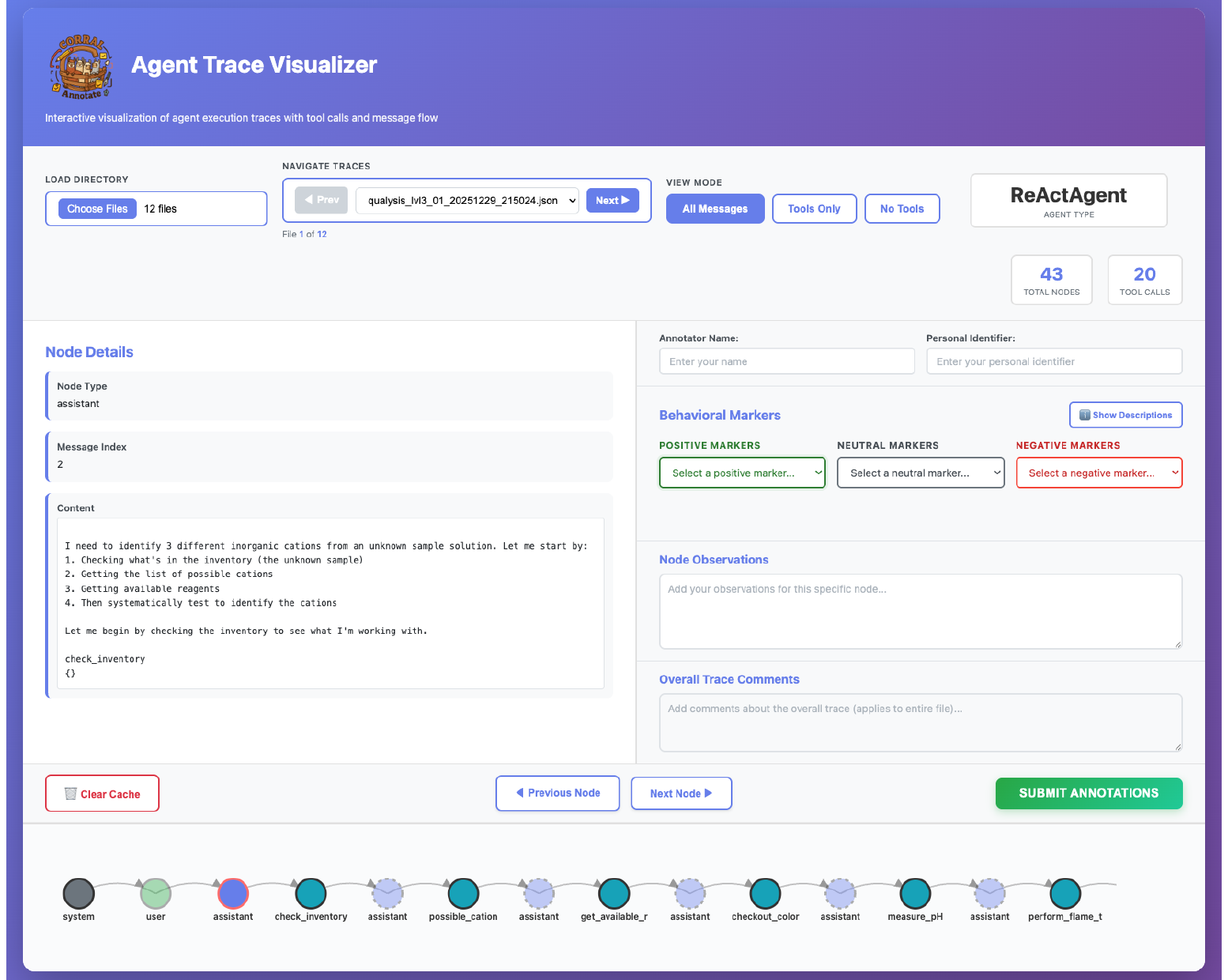}
    \caption{\textbf{Screenshot of the Agent Trace Visualizer annotation interface.} The tool renders each execution trace as a navigable sequence of steps (model outputs, tool calls, and environment responses), allowing the expert annotator to assign one or more behavioral markers from a controlled taxonomy and optionally add free-text observations at the step level and an overall trace-level comment.}
    \label{fig:marker_annotation_app}
\end{figure}

Domain experts annotated agent execution traces using a custom web-based tool (see \Cref{fig:marker_annotation_app}) that renders each trace as a navigable sequence of steps: model outputs, tool calls, and environment responses. One expert was assigned to each scientific environment. Each expert reviewed all accepted traces within their assigned environment, covering both the \gptfourO and \claudesonnet models, across combinations with ReAct and tool‑calling agents for the different scopes.

For each step in a trace, the annotator assigned one or more behavioral markers drawn from a controlled taxonomy (\Cref{tab:markers}). Annotators could supplement marker selections with free-text observations at the step level and an overall comment characterizing the trace as a whole.

As a result, 773 traces were annotated end-to-end following this procedure. The detailed results of the marker counts are described in \Cref{app:marker_annotation_results}.

\begin{table}[H]
\centering
\caption{\textbf{Behavioral marker taxonomy used during expert annotation of agent execution traces.} Markers are organized into three tiers: \textit{positive} markers capture effective reasoning and task execution; \textit{neutral} markers flag unremarkable steps; and \textit{negative} markers identify failure modes ranging from hallucination and looping to premature answer submission.}
\label{tab:markers}
\small
\begin{tabular}{llp{8cm}}
\toprule
\textbf{Category} & \textbf{Marker} & \textbf{Definition} \\
\midrule
\textit{Positive} & & \\
\midrule
         & \texttt{validation\_attempt}    & The agent explicitly validates a result or intermediate output. \\
         & \texttt{backtrack\_trigger}     & The agent recognizes a dead end and pivots to an alternative approach. \\
         & \texttt{planning\_statement}    & The agent articulates an explicit plan or subgoal before acting. \\
         & \texttt{reasoning\_statement}   & The agent produces explicit reasoning connecting hypotheses to evidence. \\
         & \texttt{correct\_submission}    & The agent submits a final answer in the required format. \\
         & \texttt{todo\_list}             & The agent consults a structured task list. \\
\midrule
\textit{Neutral} & & \\
\midrule
         & \texttt{neutral}                & The step contains no remarkable activity. \\
\midrule
\textit{Negative} \\
\midrule
         & \texttt{missing\_validation}    & A validation step was warranted but omitted. \\
         & \texttt{unnecessary\_tool\_use} & A tool was invoked when not needed. \\
         & \texttt{non\_sense}             & The output is incoherent or logically inconsistent. \\
         & \texttt{loop\_instance}         & The agent repeats tool calls without incorporating new information. \\
         & \texttt{hallucination}          & The agent asserts fabricated or unjustified content. \\
         & \texttt{wrong\_planning}        & A plan is present but is factually or logically incorrect. \\
         & \texttt{wrong\_reasoning}       & Reasoning is present but leads to an incorrect conclusion. \\
         & \texttt{syntax\_error}          & The agent produces syntactically malformed output. \\
         & \texttt{early\_final\_answer}   & A final answer is submitted without sufficient justification. \\
         & \texttt{give\_up}               & The agent explicitly states it cannot solve the task. \\
         & \texttt{inefficient\_tool\_call}& The correct tool is invoked with a vague or suboptimal query. \\
         & \texttt{iteration\_limit}       & The agent reached the maximum allowed number of iterations. \\
\bottomrule
\end{tabular}
\end{table}

\section{Performance analysis}
\label{app:performance}

The base model is the primary determinant of agent performance. Scaffold, tool-documentation verbosity, and item-level factors account for a smaller share of the variance (\Cref{fig:performance}). Aggregate scores (\Cref{tab:aggregate-scores,tab:aggregate-scores-subtask}) and per-skill performance (\Cref{fig:2c_task_category}) are examined in the following subsections. A hierarchical Bayesian model of per-trial outcomes supplies the variance decomposition and a task-level calibration check (\Cref{app:binomial-subsec}). A two-parameter IRT analysis of curated question-answer items separates knowledge from reasoning ability for each model--environment pair (\Cref{sec:app_irt}). Intervention experiments on the conversation history (\Cref{app:interventions}) and ablations of tool-documentation verbosity (\Cref{app:verbosity}) complement these analyses.

\subsection{Aggregate scores} \label{app:aggregate-scores}

\Cref{tab:aggregate-scores} and \Cref{tab:aggregate-scores-subtask} present mean scores across all eight environments for each model–agent combination. \Cref{tab:aggregate-scores} reports task-level outcomes, where each cell is the mean over complete end-to-end trial episodes within the corresponding environment–scope bin. \Cref{tab:aggregate-scores-subtask} provides the analogous results at the subtask level; the assignments are divided over the individual constituent steps rather than over full trials, allowing fine-grained evaluation. Consequently, subtask scores are generally higher than task scores, because partial credit accumulates across a trial even when the overall task is not solved.

The ranking of models is largely consistent across both tables: \claudesonnet leads in most environments, followed by \gptfourO and \gptoss, although the performance gap narrows for lower‑complexity scopes (S1) and for environments that require tight multi‑step planning (e.g., Retrosynthesis). ReAct and Tool‑Calling agents perform comparably on average, with differences becoming more pronounced at higher scope levels and in workflow‑construction environments (e.g., \textit{AFM Experiment}, \textit{Molecular Simulation}), where the structured action space of Tool Calling confers a modest advantage.

\begin{table}[H]
    \centering
    \caption{\textbf{Average scores by environment and scope for each model and agent type.} R=ReAct, TC=Tool Calling. Environments are grouped by task category: \textit{Hypothesis-driven inquiry} (Spectra Elucidation, Circuit Inference), \textit{Strategic reasoning} (Retrosynthesis), and \textit{Workflow construction} (AFM Experiment, Molecular Simulation, Adsorption Surface, ML Property). Scope levels (S1–S4) reflect increasing task difficulty within each environment.}
    \label{tab:aggregate-scores}

\begin{tabular}{@{}l l w{c}{2.1em} w{c}{2.1em} w{c}{2.1em} w{c}{2.1em} w{c}{2.1em} w{c}{2.1em}@{}}
\toprule
\textbf{Environment} & \textbf{Scope} & \multicolumn{2}{c}{\textbf{Claude-4.5-Sonnet}} & \multicolumn{2}{c}{\textbf{GPT-4o}} & \multicolumn{2}{c}{\textbf{GPT-OSS-120B}} \\
\cmidrule(lr){3-4} \cmidrule(lr){5-6} \cmidrule(lr){7-8}
\multicolumn{2}{l}{} & R & TC & R & TC & R & TC \\
\midrule
\multicolumn{8}{l}{\textit{Hypothesis-driven inquiry}} \\
\midrule
Spectra Eluc. & S1 & \textbf{0.42} & 0.25 & 0.02 & 0.06 & 0.04 & 0.06 \\
 & S2 & \textbf{0.39} & 0.23 & 0.02 & 0.03 & 0.03 & 0.07 \\
Inorg. Qual. Analysis & S1 & \textbf{0.60} & 0.56 & 0.06 & 0.06 & 0.16 & 0.16 \\
 & S2 & \textbf{0.46} & 0.22 & 0.02 & 0.00 & 0.00 & 0.04 \\
 & S3 & \textbf{0.28} & 0.22 & 0.00 & 0.00 & 0.02 & 0.02 \\
Circuit Inf. & S1 & \textbf{0.73} & \textbf{0.73} & 0.00 & 0.00 & 0.20 & 0.47 \\
\midrule
\multicolumn{8}{l}{\textit{Strategic reasoning}} \\
\midrule
Retrosynthesis & S1 & 0.85 & \textbf{0.88} & 0.75 & 0.85 & 0.05 & 0.78 \\
 & S2 & \textbf{0.50} & 0.45 & 0.23 & 0.35 & 0.05 & 0.35 \\
 & S3 & \textbf{0.38} & \textbf{0.38} & 0.18 & 0.12 & 0.00 & 0.03 \\
\midrule
\multicolumn{8}{l}{\textit{Workflow construction}} \\
\midrule
AFM Exp. & S1 & \textbf{1.00} & \textbf{1.00} & 0.60 & 0.20 & 0.00 & 0.20 \\
 & S2 & \textbf{0.40} & 0.00 & 0.00 & 0.00 & 0.00 & 0.00 \\
 & S3 & \textbf{0.00} & \textbf{0.00} & \textbf{0.00} & \textbf{0.00} & \textbf{0.00} & \textbf{0.00} \\
 & S4 & 0.00 & 0.00 & 0.00 & 0.00 & 0.00 & \textbf{0.40} \\
Mol. Simulation & S1 & 0.53 & \textbf{0.80} & 0.07 & 0.13 & 0.07 & 0.60 \\
 & S2 & \textbf{0.70} & 0.60 & 0.20 & 0.10 & 0.00 & 0.40 \\
Adsorption Surf. & S1 & \textbf{1.00} & \textbf{1.00} & 0.87 & \textbf{1.00} & 0.00 & \textbf{1.00} \\
ML Property & S1 & 0.73 & \textbf{1.00} & 0.73 & 0.53 & 0.00 & 0.80 \\
\bottomrule
\end{tabular}

\end{table}

\begin{table}[H]
    \centering
    \caption{\textbf{Subtask-level average scores by environment and scope for each model and agent type.} R=ReAct, TC=Tool Calling. Each cell reports the mean over the individual subtask items belonging to that environment–scope combination, providing a finer-grained view of model capabilities.}
    \label{tab:aggregate-scores-subtask}

\begin{tabular}{@{}l l w{c}{2.1em} w{c}{2.1em} w{c}{2.1em} w{c}{2.1em} w{c}{2.1em} w{c}{2.1em}@{}}
\toprule
\textbf{Environment} & \textbf{Scope} & \multicolumn{2}{c}{\textbf{Claude-4.5-Sonnet}} & \multicolumn{2}{c}{\textbf{GPT-4o}} & \multicolumn{2}{c}{\textbf{GPT-OSS-120B}} \\
\cmidrule(lr){3-4} \cmidrule(lr){5-6} \cmidrule(lr){7-8}
\multicolumn{2}{l}{} & R & TC & R & TC & R & TC \\
\midrule
\multicolumn{8}{l}{\textit{Hypothesis-driven inquiry}} \\
\midrule
Spectra Eluc. & S1 & 0.52 & \textbf{0.54} & 0.38 & 0.40 & 0.48 & 0.50 \\
 & S2 & \textbf{0.50} & 0.49 & 0.29 & 0.32 & 0.40 & 0.47 \\
Inorg. Qual. Analysis & S1 & \textbf{0.52} & 0.50 & 0.06 & 0.24 & 0.03 & 0.22 \\
 & S2 & \textbf{0.74} & 0.63 & 0.30 & 0.27 & 0.30 & 0.31 \\
 & S3 & \textbf{0.64} & 0.51 & 0.24 & 0.24 & 0.25 & 0.29 \\
Circuit Inf. & S1 & \textbf{1.00} & 0.92 & 0.26 & 0.25 & 0.53 & 0.74 \\
\midrule
\multicolumn{8}{l}{\textit{Strategic reasoning}} \\
\midrule
Retrosynthesis & S1 & \textbf{0.96} & 0.92 & 0.69 & 0.81 & 0.52 & 0.88 \\
 & S2 & \textbf{0.48} & 0.47 & 0.23 & 0.35 & 0.05 & 0.38 \\
\midrule
\multicolumn{8}{l}{\textit{Workflow construction}} \\
\midrule
AFM Exp. & S1 & 0.71 & \textbf{1.00} & 0.83 & 0.54 & 0.00 & 0.66 \\
 & S2 & \textbf{0.66} & 0.49 & 0.40 & 0.31 & 0.00 & 0.37 \\
 & S3 & 0.33 & \textbf{0.40} & 0.35 & 0.22 & 0.00 & 0.18 \\
 & S4 & 0.00 & 0.00 & \textbf{0.27} & 0.00 & 0.00 & 0.20 \\
Mol. Simulation & S1 & 0.77 & \textbf{0.97} & 0.66 & 0.59 & 0.16 & 0.72 \\
Adsorption Surf. & S1 & \textbf{1.00} & 0.96 & 0.53 & 0.86 & 0.27 & 0.98 \\
ML Property & S1 & 0.93 & 0.97 & 0.40 & 0.97 & 0.55 & \textbf{0.98} \\
\bottomrule
\end{tabular}

\end{table}

\subsection{Subtask-category gradient}

\begin{figure}
    \centering
    \includegraphics[width=\linewidth]{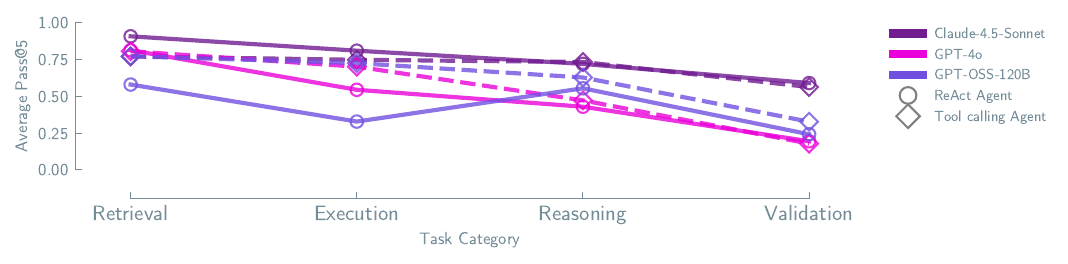}
    \caption{\textbf{Performance decreases as epistemic demand increases.} All tasks were broken down into subtasks, each classified by experts into four categories (retrieval, execution, reasoning, validation). Both scaffolds for all models show the same trend.}
    \label{fig:2c_task_category}
\end{figure}

\Cref{fig:2c_task_category} presents average \passat{} broken down by the four skill categories (retrieval, execution, reasoning, and validation) across all model--agent combinations. Across most combinations, retrieval and execution scores are higher than reasoning and validation scores, and the spread across combinations is larger for the latter two categories. Reasoning and validation require logical inference, consistency checking, and hypothesis testing, whereas retrieval and execution test more routine information access and action selection. The pattern aligns with the gradient in epistemic demand reported in the main text.

\subsection{Manual marker annotation} \label{app:marker_annotation_results}

\Cref{tab:marker-count} presents the total occurrence counts of each behavioral marker across all annotated traces, broken down by model and scaffold. Positive markers are dominated by \texttt{validation\_attempt} and \texttt{reasoning\_statement}, with \claudesonnet producing substantially more explicit reasoning steps than \gptfourO across both scaffolds (e.g., 392 vs. 69 under ReAct). ReAct consistently elicits more \texttt{planning\_statement} annotations for \claudesonnet (240 vs. 34 for tool calling), suggesting that the scaffold shapes high-level planning behavior more for that model than for \gptfourO. Among negative markers, \texttt{loop\_instance} and \texttt{wrong\_reasoning} are the most prevalent across all conditions. \gptfourO shows notably higher rates of \texttt{non\_sense} and \texttt{syntax\_error} under ReAct (141 and 100, respectively), pointing to formatting difficulties with that scaffold. \texttt{misunderstood\_tool} is almost exclusively observed in the \claudesonnet tool-calling condition (31 out of 38 occurrences), indicating a scaffold-specific failure mode.

\begin{table}[H]
\centering
\caption{\textbf{Behavioral marker counts per model and scaffold.} Total number of times each marker was observed across all annotated traces, grouped by sentiment category (Positive, Neutral, Negative). Markers are defined in \Cref{tab:markers}.}
\label{tab:marker-count}
\small
\begin{tabular}{lcccc}
\toprule
\textbf{Marker} & \multicolumn{2}{c}{\textbf{Claude-4.5-Sonnet}} & \multicolumn{2}{c}{\textbf{GPT-4o}} \\
\cmidrule(lr){2-3}\cmidrule(lr){4-5}
 & \textbf{ReAct} & \textbf{Tool calling} & \textbf{ReAct} & \textbf{Tool calling} \\
\midrule
\multicolumn{5}{l}{\textit{Positive}} \\
\midrule
\texttt{validation\_attempt} & 413 & 378 & 243 & 399 \\
\texttt{backtrack\_trigger} & 77 & 44 & 27 & 28 \\
\texttt{planning\_statement} & 240 & 34 & 33 & 61 \\
\texttt{reasoning\_statement} & 392 & 290 & 69 & 49 \\
\texttt{correct\_submission} & 128 & 137 & 30 & 33 \\
\texttt{todo\_list} & 25 & 29 & 0 & 0 \\
\midrule
\multicolumn{5}{l}{\textit{Neutral}} \\
\midrule
\texttt{neutral} & 1120 & 1287 & 1012 & 1056 \\
\midrule
\multicolumn{5}{l}{\textit{Negative}} \\
\midrule
\texttt{missing\_validation} & 61 & 74 & 60 & 81 \\
\texttt{unnecessary\_tool\_use} & 96 & 106 & 25 & 81 \\
\texttt{non\_sense} & 9 & 15 & 141 & 63 \\
\texttt{loop\_instance} & 255 & 401 & 390 & 419 \\
\texttt{hallucination} & 53 & 48 & 139 & 28 \\
\texttt{wrong\_planning} & 43 & 92 & 28 & 48 \\
\texttt{wrong\_reasoning} & 319 & 332 & 184 & 138 \\
\texttt{syntax\_error} & 8 & 17 & 100 & 20 \\
\texttt{early\_final\_answer} & 19 & 26 & 57 & 76 \\
\texttt{give\_up} & 8 & 25 & 32 & 22 \\
\texttt{inefficient\_tool\_call} & 28 & 28 & 52 & 29 \\
\texttt{iteration\_limit} & 20 & 20 & 24 & 30 \\
\texttt{misunderstood\_tool} & 3 & 31 & 0 & 4 \\
\bottomrule
\end{tabular}
\end{table}

\subsection{Latent factor modelling}
\label{sec:app_irt}

\begin{figure}
    \centering
    \includegraphics[width=1\linewidth]{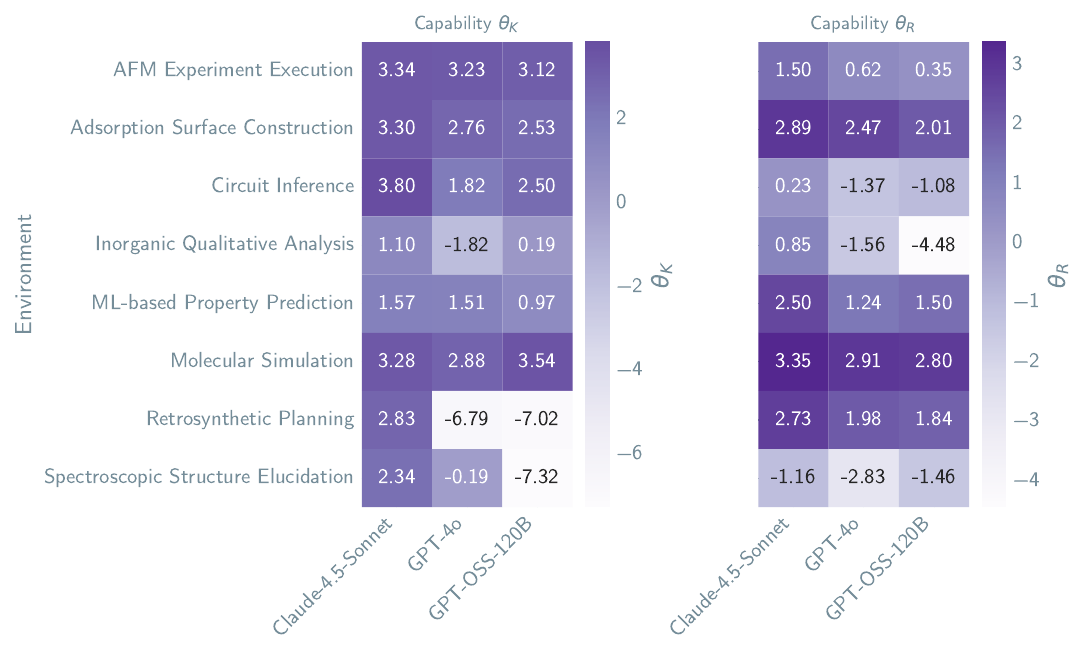}
    \caption{\textbf{Model capabilities vary sharply across scientific domains, with knowledge and reasoning deficits concentrated in the hypothesis-driven inquiry domain.}
      Side-by-side heatmaps of the latent knowledge ($\theta_K$, left) and reasoning ($\theta_R$, right) capabilities estimated from a two-parameter IRT model fit to task outcomes
  across all agent configurations. Rows correspond to the eight benchmark environments; columns show the three evaluated models. Cell values are posterior means of the per-model,
  per-environment capability parameters, with color intensity scaled from low (light) to high (dark purple), centered at zero. Both heatmaps share a common $y$-axis for direct comparison.}
      \label{fig:capability_heatmaps}
\end{figure}

\begin{figure}[H]
    \centering
    \includegraphics[width=0.5\linewidth]{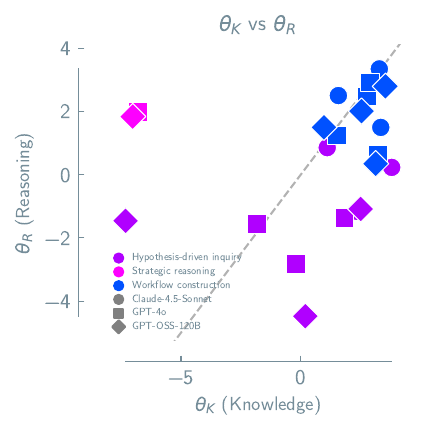}
    \caption{\textbf{Knowledge and reasoning capabilities are broadly correlated, but retrosynthesis demands reasoning far beyond what domain knowledge alone predicts.}
      Scatter plot of latent knowledge capability ($\theta_K$) against reasoning capability ($\theta_R$) for each model--environment pair, estimated from a two-parameter IRT model. Marker
  shape encodes the model (circle: \claudesonnetfull; square: \gptfourOfull; diamond: \gptossfull) and color encodes the domain group (blue: workflow construction; purple: hypothesis-driven
  inquiry; magenta: strategic reasoning). The dashed diagonal marks $\theta_K = \theta_R$; points above the line indicate environments where reasoning demands exceed knowledge demands.
  Retrosynthetic planning (magenta) sits far above the diagonal with strongly negative $\theta_K$, suggesting models rely on chain-of-thought reasoning rather than retrievable domain
  knowledge to navigate the combinatorial search space.}
      \label{fig:knowledge_vs_reasoning}
\end{figure}

\paragraph{IRT capability estimation.}
  We estimate latent knowledge and reasoning capabilities for each
  model-environment pair using a separate two-parameter logistic (2PL)
  item response theory (IRT) models \autocite{hori2022item}. For each
  of the eight benchmark environments, we prepared two sets of
  question-answer (QA) items: one targeting domain-specific factual
  knowledge and one targeting multi-step reasoning (\Cref{tab:qa-scores}; item generation and review protocol in \Cref{app:capability-items}). All three models
  were evaluated on the full QA across all environments,
  yielding binary correctness outcomes. We fit one 2PL IRT model to the
  knowledge QA responses and a second, independent 2PL IRT model to the
  reasoning QA responses. In each model, a latent ability parameter
  $\theta_{m,e}$ is estimated for every model--environment pair $(m,e)$
  via a hierarchical prior $\theta_{m,e} \sim
  \mathcal{N}(\mu_m + \nu_e,\, \sigma_\theta)$, where $\mu_m$ and
  $\nu_e$ are the sum-to-zero model and environment effects. Item
  parameters follow the standard 2PL parameterization: discrimination
  $a_i = \exp(\log a_i)$ with $\log a_i \sim \mathcal{N}(0, 0.5)$ and
  difficulty $b_i \sim \mathcal{N}(0, 2)$, giving item response
  probability $P(\text{correct}) = \mathrm{logistic}(a_i \theta - b_i)$.
  All parameters are estimated via MCMC (4 chains, 2000 post-warmup
  draws each, target acceptance 0.9); we report posterior means and
  standard deviations of $\theta$. The knowledge model yields
  $\theta_K$ and the reasoning model yield $\theta_R$, enabling a
  direct comparison of where each model's knowledge versus reasoning
  capabilities diverge across scientific domains. The extracted capabilities are plotted as a heatmap in \Cref{fig:capability_heatmaps}; a complementary scatter view comparing knowledge against reasoning abilities per model--environment pair is shown in \Cref{fig:knowledge_vs_reasoning}.

\subsection{Binomial modeling}
\label{app:binomial-subsec}

\begin{figure}[ht]
      \centering
      \includegraphics[width=\textwidth]{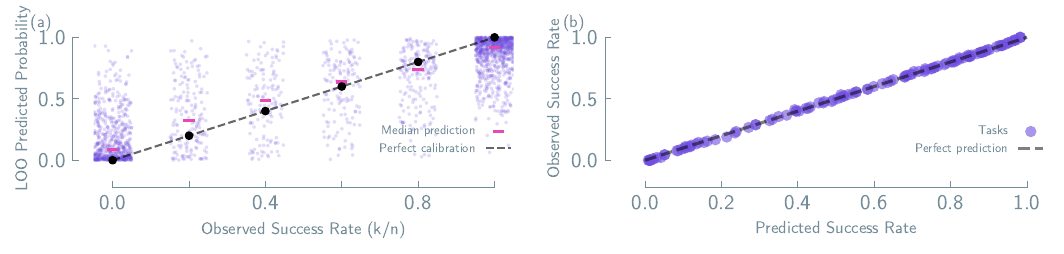}
      \caption{\textbf{The best-fitting latent factor model is well calibrated
      and accurately recovers task-level success rates.}
      \textbf{(a)}~\textit{Leave-one-out (LOO) predicted probabilities grouped by
      observed success rate ($k/n$)}. Each point is one observation, jittered
      horizontally for visibility; pink bars mark group medians. The dashed
      diagonal indicates perfect calibration. Median predictions closely
      track the diagonal across the full range, confirming that the model
      does not systematically over- or under-predict at any difficulty level.
      \textbf{(b)}~\textit{Task-averaged LOO predicted success rate versus observed
      success rate}. Each point represents one task, averaged over all
      model--scaffold--verbosity configurations. Points cluster tightly
      around the identity line, indicating that the model captures
      task-level variation rather than merely fitting marginal statistics.
      Both panels use M7 (abilities $\times$ environment $\times$ scope),
      the top-ranked model on the ELPD-LOO frontier.}
      \label{fig:lfm_fitting}
  \end{figure}

The goal of the latent factor modeling stage is to quantify how much of the variance in agent success is attributable to model capability (knowledge and reasoning), how much to agent-design choices (scaffold, verbosity). This decomposition is not uniquely determined since it depends on modeling assumptions about which factors interact and at what granularity grouping structure is imposed. For instance, task difficulty might be adequately captured at the environment level, or it might require finer environment×scope resolution. To avoid committing to a single set of assumptions, we specified a family of eight candidate models (M1–M8) that span a range of structural choices. All models share the same likelihood, priors, and core covariates (\Cref{tab:irt_notation}); they differ only in how they partition residual variance. We then select among them using Pareto-smoothed importance sampling leave-one-out cross-validation \autocite{vehtari2017practical} (PSIS-LOO; \Cref{fig:elpd_vs_complexity}), which balances predictive accuracy against model complexity without requiring a held-out test set. The variance decomposition reported in the main text (\Cref{fig:irt-main-fig}C) and the posterior predictive checks in \Cref{fig:lfm_fitting} are both based on the top-ranked model on the ELPD-LOO frontier.

\subsubsection{Model specifications}
\label{sec:models_irt}

All models include the common sum-to-zero effects; they differ in how the
remaining variance is partitioned. The common notation used across models is described in \Cref{tab:irt_notation}; the full prior specification
  for the selected model (M7) is given in \Cref{tab:m7_priors}.

\begin{table}[H]
  \centering
  \caption{Notation used in the phenomenological latent factor models (M1--M8).}
  \label{tab:irt_notation}
  \small
  \begin{tabular}{@{}lll@{}}
  \toprule
  \textbf{Symbol} & \textbf{Name} & \textbf{Description} \\
  \midrule
  \multicolumn{3}{@{}l}{\textit{IRT-derived ability covariates}} \\[2pt]
  $\tilde{\theta}_K$        & Knowledge ability   & Standardised 2PL IRT estimate (zero mean, unit variance) \\
  $\tilde{\theta}_R$        & Reasoning ability    & Standardised 2PL IRT estimate (zero mean, unit variance) \\
  \midrule
  \multicolumn{3}{@{}l}{\textit{Global ability slopes}} \\[2pt]
  $\lambda$                 & Knowledge slope      & Base regression coefficient for $\tilde{\theta}_K$ \\
  $\psi$                    & Reasoning slope      & Base regression coefficient for $\tilde{\theta}_R$ \\
  \midrule
  \multicolumn{3}{@{}l}{\textit{Sum-to-zero categorical effects}} \\[2pt]
  $\gamma_s$                & Scaffold             & Agent approach (\texttt{react}, \texttt{tool\_calling}) \\
  $\delta_\ell$             & Scope                & Task difficulty scope (1--4; 4 is hardest) \\
  $\xi_v$                   & Verbosity            & Tool-description verbosity (3 different verbosity levels) \\
  $\kappa_c$                & Category             & Task category (\texttt{task} vs.\ \texttt{subtask}) \\
  \midrule
  \multicolumn{3}{@{}l}{\textit{Grouping variables}} \\[2pt]
  $e$                       & Environment          & 8 Scientific environments  \\
  $t$                       & Task UID             & Unique task identity (task\,$\times$\,scope combinations) \\
  \midrule
  \multicolumn{3}{@{}l}{\textit{Hierarchical and interaction terms}} \\[2pt]
  $\tau_t$                  & Task effect          & Random intercept per task;\; $\tau_t \sim \mathcal{N}(0,\sigma_\tau)$ \\
  $\alpha_e$ / $\alpha_{e,\ell}$ & Group intercept & Hierarchical intercept for environment (or env\,$\times$\,scope) \\
  $\theta_e$ / $\theta_{e,\ell}$ & Knowledge slope adj. & Sum-to-zero deviation from $\lambda$, by env (or env\,$\times$\,scope) \\
  $\phi_e$ / $\phi_{e,\ell}$     & Reasoning slope adj. & Sum-to-zero deviation from $\psi$, by env (or env\,$\times$\,scope) \\
  $\gamma_{(s,e)}$          & Scaffold\,$\times$\,env    & Joint scaffold--environment interaction effect \\  
  $\gamma_{(s,\ell)}$       & Scaffold\,$\times$\,scope  & Joint scaffold--scope interaction effect\\
  $\delta_{(e,\ell)}$       & Env\,$\times$\,scope       & Joint interaction effect  \\
  \midrule
  \multicolumn{3}{@{}l}{\textit{Priors}} \\[2pt]
  $\beta_0$                 & Intercept            & $\mathcal{N}(0,\,2)$ --- weakly informative on logit scale \\
  $\sigma_\tau$, $\sigma_\alpha$ & Hierarchical SDs & $\mathrm{Half}\text{-}\mathcal{N}(0.5)$ --- regularises toward homogeneity \\
  \bottomrule
  \end{tabular}
  \end{table}

\label{sec:models}

\begin{hypobox}{M1 : Basic model without interactions}
\begin{equation*}
  \eta_i
    = \beta_0
    + \lambda\,\tilde{\theta}_{K,i}
    + \psi\,\tilde{\theta}_{R,i}
    + \gamma_{s[i]}
    + \delta_{\ell[i]}
    + \xi_{v[i]}
    + \kappa_{c[i]}
    + \tau_{t[i]}
\end{equation*}
\end{hypobox}

\bigskip

  \begin{hypobox}{M2: Tasks + Environment}
  \begin{equation*}
    \eta_i
      = \beta_0
      + \lambda\,\tilde{\theta}_{K,i}
      + \psi\,\tilde{\theta}_{R,i}
      + \gamma_{s[i]}
      + \delta_{\ell[i]}
      + \xi_{v[i]}
      + \kappa_{c[i]}
      + \alpha_{e[i]}
      + \tau_{t[i]}
  \end{equation*}
  \end{hypobox}

\bigskip

\begin{hypobox}{M3: Ability Slopes $\times$ Environment}
  \begin{equation*}
    \eta_i
      = \beta_0
      + \bigl(\lambda + \theta_{e[i]}\bigr)\,\tilde{\theta}_{K,i}
      + \bigl(\psi   + \phi_{e[i]}\bigr)\,\tilde{\theta}_{R,i}
      + \gamma_{s[i]}
      + \delta_{\ell[i]}
      + \xi_{v[i]}
      + \kappa_{c[i]}
      + \alpha_{e[i]}
      + \tau_{t[i]}
  \end{equation*}
  \end{hypobox}

\bigskip

\begin{hypobox}{M4: Scaffold $\times$ Environment}
  \begin{equation*}
    \eta_i
      = \beta_0
      + \lambda\,\tilde{\theta}_{K,i}
      + \psi\,\tilde{\theta}_{R,i}
      + \gamma_{(s,e)[i]}
      + \delta_{\ell[i]}
      + \xi_{v[i]}
      + \kappa_{c[i]}
      + \tau_{t[i]}
  \end{equation*}
\end{hypobox}

\bigskip

\begin{hypobox}{M5: Scaffold $\times$ Scope Interaction}
\begin{equation*}
  \eta_i
    = \beta_0
    + \lambda\,\tilde{\theta}_{K,i}
    + \psi\,\tilde{\theta}_{R,i}
    + \xi_{v[i]}
    + \kappa_{c[i]}
    + \gamma_{(s,\ell)[i]}
    + \alpha_{e[i]}
    + \tau_{t[i]}
\end{equation*}
\end{hypobox}

\bigskip

\begin{hypobox}{M6: Environment $\times$ Scope Intercepts}
\begin{equation*}
  \eta_i
    = \beta_0
    + \lambda\,\tilde{\theta}_{K,i}
    + \psi\,\tilde{\theta}_{R,i}
    + \gamma_{s[i]}
    + \xi_{v[i]}
    + \kappa_{c[i]}
    + \delta_{(e,\ell)[i]}
    + \tau_{t[i]}
\end{equation*}
\end{hypobox}

\bigskip

\begin{hypobox}{M7: Ability Slopes $\times$ Environment--Scope (Best Model)}
\begin{equation*}
  \eta_i
    = \beta_0
    + \bigl(\lambda + \theta_{(e,\ell)[i]}\bigr)\,\tilde{\theta}_{K,i}
    + \bigl(\psi   + \phi_{(e,\ell)[i]}\bigr)\,\tilde{\theta}_{R,i}
    + \gamma_{s[i]}
    + \xi_{v[i]}
    + \kappa_{c[i]}
    + \alpha_{(e,\ell)[i]}
    + \tau_{t[i]}
\end{equation*}
\end{hypobox}

\bigskip

\begin{hypobox}{M8: Hybrid --- Ability Slopes $\times$ Environment + Env--Scope Intercept}
\begin{equation*}
  \eta_i
    = \beta_0
    + \bigl(\lambda + \theta_{e[i]}\bigr)\,\tilde{\theta}_{K,i}
    + \bigl(\psi   + \phi_{e[i]}\bigr)\,\tilde{\theta}_{R,i}
    + \gamma_{s[i]}
    + \xi_{v[i]}
    + \kappa_{c[i]}
    + \alpha_{(e,\ell)[i]}
    + \tau_{t[i]}
\end{equation*}
\end{hypobox}

 \begin{table}[H]
    \centering
    \caption{Prior specification for M7.
      All priors are weakly informative on the logit scale.
      Sum-to-zero constraints are imposed via soft centering:
      $\tilde{x} = x_{\mathrm{raw}} - \bar{x}_{\mathrm{raw}}$.
      Hierarchical effects use a non-centered parameterization
      $x = \sigma \cdot x_{\mathrm{raw}}$.
      The number of groups $J$ is listed for each categorical or hierarchical term.}
    \label{tab:m7_priors}
    \small
    \begin{tabular}{@{}llrl@{}}
    \toprule
    \textbf{Parameter} & \textbf{Prior} & $J$ & \textbf{Constraint / role} \\
    \midrule
    \multicolumn{4}{@{}l}{\textit{Global terms}} \\[2pt]
    $\beta_0$ (intercept)
        & $\mathcal{N}(0,\,2)$ & 1 & Baseline log-odds \\
    $\lambda$ (knowledge slope)
        & $\mathcal{N}(0,\,1)$ & 1 & Base effect of $\tilde{\theta}_K$ \\
    $\psi$ (reasoning slope)
        & $\mathcal{N}(0,\,1)$ & 1 & Base effect of $\tilde{\theta}_R$ \\
    \midrule
    \multicolumn{4}{@{}l}{\textit{Ability slope adjustments (env\,$\times$\,scope)}} \\[2pt]
    $\theta_{(e,\ell)}^{\mathrm{raw}}$ (knowledge)
        & $\mathcal{N}(0,\,0.5)$ & 16 & Sum-to-zero centered \\
    $\phi_{(e,\ell)}^{\mathrm{raw}}$ (reasoning)
        & $\mathcal{N}(0,\,0.5)$ & 16 & Sum-to-zero centered \\
    \midrule
    \multicolumn{4}{@{}l}{\textit{Sum-to-zero categorical effects}} \\[2pt]
    $\gamma_s^{\mathrm{raw}}$ (scaffold)
        & $\mathcal{N}(0,\,1)$ & 2 & Sum-to-zero centered \\
    $\kappa_c^{\mathrm{raw}}$ (category)
        & $\mathcal{N}(0,\,1)$ & 2 & Sum-to-zero centered \\
    $\xi_v^{\mathrm{raw}}$ (verbosity)
        & $\mathcal{N}(0,\,1)$ & 3 & Sum-to-zero centered \\
    \midrule
    \multicolumn{4}{@{}l}{\textit{Hierarchical random effects (non-centred)}} \\[2pt]
    $\sigma_\alpha$ (env\,$\times$\,scope SD)
        & $\mathrm{Half}\text{-}\mathcal{N}(0.5)$ & --- & Scale for $\alpha_{(e,\ell)}$ \\
    $\alpha_{(e,\ell)}^{\mathrm{raw}}$
        & $\mathcal{N}(0,\,1)$ & 16 & $\alpha = \sigma_\alpha \cdot \alpha^{\mathrm{raw}}$ \\
    $\sigma_\tau$ (task SD)
        & $\mathrm{Half}\text{-}\mathcal{N}(0.5)$ & --- & Scale for $\tau_t$ \\
    $\tau_t^{\mathrm{raw}}$
        & $\mathcal{N}(0,\,1)$ & 120 & $\tau = \sigma_\tau \cdot \tau^{\mathrm{raw}}$ \\
    \midrule
    \multicolumn{4}{@{}l}{\textit{Likelihood}} \\[2pt]
    $y_i \mid n_i, p_i$
        & $\mathrm{Binomial}(n_i,\,p_i)$ & --- & $\mathrm{logit}(p_i) = \eta_i$ \\
    \bottomrule
    \end{tabular}
  \end{table}

  \paragraph{Posterior predictive assessment.}
  We evaluate the predictive fidelity of the best-fitting latent factor
  model (M7, see \Cref{sec:models_irt}) using Pareto-smoothed importance sampling (PSIS) to
  approximate genuine leave-one-out predictions without refitting. In
  panel~(a), LOO-predicted probabilities are grouped by the empirical
  success rate of each observation; well-calibrated predictions fall on
  the diagonal, meaning the model assigns $\approx$60\% probability to
  observations that succeed 60\% of the time, and so on. Panel~(b)
  aggregates predictions to the task level: for each unique task, we average the LOO-predicted and observed success rates across all
  agent configurations and plot one against the other. The tight
  clustering around the identity line ($R^2 > 0.95$) confirms that the hierarchical model generalizes beyond in-sample fit, accurately recovering per-task difficulty even after accounting for model abilities, scaffold choice, and tool-description verbosity.

\begin{figure}
    \centering
    \includegraphics[width=0.5\linewidth]{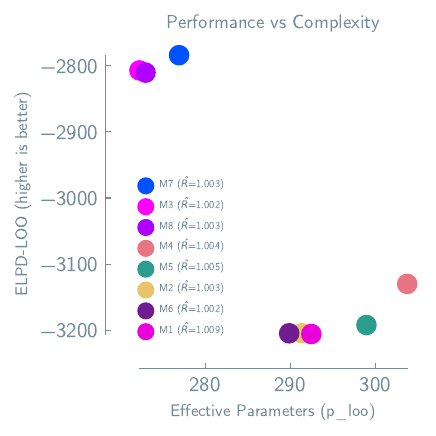}
    \caption{\textbf{Ability-based models achieve better predictive accuracy with fewer effective parameters.}
      Scatter plot of leave-one-out expected log pointwise predictive density
      (ELPD-LOO) against the effective number of parameters ($p_\mathrm{loo}$)
      for eight candidate latent factor models (M1--M8). Each marker represents
      one model specification; color distinguishes models, and the legend reports
      the maximum $\hat{R}$ across all parameters as a convergence diagnostic.
      Models in the upper-left corner are preferred: they explain held-out data
      better (higher ELPD-LOO) while using fewer effective parameters. M7
      (abilities + environment + scope) and M3 (abilities + environment) dominate
      the remaining specifications, indicating that explicit knowledge and
      reasoning ability terms capture the dominant structure in agent performance. M7 is used in our main text.
      Only models that converged ($\hat{R} < 1.01$, zero divergences) are included.}
      \label{fig:elpd_vs_complexity}
\end{figure}

\paragraph{Model comparison metrics.}
We compare candidate models using Pareto-smoothed importance sampling
leave-one-out cross-validation (PSIS-LOO) \autocite{vehtari2017practical}.
The expected log pointwise predictive density (ELPD-LOO) estimates each model's out-of-sample predictive accuracy: higher values indicate better generalization. The effective number of parameters ($p_\mathrm{loo}$) measures model complexity as the difference between the in-sample and leave-one-out log predictive densities, providing a bias-corrected analog of classical parameter counts that accounts for shrinkage in hierarchical models. Together, ELPD-LOO and $p_\mathrm{loo}$ form a predictive-accuracy--complexity trade-off analogous to the information-criterion frontier: models in the upper-left of a scatter plot achieve better prediction with less overfitting.
Convergence is assessed via the split-$\hat{R}$ statistic, where
$\hat{R} < 1.01$ across all parameters indicates reliable posterior
sampling. \Cref{fig:elpd_vs_complexity} compares complexity against performance. All modeling is implemented in PyMC~v5 \autocite{salvatier2016probabilistic} with inference via the NUTS
sampler, and post-processing uses ArviZ \cite{kumar2019arviz} for diagnostics and model
comparison.

\begin{figure}[ht]
      \centering
      \includegraphics[width=\textwidth]{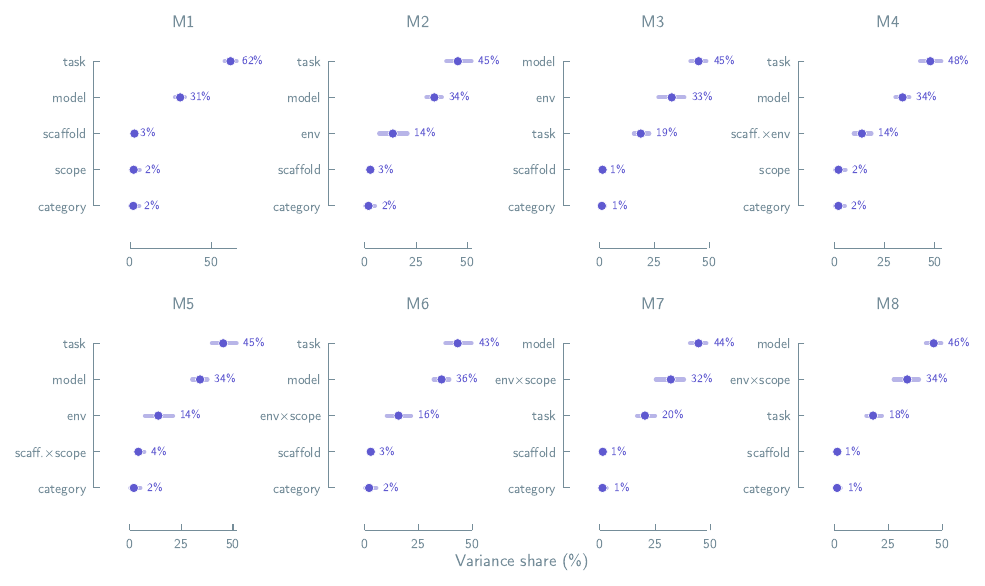}
      \caption{\textbf{Model identity dominates the variance budget regardless of
        model specification.}
        Forest plots of the top-five posterior variance-share components for
        each of the eight latent factor models (M1--M8).  Dots show the median
        and bars the 90\% credible interval over 8\,000 MCMC draws.
        In simpler specifications (M1, M2, M4, M5) task random effects absorb
        the largest share (45--62\%) because environment- and scope-level
        structure is not modeled explicitly.  Once ability slopes are allowed
        to interact with environment (M3, M7, M8), model identity rises to the
        dominant component (44--46\%), while task effects shrink correspondingly.
        Scaffold, verbosity, and category contribute $<$4\% across all eight
        specifications.}
      \label{fig:variance_forest_all}
  \end{figure}

  \paragraph{Variance decomposition across specifications.}
  \Cref{fig:variance_forest_all} compares the posterior variance budget of all eight candidate models. Two patterns are robust to specification choice. First, the model-identity component ($\sigma_{\mathrm{model}}$, driven by the IRT-derived knowledge and reasoning abilities) is consistently the largest or second-largest contributor, ranging from 31\% in the baseline (M1) to 46\% in the hybrid (M8). Second, scaffold choice, tool-description verbosity, and task category account for less than 4\% combined in every specification. The choice of LLM thus dominates over scaffold configuration. The remaining variance is split between environment-related intercepts and task-specific effects; their relative share depends on the granularity of the grouping structure, but neither overtakes the model component once ability--environment interactions are included (M3, M7, M8).

\begin{figure}
    \centering
    \includegraphics[width=0.7\linewidth]{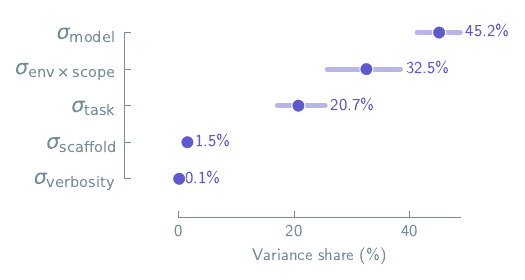}
    \caption{\textbf{The base model identity dominates the variance budget of
        the best-fitting latent factor model (M7).}
        Forest plot of posterior variance shares for the five linear-predictor
        components of M7.
        Each row shows the median (dot) and 90\% highest-density interval (bar)
        over all 8\,000 MCMC draws.
        $\sigma_{\mathrm{model}}$ (knowledge + reasoning ability, indexed by LLM
        identity) accounts for 45.2\% of the explained variance, with a
        credible interval that does not overlap any other component.
        Environment $\times$ scope intercepts and task-specific effects contribute
        most of the remaining variance (32.5\% and 20.7\%, respectively),
        while scaffold choice and tool-description verbosity are negligible
        ($<$2\% combined).}
    \label{fig:m7_variance_forest}
\end{figure}

  In the best model (see \Cref{fig:m7_variance_forest}), model identity alone explains 45.2\% [41.4, 48.8] of the variance, more than any other source and with a credible interval that does not overlap the next-largest contributor (environment\,$\times$\,scope, 32.5\%). Scaffold choice and tool-description verbosity together account for less than 2\%. Total reasoning slopes ($\psi + \phi_{(e,\ell)}$) are positive across all 16 environment\,$\times$\,scope cells (range 0.48--5.76), so higher reasoning ability predicts better task performance regardless of domain or problem scope.

\newpage
\subsection{Intervention experiments}
\label{app:interventions}

\paragraph{Protocol.} To study whether partial trajectories from prior runs can steer an agent toward recovery or failure, we implement an intervention mechanism. For every (environment, agent, task) triple, we maintain a \emph{trace registry}~$\mathcal{R}$ that maps the triple to two pools of recorded conversation traces: one drawn from successful baseline trials and one from failed baseline trials, selected from tasks with mixed outcomes (success rates between 20\% and 80\%).

  \paragraph{Trace materialisation.}
  Given an intervention type $\tau \in \{\textsc{success},
  \textsc{failed}\}$ and a step index~$k$, the hook samples a trace from
  the corresponding pool and extracts the subsequence of assistant
  turns~$\mathbf{a} = (a_1, \dots, a_N)$, where each~$a_i$ usually comprises
  thoughts \emph{and} tool invocations.  We write
  $|\mathbf{a}|_p$ for the number of assistant turns in trace~$p$.  The
  \emph{prefix slice} operator selects the turns to inject:
  \begin{equation}
    \operatorname{slice}(\mathbf{a}, k) =
    \begin{cases}
      (a_1, \dots, a_k)        & k > 0, \\[4pt]
      (a_1, \dots, a_{N+k})    & k < 0,
    \end{cases}
    \label{eq:slice}
  \end{equation}
  where positive~$k$ takes the first~$k$ steps and negative~$k$ retains
  all but the final~$|k|$.  In our experiments we use
  $k \in \{1, 2, -2, -1\}$, probing both early-stage and near-terminal
  interventions.

  The sliced steps are appended to the initial task prompt~$u_0$ to form
  the seed history~$\mathcal{H}_0$.  
  Every tool call is \emph{re-executed} against the live environment so
  those observations reflect the true current state.  The full procedure is
  given in Algorithm~\ref{alg:trace-intervention}.

  \begin{algorithm}[H]
  \caption{Trace Intervention at Step $k$}
  \label{alg:trace-intervention}
  \begin{algorithmic}[1]
  \Require Trace registry $\mathcal{R}$, intervention type $\tau$, step index $k$, task id
  \Ensure Seeded message history $\mathcal{H}_0$
  \Statex
  \State $\mathcal{P} \gets \mathcal{R}[\text{env}, \text{agent}, \text{task}].\tau\texttt{\_traces}$ \Comment{Look up trace pool}
  \State $\mathcal{P}' \gets \bigl\{\, p \in \mathcal{P} \;\big|\; (k > 0 \,\land\, |\mathbf{a}|_p \ge k) \;\lor\; (k < 0 \,\land\, |\mathbf{a}|_p > |k|) \,\bigr\}$
  \State $p^{*} \sim \mathrm{Uniform}(\mathcal{P}')$ \Comment{Sample one trace per trial}
  \State $\mathbf{a} \gets [\,m \in \textsc{Load}(p^{*}) \mid m.\texttt{role} = \texttt{"assistant"}\,]$
  \State $\mathbf{s} \gets \operatorname{slice}(\mathbf{a},\, k)$ \Comment{Eq.~\ref{eq:slice}}
  \State $\mathcal{H}_0 \gets [u_0]$ \Comment{Task prompt}
  \For{each $s_j \in \mathbf{s}$}
      \State $\textsc{Append}(\mathcal{H}_0,\; s_j)$ \Comment{Thought + action}
      \For{each action $c \in \textsc{ParseActions}(s_j)$}
          \State $o \gets \textsc{ExecuteTool}(c.\texttt{name},\; c.\texttt{args})$
          \State $\textsc{Append}(\mathcal{H}_0,\; \texttt{Observation}(o))$
      \EndFor
  \EndFor
  \Statex
  \State \Return $\mathcal{H}_0$ \Comment{Agent continues from here}
  \end{algorithmic}
  \end{algorithm}

  The agent resumes auto-regressively from~$\mathcal{H}_0$, unaware that
  the initial turns were replayed.

\begin{figure}[H]
    \centering
    \includegraphics[width=0.75\linewidth]{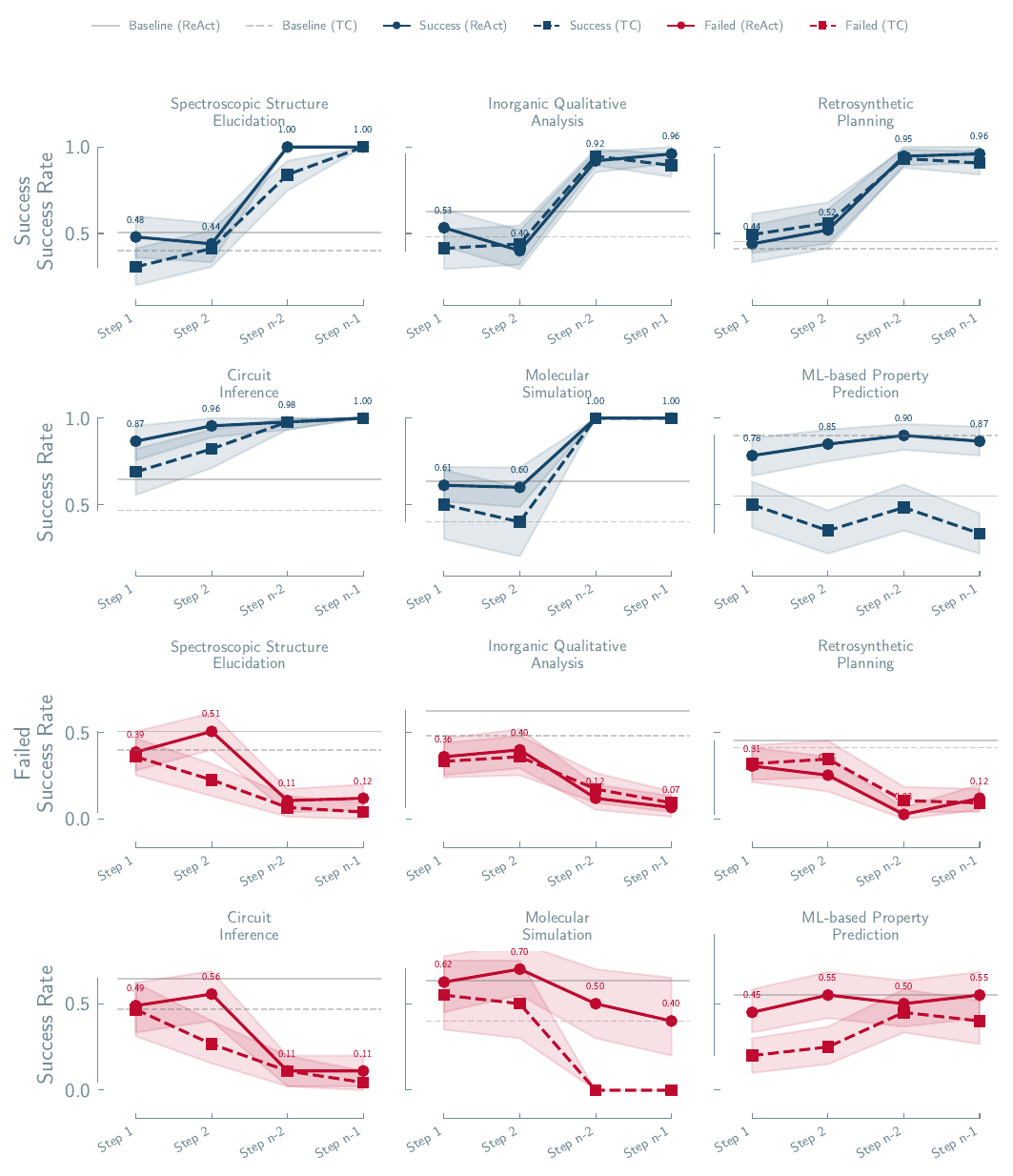}
    \caption{\textbf{Recovery curves under success and failed trace interventions across environments.}
      Top two rows show agent success rates when injecting steps from a successful trajectory;
      bottom two rows show the same for a failed trajectory.
      Environments exhibit two distinct recovery profiles.
      \emph{Hypothesis-driven inquiry} environments (\spectra, \wetlab) show minimal benefit from early successful steps---often at or below baseline---followed by a sharp jump when later steps are injected (Step~$n{-}2$, Step~$n{-}1$).
      Conversely, injecting late steps from a failed trace is catastrophic in these environments, dropping success rates to near zero (${\leq}0.12$), indicating that erroneous reasoning
  strongly anchors the agent.
      In contrast, \emph{Workflow construction} environments (\ml) exhibit monotonic, gradual improvement from early steps onward, and are more
  robust to failed traces (success rates remain ${\geq}0.40$).
      Dashed lines denote the tool-calling agent; solid lines denote ReAct.}
    \label{fig:recovery_curves_appendix}
\end{figure}

\begin{figure}[H]
    \centering
    \includegraphics[width=1\linewidth]{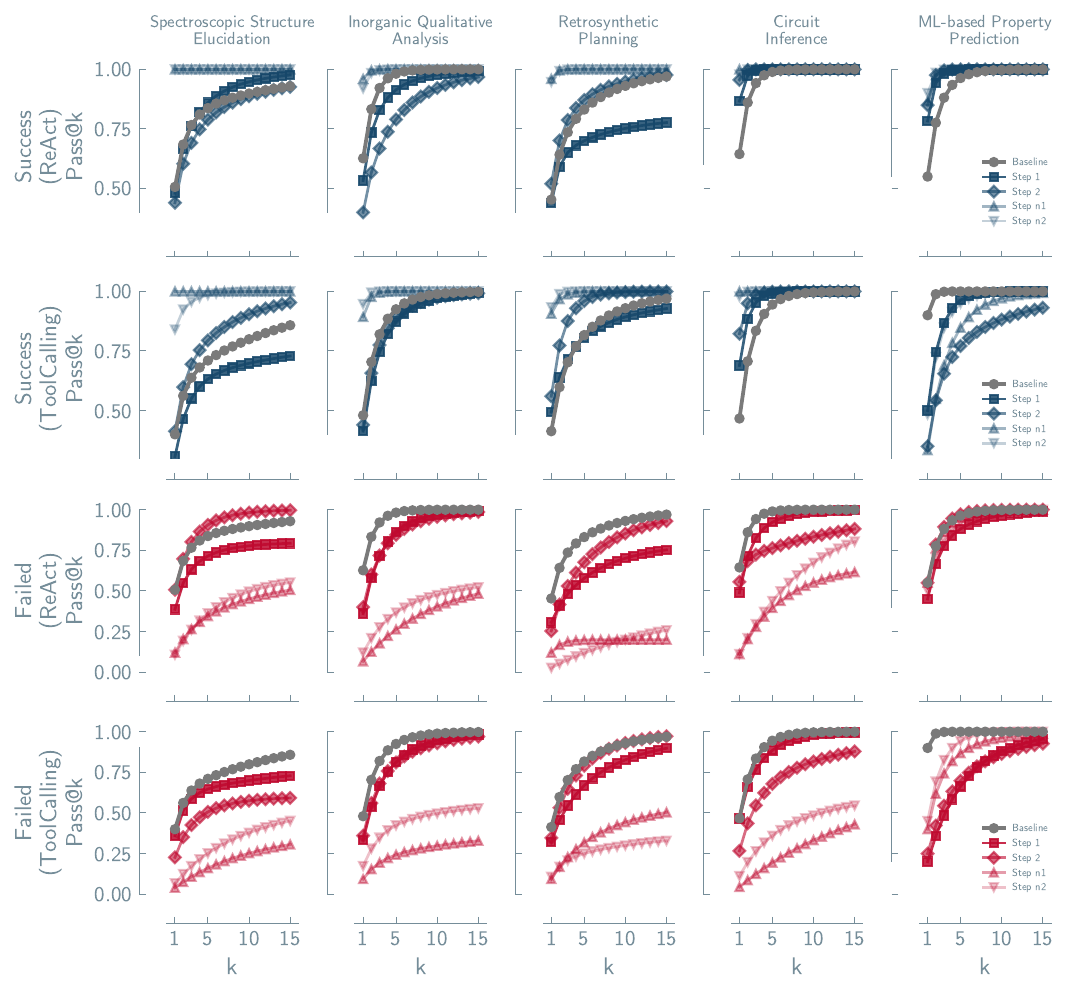}
    \caption{\textbf{\passat{} decay under trace interventions across scaffold architectures.} \passat{} as a function of $k$ under trace interventions, shown separately for ReAct and tool-calling agents.
      Rows correspond to successful interventions (top two) and failed interventions (bottom two), with columns representing each environment.
      The grey baseline curve is repeated in each panel for reference.
      Marker opacity decreases with distance from the intervention point (Step~1 darkest, Step~$n{-}2$ lightest).}
      \label{fig:pass_at_per_agent}
\end{figure}

  \paragraph{Controls.}
  Each task is run under three regimes: \emph{baseline} (no intervention), \emph{success intervention}, and \emph{failed intervention}. We run 15~trials per condition, yielding \passat{} and \passhat{} metrics for $k = 1, \dots, 15$. Failed-trace injections serve as a negative control: if injecting successful reasoning genuinely aids recovery, injecting failed reasoning should not confer the same benefit and may hinder performance. A different trace is sampled uniformly from the pool for each trial, mitigating single-trace confounds. All trials use a temperature of 0.7. Per-environment recovery curves are shown in \Cref{fig:recovery_curves_appendix}, and \passat{} decay by scaffold in \Cref{fig:pass_at_per_agent}.

\subsection{Token-level log-probability analysis}
  \label{app:logprob-results}

\begin{figure}
    \centering
    \includegraphics[width=\linewidth]{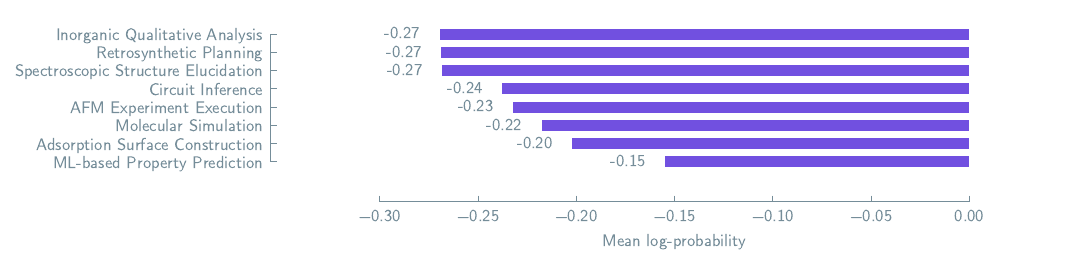}
    \caption{\textbf{Mean log probability assigned by \gptoss averaged over all tokens in completions across all traces for each environment type.} Hypothesis-driven environments yield consistently lower mean log probability than workflow-based environments.}
    \label{fig:logprob_breakdown}
\end{figure}

  We use mean token-level log-probabilities (\Cref{sec:logprob-analysis})
  as an environment-level proxy for model uncertainty.
  For each environment, we pool the log-probability of the top-1 token
  across all assistant messages and trials, excluding special control
  tokens, and report the per-environment mean $\bar{\ell}_e$
  (\Cref{fig:logprob_breakdown}).
  Workflow-construction domains (\ml,
  \catalyst, \md) exhibit the
  least negative log-probabilities ($\bar{\ell}_e \ge -0.22$),
  indicating high token-level confidence consistent with well-defined
  solution paths.
  Hypothesis-driven and strategic-reasoning domains (\wetlab,\spectra,\retro) share the most negative values
  ($\bar{\ell}_e = -0.27$).
  This ordering mirrors the gradient in epistemic demand described in
  the main text and aligns with the trace-intervention finding that
  near-complete trajectories are required to improve performance in
  exactly those environments where token-level confidence is lowest
  (\Cref{fig:combined}A).

\subsection{Verbosity detailed results}
\label{app:verbosity}

\begin{figure}[H]
    \centering
    \includegraphics[width=1\linewidth]{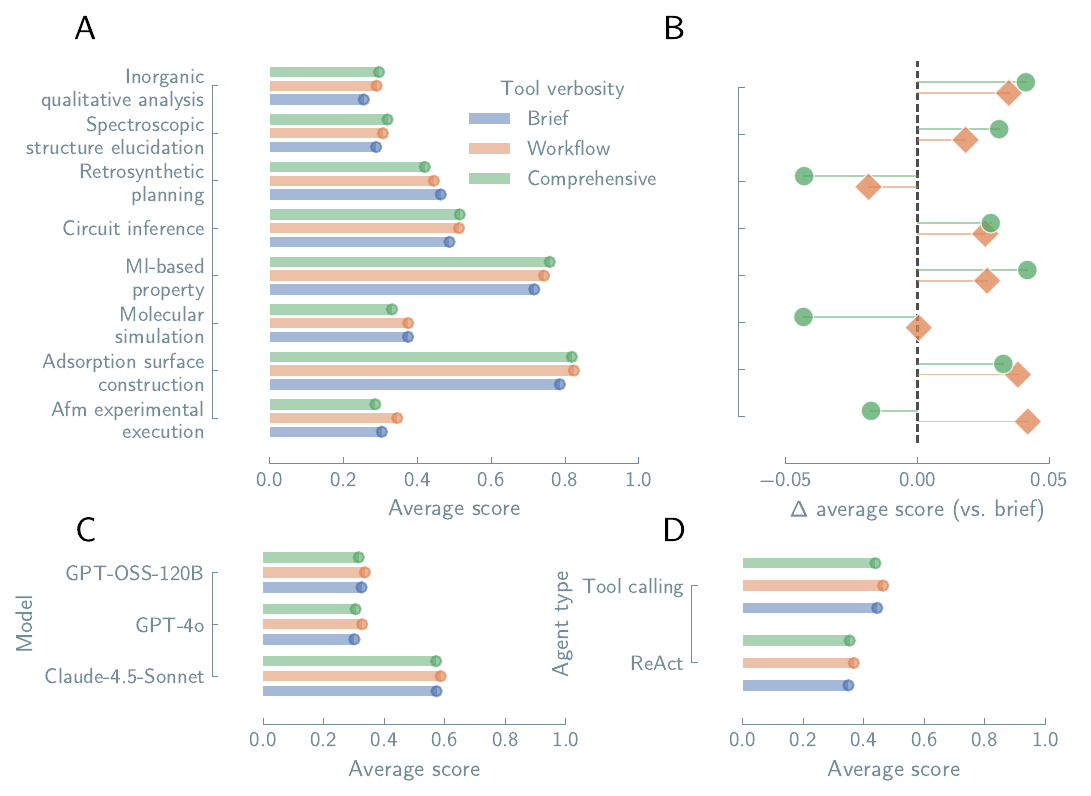}
    \caption{\textbf{Detailed results across tool-documentation verbosity levels.}
    (A) Results across environments. Scores are close across verbosity levels.
    (B) Change in workflow and comprehensive scores relative to brief. Differences are within 0.05 points in every environment.
    (C) Results per model. Each model's scores are close across the three verbosity levels.
    (D) Results per agent. Scores for the two scaffolds track each other across verbosity levels.}
    \label{fig:fig-5_app}
\end{figure}

We compare agent scores at the three tool-documentation verbosity levels defined in \Cref{sec:tool-verbosity}, holding the model, scaffold, environment, and scope fixed. Scores are aggregated per environment, per model, and per scaffold. \Cref{fig:fig-5_app} shows that verbosity has a negligible effect, consistent with the $<$0.1\% variance share attributed to verbosity by the latent factor model (\Cref{fig:irt-main-fig}C).

\section{Reasoning process analysis}
\label{app:reasoning-process}

This analysis examines how agents reach their answers, independent of whether the answer is correct. Each reasoning trace is annotated as a graph of epistemic operations (\Cref{sec:epistemological-graphs}); productive patterns and reasoning breakdowns are detected as structural templates over the graph.

\paragraph{Annotation and review.} The two-stage LLM annotation pipeline is described in \Cref{sec:epistemological-graphs} and expanded with implementation details in \Cref{app:annotation-pipeline}. Manual review by two domain-expert annotators was performed on 626 selected traces to quantify agreement with the LLM pipeline. A detailed analysis of the annotation results can be found in \Cref{app:annotation-agreement}.

\paragraph{Pattern prevalence across models.} Reasoning breakdowns dominate in both annotated models. Evidence non-uptake appears in 88\% of traces for both \claudesonnet and \gptfourO, and untested claims in 66\% and 62\% respectively (\Cref{tab:epist-full-scores}). Productive counterparts to these breakdowns are rare: convergent multi-test evidence and evidence-guided test redesign each appear in at most 1\% of traces; refutation-driven belief revision appears in 4\% and 1\%. The two models diverge most on fixed belief traces (47\% versus 67\%) and on stalled revision (21\% versus 8\%); elsewhere their profiles are close.

\paragraph{Pattern prevalence across environments.} \Cref{fig:epist_res-env} shows that reasoning breakdowns exceed productive motifs at every environment level. This is consistent with the main-text claim that undisciplined reasoning appears across task types rather than emerging only under the highest task difficulty.

\subsection{LLM annotation pipeline}
\label{app:annotation-pipeline}

The epistemological graph prevalences summarised in \Cref{tab:epist-full-scores} and \Cref{fig:epist_res-env} are produced by a two-stage annotation pipeline operating on the serialized message history of each completed agent run.

\paragraph{Input representation.} Each trace is serialized to a linear message list with role, content, and step index preserved. System prompts and iteration-limit error messages are excluded. Task descriptions remain eligible for annotation because they carry information the agent subsequently uses.

\paragraph{Stage 1: node labeling.} \claudesonnet\ is prompted to label message-level nodes with epistemic operations drawn from the fixed vocabulary \{hypothesis, test, evidence, judgment, commitment, final answer\}. Annotation proceeds in overlapping windows of 20 messages with a stride of 15, so each message appears in several windows. For each eligible node, the model returns a label, a supporting quote taken verbatim from the message, and a short gloss. Decoding uses temperature 0.7. After all windows are processed, nodes referring to the same message are merged and duplicates removed. Tool responses are assigned as evidence. 

\paragraph{Stage 2: edge construction.} The labeled graph and the original messages are passed to a second \claudesonnet\ call, using the same windowing as in stage 1. The call adds directed edges between related nodes, typed from the fixed vocabulary \{testing, observing, informing, contradicting, competing, updating\}. Each edge carries a supporting quote and the source-message index. Decoding uses temperature 0.7.

In the pilot prompt-development experiments (see \Cref{app:epist-prompts}), \claudesonnet provided the best balance between annotation quality and cost, and showed the strongest overall performance in schema adherence, quote grounding, and consistency across long traces, as judged by the experts reviewing the annotations.

\paragraph{Validation} Every graph undergoes three automated checks: (i) every supporting quote is checked for verbatim substring membership in the cited message. Under the default mode, failures are logged as warnings, and the item is retained; (ii) each edge must connect nodes that exist in the graph, and (iii) only permitted edge‑node combinations are allowed (see Appendix \ref{app:epist-prompts}). Graphs failing any of these checks are discarded, and a warning is raised. 

\Cref{tab:epist_val} shows that most validation warnings arise from non-verbatim quote mismatches rather than structural problems. Edge-stage quote mismatches account for 56.0\% of all warnings, and node-stage mismatches for 25.8\%, together over 80\% of flagged cases. Disallowed edge-node combinations are less frequent (11.1\%). Corrections on observation messages (extra nodes and type changes) and other structural fixes are rare. No schema violations were observed for any model; the constrained-generation setup reliably enforced the node and edge vocabularies.

\begin{table}[H]
    \centering
    \caption{\textbf{Quality‑control warning counts by failure mode and model.} Each annotated reasoning graph was automatically validated after generation. Non‑verbatim quote warnings occur when the cited evidence excerpt for a node (node stage) or an edge (edge stage) is not an exact substring of the source trace; such annotations are retained but flagged. \textit{Disallowed combination} denotes an edge whose relation–type triple falls outside the permitted schema vocabulary. \textit{Extra node at observation message} counts evidence/claim nodes that were injected into observation‑only messages and subsequently removed. \textit{Node type corrected to enforce E‑only rule} counts nodes whose type was automatically changed to enforce the constraint that observation messages may contain only Evidence nodes. \textit{Other} captures miscellaneous structural corrections (e.g., node additions or residual type fixes). Schema violations (invalid node types or edge relations outside the vocabulary) were never observed. Percentages are computed over the total number of warnings across all models.}
    \label{tab:epist_val}
    \begin{tabularx}{\textwidth}{Xccccc}
  \toprule
  Failure mode & \claudesonnet & \gptfour & \gptoss & Total & \% \\
  \midrule
  Non-verbatim quote (node stage) & 4,166 & 1,995 & 388 & 6,549 & 25.8 \\
  Non-verbatim quote (edge stage) & 9,234 & 3,999 & 967 & 14,200 & 56.0 \\
  Disallowed (relation, src, dst) combination & 1,402 & 1,103 & 317 & 2,822 & 11.1 \\
  Extra node at observation message (removed) & 769 & 403 & 65 & 1,237 & 4.9 \\
  Node type corrected to enforce E-only rule & 190 & 102 & 0 & 292 & 1.2 \\
  Other (node additions / type corrections) & 117 & 111 & 41 & 269 & 1.1 \\
  \midrule
  Schema violations (none observed) & 0 & 0 & 0 & 0 & 0.0 \\
  \midrule
  \textbf{Total} & \textbf{15,878} & \textbf{7,713} & \textbf{1,778} & \textbf{25,369} & 100.0 \\
  \bottomrule
\end{tabularx}

\end{table}

\subsection{Annotations prompts} \label{app:epist-prompts}

The annotation pipeline comprises two sequential passes, each consisting of a system prompt followed by a user instruction prompt. The system prompts define the model’s behavioral constraints and high-level rules. Both prompts conclude with an explicit directive to produce JSON that conforms to the required schema, without any surrounding prose or explanation.

\begin{hypobox}{System prompt for Stage 1: Node annotation}
You are a careful annotator. You MUST only extract information explicitly present in the provided messages.\\
Rules:\\
- Do NOT invent hidden thoughts or implied steps.\\
- Do not borrow any external knowledge or make assumptions beyond the text.\\
- Do not judge or correct the content, only label supported nodes.\\
- For messages with several nodes, you MUST follow the order in which they appear in the text to assign message indices.\\
- If uncertain, omit the node rather than guessing.\\
- Output JSON only, matching the required schema.
\end{hypobox}

\begin{hypobox}{User prompts for Stage 1 annotation}
Extract 0..k nodes from the provided message window.\\

Every non-Observation message must be assigned with at least one node. It is possible that some messages will have multiple nodes (e.g., a message that both states a hypothesis and describes a test). Avoid repeated nodes of the **same type** within a single message, unless the text explicitly supports multiple distinct instances.\\

Node types:\\
H = Hypothesis: a candidate explanation, or a working assumption about the system. It should be a revisable claim, proposal, suggestion, or the current best guess about the answer. Information in task definitions or environment descriptions do not count as hypotheses (H). Partial or full reiteration of the task descriptionn does not count as hypothesis (H). Correcting a typo in a tool argument does not count as a Hypothesis (H). We consider a Hypothesis (H) to be present if the agent states a claim similar to "I think the answer is X", "This suggests that X might be the solution", "The most likely explanation is X", "X could be the case if...", "It seems that the answer can be X", etc.\\
E = Evidence: all Observation messages must be assigned with and only with an Evidence (E) node. Only Observation messages can be Evidence (E) nodes, with the exception of the task description message.\\
N = Neutral: for boilerplate operations like writing or copying files, or non-scientific tool calls. Only the tool calls of a message can be assigned as Neutral (N).\\
T = Test: any information-seeking action, including experiments, evaluations, or lookups. Both the intention and the concrete tool call qualify as tests (T); what matters is that the system is seeking new scientific information to evaluate a hypothesis (H) or a judgment (J). If both the intention/plan to run a test and the actual tool call are present within the same message, count only the tool call as the test (T) node. Every tool call (with the tag <action>) must be assigned as either a Test (T) or a Neutral (N) node.\\
J = Judgment: an interpretation of test results (Observation) that goes beyond the literal repetition of the raw output. If the agent restates an observation while adding any evaluative, comparative, or inferential content, even brief it is a judgment (J).\\
F = Final Answer: Only the message that contains a final submisssion (with tag <final\_answer>), must be a assigned with a Final Answer (F) node.\\

Constraints for nodes:\\
- Only label what is explicitly present in text.\\
- Every node must have support quotes (exact substrings) and msg indices.\\
- If you normalize text, still cite original quote(s).\\

Pseudo-nodes (not explicitly stated but can be inferred):\\
C = Commitment: if from the actions of the agents or system it can be inferred that they have reached an implicit commitment to an answer that is not yet fully supported by evidence, and that they are **refusing to revise it**, then create a pseudo-node labeled C. This is a special pseudo-node that captures the commitment even if it is not explicitly stated. It is possible that the Final Answer (F) node is also a Commitment (C) if the Final Answer (F) is not fully supported by evidence.\\

Constrains for pseudo-nodes:\\
- Only Commitment (C) can be a pseudo-node.\\
- For the quote support of a pseudo-node, you can cite the text that implies the commitment, even if it is not explicit. You are allowed to add a brief explanation to clarify the implication, but it must be concise and directly tied to the quote.\\

Return JSON with keys:\\
{\\
  "nodes": [\\
    {\\
      "node\_id": "N1",\\
      "type": "H|T|E|J|C|F",\\
      "time": <int message index of earliest support>,\\
      "text": <normalized short node text>,\\
      "support": [{"msg\_idx": <int>, "quote": <exact substring from that message>}],\\
    }, ...\\
  ]\\
}
\end{hypobox}

In the second pass (Stage 2), the model receives the previously extracted nodes alongside the same message window. Its task is to identify directed edges between those nodes. The user prompt enumerates all permissible relation types and, for each type, specifies the valid source–destination pair combinations. This ensures that only structurally grounded edges are extracted. Edges that are not permitted are deleted programmatically after annotation.

\begin{hypobox}{System prompt for Stage 2: Edge annotation}
You are a careful annotator. You MUST only add edges supported by explicit text.\\
Rules:\\
- You may only connect nodes provided to you.\\
- Do not borrow any external knowledge or make assumptions beyond the text.\\
- Do not judge or correct the content, only label supported edges.\\
- Every edge MUST include at least one support quote with message indices.\\
- If uncertain, omit the edge rather than guessing.\\
- Output JSON only, matching the required schema.
\end{hypobox}

\begin{hypobox}{User prompts for Stage 2 annotation}
Given the message window and a list of extracted nodes (with node\_id and text), extract supported edges among these nodes.\\

Only the following edge types are allowed, any other combination is forbidden:\\
- tests: Allowed only between: H -> T, J -> T.\\
    H -> T: the Test (T) directly addresses the Hypothesis' claim (H), or attempts to falsify or verify it.\\
    J -> T: the Test (T) is designed in response to the Judgment (J), or the Judgment (J) motivates the test design.\\
- observes: Allowed only between: T -> E.\\
    T -> E: the Evidence (E) is a direct result of the Test (T).\\
- updates\_to: Allowed only between: H -> H.\\
    H -> H: the later Hypothesis (H) is a revision of the earlier one, based on the nodes in between.\\
- competes\_with: Allowed only between: H -> H.\\
    H -> H: the two Hypotheses (H) are alternative explanations that are directly compared or evaluated against each other. Both hypotheses should be plausible and co-existing at the same time. If another hypothesis is introduced later as a revision of the first one, then it should be connected with updates\_to instead of competes\_with.\\
- contradicts: Allowed only between: E -> H, J -> H.\\
    E -> H: the Evidence (E) contradicts the claim of the Hypothesis (H).\\
    J -> H: the Judgment (J) contradicts the claim of the Hypothesis (H).\\
- informs: Allowed only between: E -> H, E -> J, E -> C, J -> C, J -> H, J -> J.\\
    E -> H: the Evidence (E) provides information relevant to the claim of the Hypothesis (H).\\
    E -> J: the Judgment (J) is an interpretation of the Evidence (E).\\
    E -> C: the Commitment (C) is informed by the Evidence (E) but not necessarily in an explicit way.\\
    J -> H: the Judgment (J) provides information relevant to the claim of the Hypothesis (H).\\
    J -> J: the later Judgment (J) is a refinement, an extension, or a combination of one or serveral earlier judgments (J).\\
    J -> C: the Commitment (C) is informed by the Judgment (J) but not necessarily in an explicit way.\\

Return JSON with keys:\\
{\\
  "edges": [\\
    {\\
      "src": "<node\_id>",\\
      "dst": "<node\_id>",\\
      "relation": "<one of allowed>",\\
      "time": <int message index of earliest support>,\\
      "support": [{"msg\_idx": <int>, "quote": <exact substring from that message>}]\\
    }, ...\\
  ]\\
}\\
\end{hypobox}

\paragraph{Prompt optimization}

Two experts optimized the annotation prompts for both stages by defining node and edge definitions in an iterative process on a 50-trace training set. The experts manually reviewed the LLM’s annotations and suggested improvements. When both agreed that the annotations reliably captured the underlying reasoning patterns, they proceeded with the full annotation.

\paragraph{Effect of prompt optimization.}

\begin{figure}[H]
    \centering
    \includegraphics[width=1\linewidth]{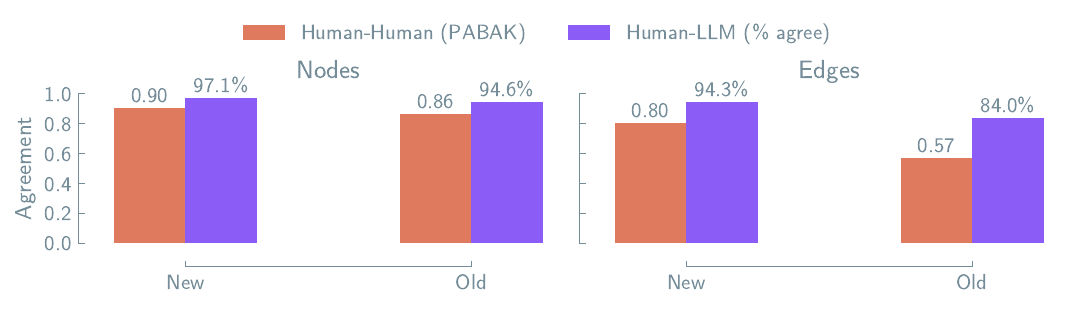}
    \caption{\textbf{Effect of prompt optimization on annotation agreement.} Human‑human (PABAK) and human‑LLM (percent agreement) scores for node and edge annotations, comparing the initial round (old) and the optimized round (new).}
    \label{fig:prompt-optimization}
\end{figure}

Prompt optimization led to substantial improvements in annotation agreement (\Cref{fig:prompt-optimization}). Human‑human agreement measured by PABAK increased from 85.8\% to 92.6\%, while human‑LLM percent agreement rose from 89.3\% to 95.7\%. The gains were similar for node and edge annotations.

\subsection{Pattern taxonomies}
\label{app:pattern-taxonomies}

Productive motifs and reasoning breakdowns are defined as explicit structural templates over the annotated graphs. Each template specifies the required node types, edge types, and their configuration; traces are scored by detecting instances of these templates. Templates are organized into three families. \textit{Hypothesis handling} captures how candidate explanations are generated, weighed against evidence, and revised. \textit{Evidence handling} captures how the results of tests are incorporated into belief. \textit{Inquiry control} captures how the overall test strategy is planned and adjusted.

\paragraph{Productive motifs.} Productive motifs are drawn from recognized practices of disciplined inquiry: Popperian falsification \autocite{popper2005logic}, exploratory-to-confirmatory transitions \autocite{tukey1977exploratory}, triangulation by convergent multi-test evidence, and active-learning-style test redesign \autocite{lindley1956measure, fedorov2013theory}. \Cref{tab:prod-motifs-def} lists every template with its short description and graph form.

\begin{table}[H]
    \centering
    \caption{\textbf{Descriptions and graph showing the productive motifs in the epistemological analysis.} Note that \textit{data-first hypothesis} and \textit{iterative test refinement} include two graphs each that refer to the same productive reasoning behavior. }
    \label{tab:prod-motifs-def}
    \definecolor{nodefill}{HTML}{DFE3E8}%
\tikzset{%
  rnode/.style={circle, draw, thick, minimum size=5mm, inner sep=1pt,
                font=\scriptsize\bfseries, fill=nodefill},%
  rlbl/.style={font=\tiny, midway, above},%
  rlblb/.style={font=\tiny, midway, below},%
  missing/.style={dashed, gray},%
  missingnode/.style={rnode, dashed, gray, text=gray, fill=none},%
}%

\begin{tabularx}{\textwidth}{p{2.2cm}cX}
\toprule
Group & Graph & Description \\
\midrule
\multirow{4}{=}{Hypothesis handling} & \begin{tikzpicture}[>=Stealth, baseline=(current bounding box.center), node distance=7mm]
  \node[rnode] (e) {E};
  \node[rnode, right=of e] (j) {J};
  \node[right=4mm of j, draw=none, font=\scriptsize] (dots) {\ldots};
  \node[rnode, right=4mm of dots] (h) {H};
  \node[rnode, right=of h] (t) {T};
  \draw[->] (e) -- node[rlbl] {inf.} (j);
  \draw[->] (h) -- node[rlbl] {tests} (t);
\end{tikzpicture} & \textbf{Evidence led hypothesis generation}. Evidence is observed first; a hypothesis is formed afterward \\
 & \begin{tikzpicture}[>=Stealth, baseline=(current bounding box.center), node distance=7mm]
  \node[rnode] (h1) {H$_1$};
  \node[rnode, right=15mm of h1] (h2) {H$_2$};
  \node[rnode, below left=5mm and 0mm of h1] (t1) {T};
  \node[rnode, below right=5mm and 0mm of h2] (t2) {T};
  \draw[<->, dashed] (h1) -- node[rlbl] {competes} (h2);
  \draw[->] (h1) -- node[left, font=\tiny] {tests} (t1);
  \draw[->] (h2) -- node[right, font=\tiny] {tests} (t2);
\end{tikzpicture} & \textbf{Hypothesis reranking}. Competing hypotheses are compared as new evidence arrives \\
 & \begin{tikzpicture}[>=Stealth, baseline=(current bounding box.center), node distance=7mm]
  \node[rnode] (h1) {H};
  \node[rnode, right=of h1] (t) {T};
  \node[rnode, right=of t] (e) {E};
  \node[rnode, right=of e] (j) {J};
  \node[rnode, right=of j] (h2) {H$_2$};
  \draw[->] (h1) -- node[rlbl] {tests} (t);
  \draw[->] (t) -- node[rlbl] {obs.} (e);
  \draw[->] (e) -- node[rlbl] {inf.} (j);
  \draw[->] (h1) to[bend right=40] node[rlblb] {upd.} (h2);
\end{tikzpicture} & \textbf{Refutation driven belief revision}. Evidence triggers a belief update to a new hypothesis \\
 & \begin{tikzpicture}[>=Stealth, baseline=(current bounding box.center), node distance=7mm]
  \node[rnode] (t1) {T};
  \node[rnode, right=of t1] (e) {E};
  \node[right=4mm of e, draw=none, font=\scriptsize] (dots) {\ldots};
  \node[rnode, right=4mm of dots] (h) {H};
  \node[rnode, right=of h] (t2) {T};
  \draw[->] (t1) -- node[rlbl] {obs.} (e);
  \draw[->] (h) -- node[rlbl] {tests} (t2);
\end{tikzpicture} & \textbf{Explore then test transition}. Exploration precedes hypothesis formation, which then drives testing \\
\midrule[0.015em]
\multirow{1}{=}{Evidence handling} & \begin{tikzpicture}[>=Stealth, baseline=(current bounding box.center), node distance=7mm]
  \node[rnode] (h) {H};
  \node[rnode, right=10mm of h, yshift=7mm] (t1) {T$_1$};
  \node[rnode, right=10mm of h] (t2) {T$_2$};
  \node[rnode, right=10mm of h, yshift=-7mm] (t3) {T$_3$};
  \node[rnode, right=10mm of t2] (e) {E};
  \draw[->] (h) -- (t1);
  \draw[->] (h) -- (t2);
  \draw[->] (h) -- (t3);
  \draw[->] (t1) -- (e);
  \draw[->] (t2) -- (e);
  \draw[->] (t3) -- (e);
\end{tikzpicture} & \textbf{Convergent multi test evidence}. One hypothesis is evaluated via multiple independent tests \\
\midrule[0.015em]
\multirow{2}{=}{Inquiry control} & \begin{tikzpicture}[>=Stealth, baseline=(current bounding box.center), node distance=7mm]
  \node[rnode] (h) {H};
  \node[rnode, right=of h] (t) {T};
  \node[rnode, right=of t] (e) {E};
  \node[rnode, right=of e] (j) {J};
  \draw[->] (h) -- node[rlbl] {tests} (t);
  \draw[->] (t) -- node[rlbl] {obs.} (e);
  \draw[->] (e) -- node[rlbl] {inf.} (j);
  \draw[->, bend left=50] (j) to node[rlblb] {tests} (t);
\end{tikzpicture} & \textbf{Fixed hypothesis test tuning}. Hypothesis is held fixed while tests are iteratively adjusted \\
 & \begin{tikzpicture}[>=Stealth, baseline=(current bounding box.center), node distance=7mm]
  \node[rnode] (j) {J};
  \node[rnode, right=of j] (t) {T};
  \node[rnode, right=of t] (e) {E};
  \draw[->] (j) -- node[rlbl] {tests} (t);
  \draw[->] (t) -- node[rlbl] {obs.} (e);
\end{tikzpicture} & \textbf{Evidence guided test redesign}. A judgment motivates a new test, which then produces new evidence \\
\bottomrule
\end{tabularx}
\end{table}

\clearpage

\paragraph{Reasoning breakdowns.} Reasoning breakdowns are drawn from established work on scientific-reasoning failure and from patterns observed during trace review. Hypothesis-handling breakdowns include untested claims, contradictions left without repair, and premature commitment. Evidence-handling breakdowns include evidence non-uptake, disconnected evidence, and unsupported judgment. Inquiry-control breakdowns include fixed belief traces, stalled revision, and fixed hypothesis test tuning (ad-hoc rescue in the Popperian sense). Both capture configurations in which the agent does not revise its plan in response to new information. \Cref{tab:reas-break-def} lists every template with its short description and graph form.

\begin{table}[H]
    \centering
    \caption{\textbf{Definitions and graphs illustrating the different reasoning breakdowns.} Each pattern is detailed as a graph and a short description. Note that \textit{untested hypothesis}, \textit{unused evidence}, and \textit{absent/stalled revision} include two different graphs per pattern, which are different cases of the same breakdown.}
    \label{tab:reas-break-def}
    \definecolor{nodefill}{HTML}{DFE3E8}%
\tikzset{%
  rnode/.style={circle, draw, thick, minimum size=5mm, inner sep=1pt,
                font=\scriptsize\bfseries, fill=nodefill},%
  rlbl/.style={font=\tiny, midway, above},%
  rlblb/.style={font=\tiny, midway, below},%
  missing/.style={dashed, gray},%
  missingnode/.style={rnode, dashed, gray, text=gray, fill=none},%
}%

\begin{tabularx}{\textwidth}{p{1.6cm}cX}
\toprule
Group & Graph & Description \\
\midrule
\multirow{4}{=}{Hypothesis handling} & \begin{tikzpicture}[>=Stealth, baseline=(current bounding box.center), node distance=7mm]
  \node[rnode] (h) {H};
  \node[missingnode, right=of h] (t) {T};
  \draw[->, missing] (h) -- node[rlbl, text=gray] {tests} (t);
  \draw[red, thick] (t.north west) -- (t.south east);
  \draw[red, thick] (t.north east) -- (t.south west);
\end{tikzpicture} & \textbf{Untested claim}. Hypothesis never linked to a test \\
 & \begin{tikzpicture}[>=Stealth, baseline=(current bounding box.center), node distance=7mm]
  \node[rnode] (h) {H};
  \node[rnode, below left=5mm and 1mm of h] (es) {E};
  \node[rnode, right=12mm of h] (j) {J};
  \node[rnode, right=of j] (c) {C};
  \node[missingnode, below right=5mm and 1mm of h] (ec) {E$_{\!c}$};
  \draw[->] (es) -- node[left, font=\tiny] {inf.} (h);
  \draw[->] (j) -- node[rlbl] {inf.} (h);
  \draw[->] (j) -- node[rlbl] {inf.} (c);
  \draw[->, missing] (ec) -- node[right, font=\tiny, text=gray] {contr.} (h);
  \draw[red, thick] (ec.north west) -- (ec.south east);
  \draw[red, thick] (ec.north east) -- (ec.south west);
\end{tikzpicture} & \textbf{One sided confirmation}. Commitment (explicit or inferred) without contradicting evidence \\
 & \begin{tikzpicture}[>=Stealth, baseline=(current bounding box.center), node distance=7mm]
  \node[rnode] (e) {E};
  \node[rnode, right=of e] (h) {H};
  \node[missingnode, right=of h] (h2) {H$_2$};
  \draw[->] (e) -- node[rlbl] {contr.} (h);
  \draw[->, missing] (h) -- node[rlbl, text=gray] {upd.} (h2);
  \draw[red, thick] (h2.north west) -- (h2.south east);
  \draw[red, thick] (h2.north east) -- (h2.south west);
\end{tikzpicture} & \textbf{Contradiction without repair}. Contradiction unresolved by any update or alternative \\
 & \begin{tikzpicture}[>=Stealth, baseline=(current bounding box.center), node distance=7mm]
  \node[rnode] (j) {J};
  \node[rnode, above right=5mm and 8mm of j] (h) {H};
  \node[rnode, below right=5mm and 8mm of j] (c) {C};
  \node[missingnode, right=of h] (t) {T};
  \draw[->] (j) -- node[rlbl] {inf.} (h);
  \draw[->] (j) -- node[rlblb] {inf.} (c);
  \draw[->, missing] (h) -- node[rlbl, text=gray] {tests} (t);
  \draw[red, thick] (t.north west) -- (t.south east);
  \draw[red, thick] (t.north east) -- (t.south west);
\end{tikzpicture} & \textbf{Premature commitment}. Commitment (explicit or inferred) to a hypothesis without testing it first \\
\midrule[0.03em]
\multirow{4}{=}{Evidence handling} & \begin{tikzpicture}[>=Stealth, baseline=(current bounding box.center), node distance=7mm]
  \node[rnode] (e) {E};
  \node[missingnode, right=of e] (j) {J};
  \draw[->, missing] (e) -- node[rlbl, text=gray] {inf.} (j);
  \draw[red, thick] (j.north west) -- (j.south east);
  \draw[red, thick] (j.north east) -- (j.south west);
\end{tikzpicture} & \textbf{Evidence non uptake}. Evidence collected but never used \\
 & \begin{tikzpicture}[>=Stealth, baseline=(current bounding box.center), node distance=7mm]
  \node[rnode] (e) {E};
  \node[missingnode, left=of e] (t) {T};
  \node[missingnode, right=of e] (j) {J};
  \draw[->, missing] (t) -- (e);
  \draw[->, missing] (e) -- (j);
  \draw[red, thick] (t.north west) -- (t.south east);
  \draw[red, thick] (t.north east) -- (t.south west);
  \draw[red, thick] (j.north west) -- (j.south east);
  \draw[red, thick] (j.north east) -- (j.south west);
\end{tikzpicture} & \textbf{Disconnected evidence}. Evidence node with no edges \\
 & \begin{tikzpicture}[>=Stealth, baseline=(current bounding box.center), node distance=7mm]
  \node[missingnode] (e) {E};
  \node[rnode, right=of e] (j) {J};
  \draw[->, missing] (e) -- node[rlbl, text=gray] {inf.} (j);
  \draw[red, thick] (e.north west) -- (e.south east);
  \draw[red, thick] (e.north east) -- (e.south west);
\end{tikzpicture} & \textbf{Unsupported judgment}. Judgment made without supporting evidence \\
 & \begin{tikzpicture}[>=Stealth, baseline=(current bounding box.center), node distance=7mm]
  \node[rnode] (t) {T};
  \node[missingnode, right=of t] (e) {E};
  \draw[->, missing] (t) -- node[rlbl, text=gray] {obs.} (e);
  \draw[red, thick] (e.north west) -- (e.south east);
  \draw[red, thick] (e.north east) -- (e.south west);
\end{tikzpicture} & \textbf{Uninformative test}. Test produces no observed evidence \\
\midrule[0.03em]
\multirow{3}{=}{Inquiry control} & \begin{tikzpicture}[>=Stealth, baseline=(current bounding box.center), node distance=7mm]
  \node[rnode] (h) {H};
  \node[rnode, right=of h] (t) {T};
  \node[rnode, right=of t] (e) {E};
  \node[rnode, right=of e] (j) {J};
  \node[missingnode, below=5mm of h] (h2) {H$_2$};
  \draw[->] (h) -- (t);
  \draw[->] (t) -- (e);
  \draw[->] (e) -- (j);
  \draw[->, missing] (h) -- node[left, font=\tiny, text=gray] {upd.} (h2);
  \draw[red, thick] (h2.north west) -- (h2.south east);
  \draw[red, thick] (h2.north east) -- (h2.south west);
\end{tikzpicture} & \textbf{Fixed belief trace}. No hypothesis revision in the entire trace \\
 & \begin{tikzpicture}[>=Stealth, baseline=(current bounding box.center), node distance=7mm]
  \node[rnode] (c) {C};
  \node[right=4mm of c, draw=none, font=\scriptsize] (dots) {\ldots};
  \node[rnode, right=4mm of dots] (h) {H};
  \node[rnode, right=of h] (t) {T};
  \node[rnode, right=of t] (e) {E};
  \draw[->] (h) -- node[rlbl] {tests} (t);
  \draw[->] (t) -- node[rlbl] {obs.} (e);
\end{tikzpicture} & \textbf{Precommitted test plan}. Commitment (explicit or inferred) before evidence collection \\
 & \begin{tikzpicture}[>=Stealth, baseline=(current bounding box.center), node distance=7mm]
  \node[rnode] (h1) {H};
  \node[rnode, right=of h1] (h2) {H$_2$};
  \node[right=7mm of h2, draw=none, font=\scriptsize, text=gray] (none) {$\varnothing$};
  \draw[->] (h1) -- node[rlbl] {upd.} (h2);
  \draw[->, missing] (h2) -- (none);
\end{tikzpicture} & \textbf{Stalled revision}. Revised hypothesis never tested \\
\bottomrule
\end{tabularx}
\end{table}

\subsection{Detailed annotation results}

\paragraph{Detailed per-model results.} \Cref{tab:epist-full-scores} reports, for each model, the fraction of traces that exhibit each motif or breakdown, averaged across environments. The table is organized into three reasoning groups (hypothesis handling, evidence handling, and inquiry control), with productive motifs and breakdowns listed separately within each group. Breakdown prevalence consistently exceeds productive-motif prevalence across all models, confirming that the aggregate trend reported in \Cref{fig:epistemology} is not driven by any single system.

\begin{table}[H]
    \centering
    \caption{\textbf{Detailed pattern presence for the epistemological analysis}. The table presents the percentages for each model included in our work, averaged across environments.}
    \label{tab:epist-full-scores}
    \begin{tabularx}{\linewidth}{X c c c}
\toprule
Pattern & Claude-4.5-Sonnet & GPT-4o & GPT-OSS-120B \\
\midrule
\multicolumn{4}{l}{\textit{Hypothesis handling}} \\
\midrule
Evidence-led
hypothesis generation & 0.63 & 0.78 & 0.51 \\
Hypothesis reranking & 0.17 & 0.07 & 0.04 \\
Refutation-driven belief revision & 0.41 & 0.21 & 0.13 \\
Explore-then-test transition & 0.59 & 0.68 & 0.16 \\
Untested claim & 0.48 & 0.44 & 0.72 \\
One-sided confirmation & 0.06 & 0.13 & 0.09 \\
Contradiction without repair & 0.23 & 0.24 & 0.08 \\
Premature commitment & 0.05 & 0.10 & 0.11 \\
\midrule
\multicolumn{4}{l}{\textit{Evidence handling}} \\
\midrule
Convergent multi-test
evidence & 0.13 & 0.02 & 0.06 \\
Evidence non-uptake & 0.81 & 0.77 & 0.35 \\
Disconnected evidence & 0.30 & 0.20 & 0.02 \\
Unsupported judgment & 0.16 & 0.09 & 0.21 \\
Uninformative test & 0.41 & 0.34 & 0.14 \\
\midrule
\multicolumn{4}{l}{\textit{Inquiry control}} \\
\midrule
Fixed hypothesis test tuning & 0.13 & 0.17 & 0.04 \\
Evidence-guided test redesign & 0.47 & 0.62 & 0.16 \\
Fixed belief trace & 0.57 & 0.78 & 0.82 \\
Precommitted test plan & 0.02 & 0.00 & 0.01 \\
Stalled revision & 0.19 & 0.08 & 0.10 \\
\bottomrule
\end{tabularx}
\end{table}

\begin{figure}[H]
    \centering
    \includegraphics[width=1\linewidth]{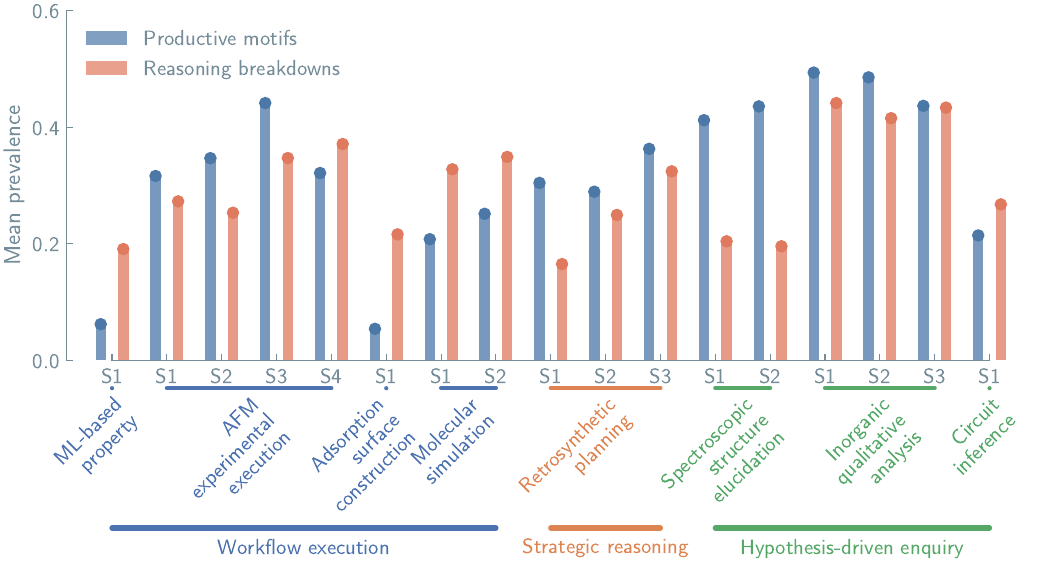}
    \caption{\textbf{Motif mean prevalence detailed for each environment.} Values are averaged across the corresponding annotated pattern sets and aggregated across models. Reasoning breakdowns are more prevalent than productive motifs across all environment scopes.
    }
    \label{fig:epist_res-env}
\end{figure}

\paragraph{Environment-level prevalence.} \Cref{fig:epist_res-env} breaks down the same prevalence estimates by environment and subtask level, aggregating across models. Each row represents one environment (e.g., \textit{Spectroscopic structure elucidation S1}) and is color-coded according to its category among the three environment types: workflow execution, strategic reasoning, and hypothesis-driven inquiry. Across every environment level, reasoning breakdowns consistently outnumber productive motifs. Hypothesis-driven environments show the highest overall annotation density, which aligns with their greater demands for iterative belief revision.

\begin{figure}[H]
    \centering
    \includegraphics[width=1\linewidth]{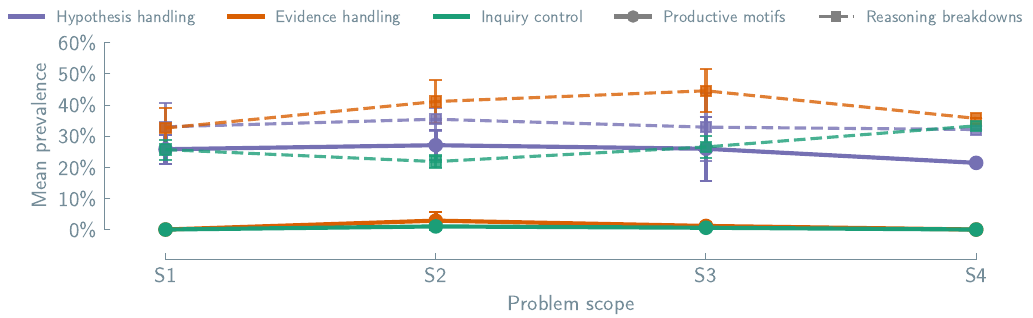}
    \caption{\textbf{Epistemic operation prevalence remains invariant to increasing problem scope.} Mean prevalence of productive motifs (solid lines) and reasoning breakdowns (dashed lines) across the three epistemic groups (hypothesis handling, evidence handling, and inquiry control), plotted against increasing problem scope (S1--S4), averaged over all multi-level environments. Error bars denote the standard error of the mean across environments. Despite a fourfold increase in task complexity, neither the absolute prevalence nor the relative ordering of epistemic operations shifts systematically, indicating that agents apply a fixed repertoire of reasoning strategies regardless of problem scope.}
    \label{fig:scope_invariance}
\end{figure}

\paragraph{Scope invariance across subtask levels.} \Cref{fig:scope_invariance} examines how epistemic operation prevalence changes as problem scope increases from S1 to S4, averaging across all multi-level environments. Despite the fourfold increase in task complexity, neither the absolute prevalence nor the relative ordering of these operations shifts systematically. This invariance suggests that agents do not adapt their epistemic behavior to broader or more complex problem settings.

\subsection{Annotation agreement analysis} \label{app:annotation-agreement}

To validate the automated LLM annotation pipeline, three domain-expert annotators independently reviewed a representative sample of 25 traces drawn from the full set of 626 traces, covering both annotation rounds (before and after prompt optimization). We report inter-annotator agreement at two levels: human--human and human--LLM. Because the label distribution is heavily skewed toward correct, which deflates Cohen's $\kappa$, we additionally report the prevalence- and bias-adjusted kappa (PABAK) as a more informative chance-corrected measure.

\paragraph{Human–human agreement.}
\Cref{tab:agreement-hh-new} summarizes pairwise agreement among the three annotators on the optimized prompt annotations. Overall percent agreement reaches 92.6\%, with a mean PABAK of 0.853, indicating substantial agreement. Node-level annotations show higher concordance (PABAK up to 0.976) than edge-level ones (PABAK 0.737--0.913); this is expected, as judging relational edges requires more nuanced interpretation of the trace. Cohen's $\kappa$ values are low (overall 0.067) due to the extreme prevalence of the \emph{correct} class. PABAK corrects for this imbalance and confirms that observed disagreements are rare.

\begin{table}[H]
\centering
\caption{\textbf{Human–human inter-annotator agreement on optimized annotations.} Pairwise Cohen's $\kappa$, PABAK, and percent agreement for nodes and edges across three domain experts. Low $\kappa$ values reflect the high prevalence of the \emph{correct} class; PABAK corrects for this imbalance.}
\label{tab:agreement-hh-new}
\begin{tabular}{l c c c c}
\toprule
Pair & Cohen's $\kappa$ & PABAK & \% Agree & $n$ \\
\midrule
\multicolumn{5}{l}{\textit{Nodes}} \\
\midrule
  R1 vs R2 & -0.014 & 0.841 & 92.1\% & 542 \\
  R1 vs R3 & 0.197 & 0.889 & 94.5\% & 416 \\
  R2 vs R3 & -0.006 & 0.976 & 98.8\% & 415 \\
\midrule
\multicolumn{5}{l}{\textit{Edges}} \\
\midrule
  R1 vs R2 & -0.005 & 0.737 & 86.8\% & 433 \\
  R1 vs R3 & 0.233 & 0.761 & 88.0\% & 343 \\
  R2 vs R3 & -0.005 & 0.913 & 95.6\% & 343 \\
\midrule
Overall & 0.067 & 0.853 & 92.6\% & 2492 \\
\bottomrule
\end{tabular}

\end{table}

\paragraph{Human–LLM agreement.}
\Cref{tab:agreement-hl-new} reports the percent agreement between each annotator and the LLM pipeline on the optimized annotations. Overall human--LLM agreement is 95.7\%, which exceeds the human--human baseline of 92.6\%. For nodes, all three reviewers agree with the LLM on at least 92.8\% of items, with two reviewers reaching 99.3\%. Edge agreement is slightly lower (87.1\%--99.8\%), consistent with the pattern observed in the human--human analysis. These results indicate that the LLM pipeline produces annotations whose quality is comparable to, and in aggregate slightly above, the level of agreement among human experts.

\begin{table}[H]
\centering
\caption{\textbf{Human–LLM agreement on optimized annotations.} Percent agreement between each of the three annotators and the LLM pipeline, reported separately for nodes and edges. Overall agreement (95.7\%) exceeds the human–human baseline, showing that the automated pipeline operates within the range of human expert judgment.}
\label{tab:agreement-hl-new}
\begin{tabular}{l c c}
\toprule
Reviewer & \% Agree & $n$ \\
\midrule
\multicolumn{3}{l}{\textit{Nodes}} \\
\midrule
  R1 & 92.8\% & 543 \\
  R2 & 99.3\% & 542 \\
  R3 & 99.3\% & 416 \\
\midrule
\multicolumn{3}{l}{\textit{Edges}} \\
\midrule
  R1 & 87.1\% & 433 \\
  R2 & 99.8\% & 433 \\
  R3 & 95.9\% & 343 \\
\midrule
Overall & 95.7\% & 2710 \\
\bottomrule
\end{tabular}

\end{table}

\subsection{Other agent behavior analysis}

\begin{figure}[H]
    \centering
    \includegraphics[width=1\linewidth]{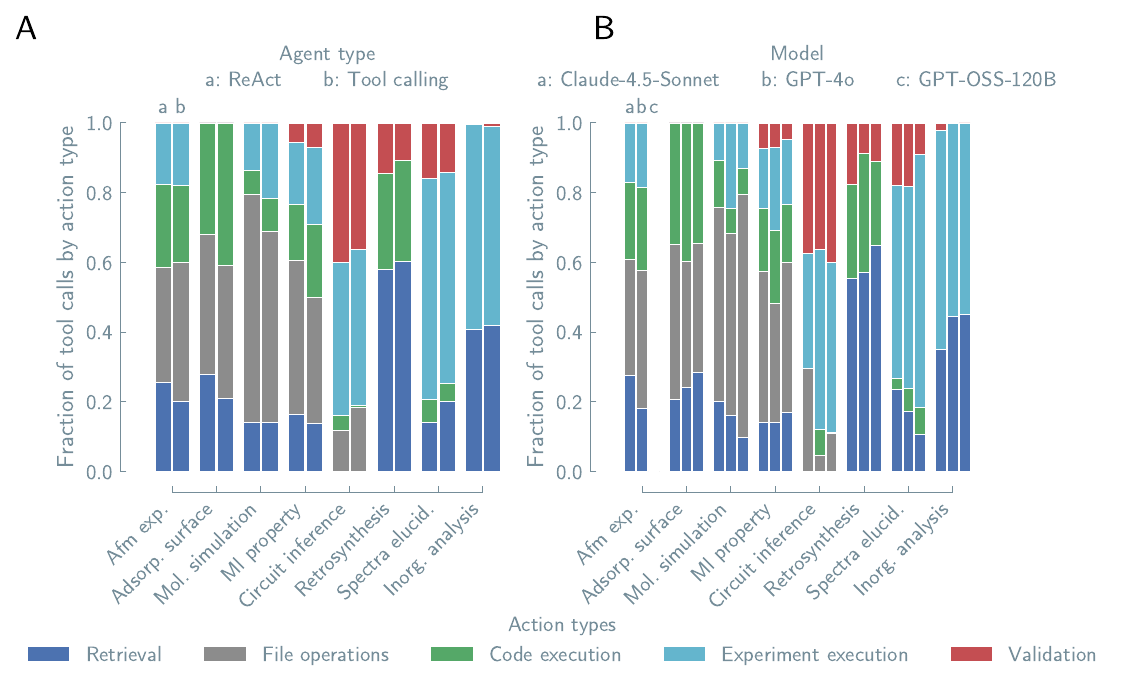}
    \caption{\textbf{Detailed results of action distributions in \corral}. Tools are classified according to the task type they perform. (A) Action distribution percentages by agent type. Differences between agents are minor, with both showing similar tool use patterns across environments. (B) Action distribution percentages by model type. Notable differences across models emerge in specific environments: only GPT-4o employs code execution tools in the Resistor environment. In AFM, both Claude and GPT-4o use these tools substantially more than GPT-Oss, whereas in MD, Claude shows the highest usage among the three models.}
    \label{fig:fig-4-app-a}
\end{figure}

\begin{figure}[H]
    \centering
    \includegraphics[width=1\linewidth]{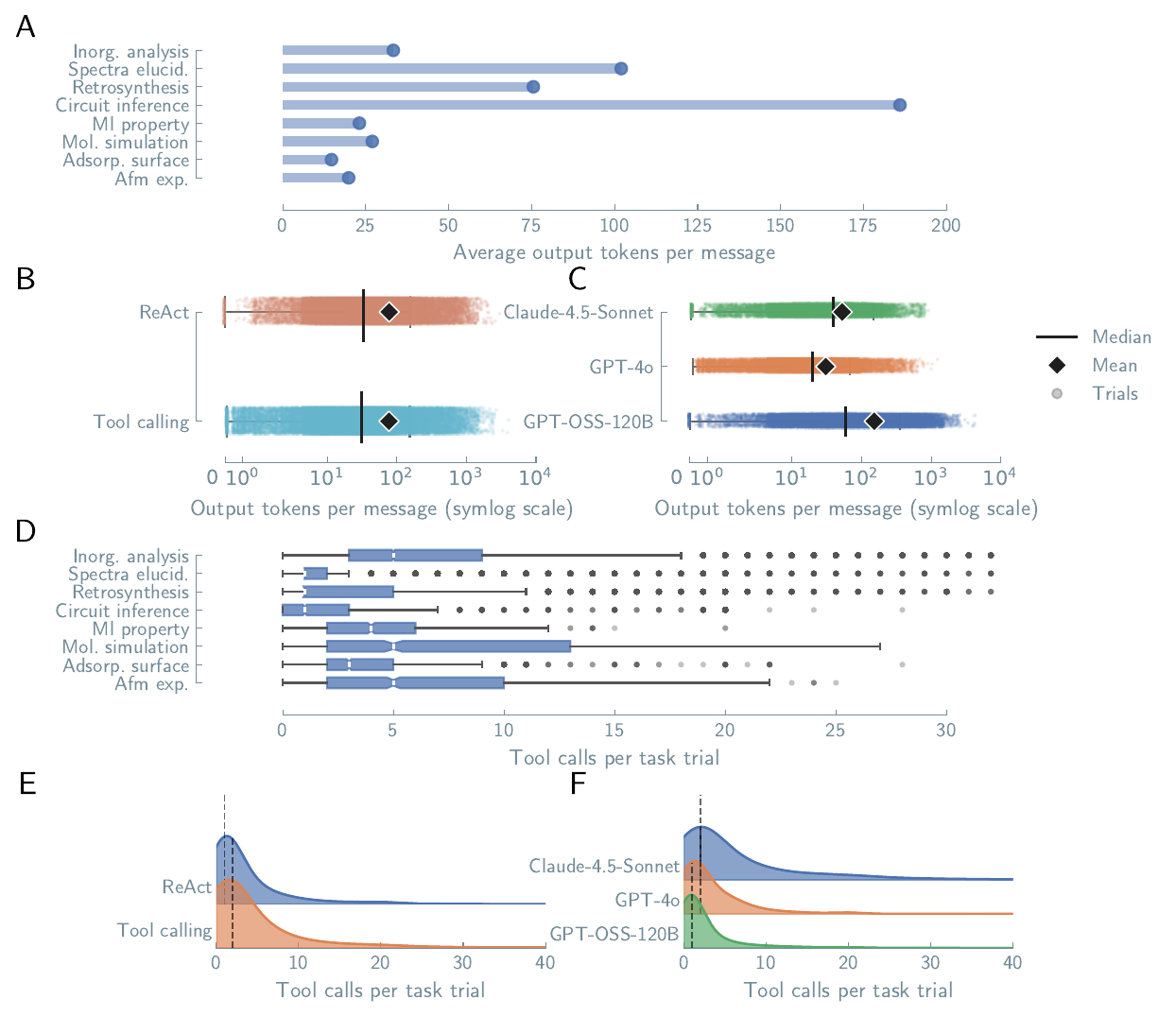}
    \caption{\textbf{Detailed results on tool calls and output tokens across environments and models.}
    (A) Average output tokens per message across environments. Resistor yields markedly more tokens per message than the other environments. Environments with lower performance tend to produce a higher number of tokens.
    (B) Tokens per message for each agent. Despite ReAct explicitly incorporating a Thought category to prompt reasoning, the token distributions across agents are similar.
    (C) Tokens per message for each model. GPT-Oss is the most verbose model, most likely due to its use of reasoning tokens.
    (D) Tool calls per trial across environments. Average number of tool calls per trial, aggregated over models and agents.
    (E) Tool distribution for each agent. Both agents exhibit similar tool‑call distributions. Although the tool‑calling agent shows a slightly longer tail, the overall differences are minimal.
    (F) Number of tool calls per model. Claude, the best‑performing model, also tends to make more tool calls than the other models.}
    \label{fig:fig-4-app-b}
\end{figure}

To complement the main performance results, we also analyze how agents interact with the environment at a behavioral level. Specifically, we examine the distribution of tool actions, the verbosity of model outputs, and the number of tool calls per trial. These analyses help clarify whether differences in task performance stem from distinct interaction strategies or from more subtle variations in how models use the same tool ecosystem. \Cref{fig:fig-4-app-a,fig:fig-4-app-b} summarize these behavioral patterns across environments, agent scaffolds, and model families.

\Cref{fig:fig-4-app-a} shows the distribution of action categories across environments, split by agent type and by model. The two agent scaffolds exhibit similar action profiles, suggesting that ReAct and tool-calling agents rely on comparable tool-use patterns. In contrast, differences across models are more pronounced in specific environments, indicating that the model family has a stronger influence than the agent scaffold on which types of tools are invoked.

\Cref{fig:fig-4-app-b} provides a complementary view based on output length and tool-use intensity. The figure shows that verbosity varies substantially across environments (some settings elicit longer responses than others) and that models also differ in how many tokens they produce and how many tools they call during a trial. The behavioral differences in this figure are stronger than those in the action-mixture analysis, suggesting that models differ less in what types of actions they take than in how extensively they reason and interact with tools while solving the task.

\clearpage

\begin{longtable}{p{\textwidth}}
\caption{\textbf{Tools for the wetlab environment.} The level of verbosity shown is \texttt{[BRIEF]}. Refer to \url{https://lamalab-org.github.io/corral} to inspect other verbosity levels, and all the tools for the other environments included in this work.}\label{tab:wetlab-tools} \\
\endfirsthead

\multicolumn{1}{c}{\tablename\ \thetable{} -- continued from previous page} \\
\endhead

\multicolumn{1}{r}{Continued on next page} \\
\endfoot

\endlastfoot

\textbf{\texttt{add\_a\_solution}} \\
\begin{description}
    \item[\textbf{Description:}] Adds a specific volume of `sol2\_label` to all of `sol1\_label` and returns observations about the changes of solution color and precipitation amount and color.

    \item[\textbf{Arguments:}]
    \begin{description}
        \item \texttt{test\_label} (str): label of the resulting solution 
        \item \texttt{sol1\_label} (str): label of the first solution
        \item \texttt{sol2\_label} (str): label of the second solution
        \item \texttt{sol2\_vol} (int): volume of the second solution to draw
    \end{description}
    \item[\textbf{Returns:}] a string containing the observations from the test
\end{description} \\
\midrule
\textbf{\texttt{add\_precipitate\_to\_solution}} \\
\begin{description}
    \item[\textbf{Description:}] Adds a precipitate to a specific volume of a solution and returns observations about the changes in the amount/color of the added precipitate or the solution color.

    \item[\textbf{Arguments:}]
    \begin{description}
        \item \texttt{test\_label} (str): label of the resulting mixture
        \item \texttt{prec\_label} (str): label of the precipitate
        \item \texttt{sol\_label} (str): label of the solution
        \item \texttt{sol\_vol} (int): volume of the solution
    \end{description}
    \item[\textbf{Returns:}] a string containing the observations from the test
\end{description} \\
\midrule
\textbf{\texttt{check\_inventory}} \\
\begin{description}
    \item[\textbf{Description:}] Returns the current contents of the Inventory
    \item[\textbf{Arguments:}]
    None
    \item[\textbf{Returns:}] a string containing the solutions and precipitates in the Inventory
\end{description} \\
\midrule
\textbf{\texttt{checkout\_color}} \\
\begin{description}
    \item[\textbf{Description:}] Observe the color of a solution or precipitate.
    \item[\textbf{Arguments:}]
    \begin{description}
        \item \texttt{label} (str): label of the target object
    \end{description}
    \item[\textbf{Returns:}] the type (reagent, solution, or precipitate) and the color of the object
\end{description} \\
\midrule
\textbf{\texttt{filter\_solution}} \\
\begin{description}
    \item[\textbf{Description:}] Separates the precipitate from the supernatant solution.
    \item[\textbf{Arguments:}]
    \begin{description}
        \item \texttt{label} (str): label of the target solution
    \end{description}
    \item[\textbf{Returns:}] a string with a message about the success/failure of the filtration
\end{description} \\
\midrule
\textbf{\texttt{get\_available\_reagents}} \\
\begin{description}
    \item[\textbf{Description:}] Returns the list of available reagent solutions.
    \item[\textbf{Arguments:}]
    None
    \item[\textbf{Returns:}] a string containing all available reagents
\end{description} \\
\midrule
\textbf{\texttt{lookup\_flame\_colors}} \\
\begin{description}
    \item[\textbf{Description:}] Returns the characteristic flame colors of cations.
    \item[\textbf{Arguments:}]
    None
    \item[\textbf{Returns:}] the list of characteristic flame colors
\end{description} \\
\midrule
\textbf{\texttt{lookup\_precipitate\_colors}} \\
\begin{description}
    \item[\textbf{Description:}] Returns the list of colored precipitates and their color.
    \item[\textbf{Arguments:}]
    None
    \item[\textbf{Returns:}] the list of precipitate colors
\end{description} \\
\midrule
\textbf{\texttt{measure\_pH}} \\
\begin{description}
    \item[\textbf{Description:}] Measures the pH of the solution using a pH paper.
    \item[\textbf{Arguments:}]
    \begin{description}
        \item \texttt{label} (str): label of the target solution
    \end{description}
    \item[\textbf{Returns:}] the closest integer value to the actual pH of the solution
\end{description} \\
\midrule
\textbf{\texttt{mix\_two\_solutions}} \\
\begin{description}
    \item[\textbf{Description:}] Mixes two solutions with the given volumes an returns observations about precipitation and color of the resulting solution
    \item[\textbf{Arguments:}]
    \begin{description}
        \item \texttt{test\_label} (str): label of the resulting solution
        \item \texttt{sol1\_label} (str): label of the first solution
        \item \texttt{sol1\_vol} (int): volume of the first solution to draw
        \item \texttt{sol2\_label} (str): label of the second solution
        \item \texttt{sol2\_vol} (int): volume of the second solution to draw
    \end{description}
    \item[\textbf{Returns:}] a string containing the observations from the test
\end{description} \\
\midrule
\textbf{\texttt{perform\_flame\_test}} \\
\begin{description}
    \item[\textbf{Description:}] Performs a flame test on the solution. Cost = 1 mL
    \item[\textbf{Arguments:}]
    \begin{description}
        \item \texttt{label} (str): label of the target solution
    \end{description}
    \item[\textbf{Returns:}] the resulting observation from the flame test
\end{description} \\
\midrule
\textbf{\texttt{possible\_anions}} \\
\begin{description}
    \item[\textbf{Description:}] Returns the list of possible anions.
    \item[\textbf{Arguments:}]
    None
    \item[\textbf{Returns:}] a string containing all the possible anions
\end{description} \\
\midrule
\textbf{\texttt{possible\_cations}} \\
\begin{description}
    \item[\textbf{Description:}] Returns the list of possible cations.
    \item[\textbf{Arguments:}]
    None
    \item[\textbf{Returns:}] a string containing all the possible cations
\end{description} \\
\midrule
\textbf{\texttt{simulate\_color\_mixture}} \\
\begin{description}
    \item[\textbf{Description:}] Given a mixture of precipitate colors, it mixes them with the given fractions and returns the name of the resulting precipitate color.
    \item[\textbf{Arguments:}]
    \begin{description}
        \item \texttt{mixture} (list[tuple[str, float]]): mixture components and their fractions as a list of tuples
    \end{description}
    \item[\textbf{Returns:}] name of the resulting color
\end{description} \\
\end{longtable}


\definecolor{colENU}{RGB}{198,219,239}   
\definecolor{colUC}{RGB}{253,205,172}    
\definecolor{colFB}{RGB}{203,213,232}    
\definecolor{colCR}{RGB}{252,187,161}    

\definecolor{nodeE}{RGB}{86,146,194}     
\definecolor{nodeH}{RGB}{196,124,58}     
\definecolor{nodeJ}{RGB}{100,140,100}    
\definecolor{nodeT}{RGB}{136,136,136}    
\definecolor{nodeC}{RGB}{170,68,68}      
\definecolor{nodeF}{RGB}{80,80,80}       

\newcommand{\nodebadge}[1]{%
  \tikz[baseline=(n.base)]{%
    \node[circle, draw, fill=node#1!25, inner sep=0pt,
          minimum size=9pt, font=\sffamily\bfseries\tiny] (n) {#1};%
  }%
}

\subsection{Illustrative traces} \label{app:illustrative-traces}

For each major breakdown category, we include one representative trace excerpt
with the supporting quote the annotator used to assign the pattern.  For every
message in the excerpt, the epistemic nodes extracted by the annotator are
listed, together with the edges that link them.  A
\nodebadge{H}~denotes a \emph{hypothesis},
\nodebadge{T}~a \emph{test},
\nodebadge{E}~an \emph{evidence} observation,
\nodebadge{J}~a \emph{judgment}, and
\nodebadge{C}~a \emph{commitment} (premature answer).
Missing edges that \emph{should} appear in a sound reasoning chain are marked
with \textcolor{red!70}{\textsf{\bfseries\small$\times$}}.
The full annotated traces, with node and edge graphs, filtering by pattern,
and alternative candidate annotations are browsable at
\url{https://lamalab-org.github.io/corral/#explainers}.

\begin{tcolorbox}[
  colback=colENU!8, colframe=colENU!60!black,
  title={\small\sffamily\bfseries Evidence non-uptake},
  fontupper=\small,
  left=4pt, right=4pt, top=4pt, bottom=4pt,
  boxrule=0.4pt, arc=2pt,
  subtitle style={colback=colENU!25},
]
\textbf{Model:} Claude Sonnet 4.5\quad
\textbf{Environment:} NMR structure elucidation (spectra)\quad
\textbf{Scope:} 1\quad
\textbf{Trial:} 20

\tcblower
\begin{description}[font=\sffamily\bfseries\small, leftmargin=12pt, style=sameline, itemsep=6pt]

\item[Msg\,11 \normalfont\textsf{(Observation)}]
\texttt{obtain\_isomers\_from\_molecular\_formula}
$\to$~20~candidate SMILES returned for C$_{18}$H$_{14}$O$_2$, including
\texttt{COC(=O)/C=C\textbackslash{}C1=CC2=\allowbreak{}CC3=CC=CC=C3C=C2C=C1} (the correct answer)
.\\[2pt]
\nodebadge{T}~N13 \emph{Obtain isomers for molecular formula C$_{18}$H$_{14}$O$_2$}
$\xrightarrow{\textsf{observes}}$
\nodebadge{E}~N14 \emph{20 isomers returned, incl.\ several with methyl ester groups}

\item[Msg\,12 \normalfont\textsf{(Agent)}]
\enquote{\emph{I got 20 isomers with the molecular formula C$_{18}$H$_{14}$O$_2$.
Looking at the list, I can see several candidates.  Let me analyze which one
matches our NMR data\,\ldots}}\\
\textcolor{gray}{\footnotesize $\rightarrow$~calls \texttt{retrieve\_dbe\_formula}}\\[2pt]
No node extracted—the agent begins a tangential calculation instead of
using the isomer list.

\item[Msg\,14 \normalfont\textsf{(Agent)}]
\enquote{\emph{Let me recalculate the DBE for C$_{18}$H$_{14}$O$_2$\,\ldots\ Looking
back at the isomers list, let me focus on those with methyl ester
groups\,\ldots}}\\
\textcolor{gray}{\footnotesize $\rightarrow$~calls \texttt{simulate\_spectra} for methyl 9-phenanthrenecarboxylate}\\[2pt]
\nodebadge{T}~N15 \emph{Simulate spectra for methyl 9-phenanthrenecarboxylate
to compare with experimental}
\end{description}

\medskip
\begin{center}
\begin{tikzpicture}[
  ncirc/.style={circle, draw, inner sep=1pt, minimum size=18pt,
                font=\small\sffamily\bfseries},
  nlbl/.style={font=\footnotesize, below=2pt},
]
\node[ncirc, fill=nodeT!25] (T) {T};
\node[nlbl] at (T.south) {N13};
\node[ncirc, fill=nodeE!25, right=55pt of T] (E) {E};
\node[nlbl] at (E.south) {N14};
\node[ncirc, draw=gray!40, text=gray!40, right=55pt of E] (JH) {J/H};
\draw[->, thick] (T) -- node[above, font=\footnotesize] {observes} (E);
\draw[->, gray!40, dashed, thick] (E) -- node[above, font=\footnotesize, text=gray] {informs} (JH);
\node[red!70, font=\bfseries\small] at ($(E)!0.5!(JH)+(0,-0.3)$) {\texttimes};
\end{tikzpicture}
\end{center}

\smallskip\noindent
\textbf{Annotator quote:}
\emph{\enquote{\texttt{COC(=O)/C=C\textbackslash{}C1=CC2=\allowbreak{}CC3=CC=CC=C3C=C2C=C1},
\texttt{COC(=O)C12C=CC(\allowbreak{}C3=CC=CC=C31)C4=CC=CC=C24}}}. The 20 candidate
structures include the correct answer, yet node~N14 has \textbf{zero outgoing
\textsf{informs} edges}: the isomer list is never used to constrain any
judgment or hypothesis.  The agent instead guesses structures one by one
(phenanthrenecarboxylate, anthracenecarboxylate,~\ldots) without consulting
the evidence it already obtained.

\end{tcolorbox}

\begin{tcolorbox}[
  colback=colUC!8, colframe=colUC!60!black,
  title={\small\sffamily\bfseries Untested claim},
  fontupper=\small,
  left=4pt, right=4pt, top=4pt, bottom=4pt,
  boxrule=0.4pt, arc=2pt,
  subtitle style={colback=colUC!25},
]
\textbf{Model:} Claude Sonnet 4.5\quad
\textbf{Environment:} NMR structure elucidation (spectra)\quad
\textbf{Scope:} 1\quad
\textbf{Trial:} 22

\tcblower
\begin{description}[font=\sffamily\bfseries\small, leftmargin=12pt, style=sameline, itemsep=6pt]

\item[Msg\,3 \normalfont\textsf{(Observation)}]
\texttt{proton\_nmr\_spectra}
$\to$~\emph{\enquote{Deltas 2.23 (s, 3H), 2.81 (t, $J$\,=\,7.1\,Hz, 2H),
2.98 (t, $J$\,=\,7.1\,Hz, 2H), 7.17 (ddd, 7H), 7.90 (ddd, 2H).}}\\[2pt]
\nodebadge{T}~N1 \emph{Get proton NMR spectra to gather structural information}
$\xrightarrow{\textsf{observes}}$
\nodebadge{E}~N2 \emph{Proton NMR: 2.23\,(s,\,3H), 2.81\,(t,\,2H),
2.98\,(t,\,2H), 7.17\,(ddd,\,7H), 7.90\,(ddd,\,2H)}

\item[Msg\,4 \normalfont\textsf{(Agent)}]
\enquote{\emph{The two triplets at 2.81 and 2.98\,ppm suggest two CH$_2$ groups
coupled to each other (likely -CH$_2$-CH$_2$-).  The aromatic signals suggest a
substituted benzene ring.  The singlet at 2.23\,ppm suggests a methyl group
attached to a carbonyl (acetyl group).}}\\
\textcolor{gray}{\footnotesize $\rightarrow$~calls
\texttt{mass\_spectrometry\_spectra}}\\[2pt]
\nodebadge{E}~N2 $\xrightarrow{\textsf{informs}}$
\nodebadge{H}~N3 \emph{Two coupled CH$_2$ groups; singlet at 2.23\,ppm
suggests acetyl group; substituted benzene ring}

\item[Msg\,5 \normalfont\textsf{(Observation)}]
\texttt{mass\_spectrometry\_spectra} $\to$ \emph{\enquote{m/z 224.12 (intensity
100)\,\ldots}}\\[2pt]
The agent moves directly to mass spectrometry: no test is designed to
evaluate the CH$_2$-CH$_2$ or acetyl-group hypothesis.

\end{description}

\medskip
\begin{center}
\begin{tikzpicture}[
  ncirc/.style={circle, draw, inner sep=1pt, minimum size=18pt,
                font=\small\sffamily\bfseries},
  nlbl/.style={font=\footnotesize, below=2pt},
]
\node[ncirc, fill=nodeE!25] (E) {E};
\node[nlbl] at (E.south) {N2};
\node[ncirc, fill=nodeH!25, right=55pt of E] (H) {H};
\node[nlbl] at (H.south) {N3};
\node[ncirc, draw=gray!40, text=gray!40, right=55pt of H] (T) {T};
\draw[->, thick] (E) -- node[above, font=\footnotesize] {informs} (H);
\draw[->, gray!40, dashed, thick] (H) -- node[above, font=\footnotesize, text=gray] {tests} (T);
\node[red!70, font=\bfseries\small] at ($(H)!0.5!(T)+(0,-0.3)$) {\texttimes};
\end{tikzpicture}
\end{center}

\smallskip\noindent
\textbf{Annotator quote:}
\emph{\enquote{The two triplets at 2.81 and 2.98\,ppm suggest two CH$_2$ groups
coupled to each other (likely -CH$_2$-CH$_2$-).  The aromatic signals
suggest a substituted benzene ring.  The singlet at 2.23\,ppm suggests a
methyl group attached to a carbonyl (acetyl group).}}
Node~N3 carries a detailed structural interpretation but has \textbf{zero
outgoing \textsf{tests} edges}.  The agent asserts the presence of an acetyl
group and coupled CH$_2$ groups as fact, then immediately requests mass
spectrometry to determine molecular weight rather than designing an
experiment (e.g., simulating $^{13}$C NMR for a candidate) to verify
the claim.

\end{tcolorbox}

\begin{tcolorbox}[
  colback=colFB!8, colframe=colFB!60!black,
  title={\small\sffamily\bfseries Fixed belief trace},
  fontupper=\small,
  left=4pt, right=4pt, top=4pt, bottom=4pt,
  boxrule=0.4pt, arc=2pt,
  subtitle style={colback=colFB!25},
]
\textbf{Model:} GPT-4o\quad
\textbf{Environment:} LAMMPS molecular dynamics (md)\quad
\textbf{Scope:} 2\quad
\textbf{Trial:} 33

\tcblower
\begin{description}[font=\sffamily\bfseries\small, leftmargin=12pt, style=sameline, itemsep=6pt]

\item[Msg\,15 \normalfont\textsf{(Observation)}]
\texttt{list\_files} $\to$ \emph{\enquote{\{\enquote{files}: [\enquote{/potentials/BKS},
\enquote{/potentials/EAM}, \enquote{/potentials/SW}, \enquote{/potentials/TERSOFF}]\}}}\\[2pt]
\nodebadge{T}~N10 \emph{List files in /potentials/ to find correct SW
potential path}
$\xrightarrow{\textsf{observes}}$
\nodebadge{E}~N11 \emph{Potentials directory contains BKS, EAM, SW, TERSOFF
subdirectories}

\item[Msg\,16 \normalfont\textsf{(Agent)}]
\enquote{\emph{The \texttt{/potentials/} directory contains a file named
\texttt{SW}, which likely corresponds to the Stillinger-Weber potential.
I will update the LAMMPS input script to use this file\,\ldots}}\\
\textcolor{gray}{\footnotesize $\rightarrow$~calls \texttt{write\_file}
to set \texttt{pair\_coeff * * /potentials/SW Si}}\\[2pt]
\nodebadge{E}~N11 $\xrightarrow{\textsf{informs}}$
\nodebadge{H}~N12 \emph{The SW potential file is located at /potentials/SW}

\item[] {\color{gray} $\vdots$ \footnotesize\itshape three error messages and re-writes later}

\item[Msg\,27 \normalfont\textsf{(Observation)}]
\texttt{run\_lammps} $\to$ \emph{\enquote{ERROR: Lost atoms: original 8
current 0}}\\[2pt]
\nodebadge{T}~N33 \emph{Run LAMMPS simulation with updated script using
/potentials/SW/Si.sw}
$\xrightarrow{\textsf{observes}}$
\nodebadge{E}~N34 \emph{LAMMPS simulation failed: Lost atoms}

\item[Msg\,28 \normalfont\textsf{(Agent)}]
\enquote{\emph{This might be due to the rapid heating rate or insufficient
equilibration.  I will adjust the heating rate and ensure proper
equilibration\,\ldots}}\\[2pt]
\nodebadge{E}~N34 \;(no \textsf{informs} to any existing H)\\
\nodebadge{H}~N35 \emph{Lost atoms error may be due to rapid heating rate
or insufficient equilibration}

\end{description}

\medskip
\begin{center}
\begin{tikzpicture}[
  ncirc/.style={circle, draw, inner sep=1pt, minimum size=18pt,
                font=\small\sffamily\bfseries},
  nlbl/.style={font=\footnotesize, below=2pt},
]
\node[ncirc, fill=nodeH!25] (H1) {H};
\node[nlbl] at (H1.south) {N12};
\node[ncirc, fill=nodeH!25, right=60pt of H1] (H2) {H};
\node[nlbl] at (H2.south) {N19};
\node[right=16pt of H2, font=\small, text=gray] {$\cdots$};
\node[ncirc, fill=nodeH!25, right=75pt of H2] (H5) {H};
\node[nlbl] at (H5.south) {N46};
\node[ncirc, fill=nodeH!25, right=60pt of H5] (H6) {H};
\node[nlbl] at (H6.south) {N57};
\draw[->, gray!40, dashed, thick] (H1) -- node[above, font=\footnotesize, text=gray] {updates\_to} (H2);
\node[red!70, font=\bfseries\small] at ($(H1)!0.5!(H2)+(0,-0.3)$) {\texttimes};
\draw[->, gray!40, dashed, thick] (H5) -- node[above, font=\footnotesize, text=gray] {updates\_to} (H6);
\node[red!70, font=\bfseries\small] at ($(H5)!0.5!(H6)+(0,-0.3)$) {\texttimes};
\end{tikzpicture}
\end{center}

\smallskip\noindent
\textbf{Annotator quotes:}\\
\noindent N12 ($\mathsf{msg\,16}$): \emph{\enquote{The \texttt{/potentials/}
directory contains a file named \texttt{SW}, which likely corresponds to the
Stillinger-Weber potential.}}\\
\noindent N35 ($\mathsf{msg\,28}$): \emph{\enquote{This might be due to the rapid
heating rate or insufficient equilibration.}}

\smallskip\noindent
The trace contains \textbf{6 hypothesis nodes and zero \textsf{updates\_to}
edges}.  The first hypothesis~(N12) mistakes the directory
\texttt{/potentials/SW} for a file, producing an error that is never traced
back to the belief.  After two more failures, the agent introduces a second
independent hypothesis~(N35), attributing the lost atoms to heating rate, without
revising the earlier structural error.  Both beliefs persist unmodified
through the remaining 15~messages.

\end{tcolorbox}

\begin{tcolorbox}[
  colback=colCR!8, colframe=colCR!60!black,
  title={\small\sffamily\bfseries Contradiction without repair},
  fontupper=\small,
  left=4pt, right=4pt, top=4pt, bottom=4pt,
  boxrule=0.4pt, arc=2pt,
  subtitle style={colback=colCR!25},
]
\textbf{Model:} Claude Sonnet 4.5\quad
\textbf{Environment:} NMR structure elucidation (spectra)\quad
\textbf{Scope:} 2\quad
\textbf{Trial:} 45

\tcblower
\begin{description}[font=\sffamily\bfseries\small, leftmargin=12pt, style=sameline, itemsep=6pt]

\item[Msg\,17 \normalfont\textsf{(Observation)}]
\texttt{simulate\_spectra} for \texttt{CC(C)OC(=O)c1ccc\-cc1N(C)C}
$\to$~$^{1}$H NMR: 1.46\,(d,\,\textbf{6H}),\;2.97\,(s,\,6H),\;
4.75\,(p,\,1H),\;aromatic protons.\\[2pt]
\nodebadge{T}~N31 \emph{Simulate spectra for
\texttt{CC(C)OC(=O)c1ccccc1N(C)C}}
$\xrightarrow{\textsf{observes}}$
\nodebadge{E}~N25 \emph{Simulated: 1.46\,ppm (d,\,\textbf{6H});
experimental: 1.43\,ppm (d,\,\textbf{3H})}

\item[Msg\,18 \normalfont\textsf{(Agent)}]
\enquote{\emph{H-NMR: 1.46\,ppm (d,\,6H) — This is for isopropyl CH$_3$ groups, but
experimental shows 1.43\,ppm (d,\,3H).  The experimental data has only 3H in
the doublet, not 6H.  \textbf{So this isn't isopropyl ester.}  The
experimental data clearly shows CH(CH$_3$), not CH(CH$_3$)$_2$.}}\\
\textcolor{gray}{\footnotesize $\rightarrow$~calls
\texttt{get\_formula\_from\_smiles}}\\[2pt]
\nodebadge{E}~N25 $\xrightarrow{\textsf{informs}}$
\nodebadge{J}~N33 \emph{Simulated isopropyl ester shows 6H doublet at
1.46\,ppm, but experimental shows only 3H doublet at 1.43\,ppm}\\[2pt]
\nodebadge{J}~N33 $\xrightarrow{\textsf{contradicts}}$
\nodebadge{H}~N30 \emph{The compound is 2-(dimethylamino)benzoic acid
isopropyl ester}

\item[Msg\,20 \normalfont\textsf{(Agent --- final answer)}]
\enquote{\emph{Simulations are approximations and can have errors\,\ldots\
The key evidence: (1)~quartet at 5.05\,ppm, (2)~singlet at 2.69\,ppm
(6H) = N(CH$_3$)$_2$, (3)~4 aromatic protons, (4)~carbonyl at
168.30\,ppm.}}
\par\texttt{<final\_answer>CC(C)OC(=O)c1ccc\-cc1N(C)C</final\_answer>}\\[2pt]
\nodebadge{J}~N36 \emph{Despite the 3H vs.\,6H discrepancy, the overall
evidence supports the isopropyl ester}\\
\nodebadge{C}~N37 \emph{Agent commits despite unresolved discrepancy}

\end{description}

\medskip
\begin{center}
\begin{tikzpicture}[
  ncirc/.style={circle, draw, inner sep=1pt, minimum size=18pt,
                font=\small\sffamily\bfseries},
  nlbl/.style={font=\footnotesize, below=2pt},
]
\node[ncirc, fill=nodeE!25] (E) {E};
\node[nlbl] at (E.south) {N25};
\node[ncirc, fill=nodeJ!25, right=55pt of E] (J) {J};
\node[nlbl] at (J.south) {N33};
\node[ncirc, fill=nodeH!25, right=55pt of J] (H) {H};
\node[nlbl] at (H.south) {N30};
\node[ncirc, draw=gray!40, text=gray!40, right=55pt of H] (Hrev) {H$'$};
\draw[->, thick] (E) -- node[above, font=\footnotesize] {informs} (J);
\draw[->, thick, red!60!black] (J) -- node[above, font=\footnotesize] {contradicts} (H);
\draw[->, gray!40, dashed, thick] (H) -- node[above, font=\footnotesize, text=gray] {updates\_to} (Hrev);
\node[red!70, font=\bfseries\small] at ($(H)!0.5!(Hrev)+(0,-0.3)$) {\texttimes};
\end{tikzpicture}
\end{center}

\smallskip\noindent
\textbf{Annotator quote:}
\emph{\enquote{H-NMR: 1.46\,ppm (d,\,6H) — This is for isopropyl CH$_3$ groups,
but experimental shows 1.43\,ppm (d,\,3H)\,\ldots\ The experimental data
has only 3H in the doublet, not 6H.  So this isn't isopropyl ester.}}
— The agent \emph{explicitly} notes that the simulated $^{1}$H~NMR
contradicts its own hypothesis~(N30).  Yet no revised hypothesis is generated:
the \textsf{contradicts} edge from~N33 to~N30 has \textbf{no accompanying
\textsf{updates\_to}~edge}.  Instead, the agent dismisses the contradiction
as a simulation artifact and submits the isopropyl ester
(\texttt{CC(C)OC(=O)c1ccccc1N(C)C}) as the final answer~(N38).
\end{tcolorbox}

\clearpage
\section{Scope and limitations}
\label{app:limitations}

The results reported here are bounded by deliberate design choices and by practical constraints on model access and compute. We state both so that comparisons to future work, whether with richer scaffolds, broader model coverage, or larger compute budgets, can be made on equal terms.

\paragraph{Scope of the evaluation.}
Each task is treated as an independent episode: agents receive no carry-over of knowledge, strategies, or refined heuristics across tasks, in contrast with the iterative accumulation of intuition that characterizes real scientific work. We evaluate two simple scaffolds, ReAct and native tool calling, with a flat, append-only conversation history and a simple truncation heuristic (earlier non-assistant messages shortened to 100 characters) when the context window is exceeded. Advanced orchestration is outside the scope of this evaluation: plan-then-execute pipelines, hierarchical planners, explicit summarisation or retrieval, and multi-agent architectures such as debate or specialist delegation. This isolates the contribution of the base model and scaffold from engineering-intensive orchestration; the scores reported here, therefore, bound what minimal scaffolds achieve rather than what is achievable in general. All agents operate under a fixed budget of 20--40 LLM calls per task, calibrated in pilot runs to be sufficient for task completion, and at a fixed sampling temperature of $0.0$.
We do not vary this budget per task nor conduct a temperature sensitivity analysis.

\paragraph{Model coverage and compute.}
Our evaluation spans three models, \claudesonnet, \gptfourO, and \gptossfull, selected to span proprietary and open-weight systems at the current capability frontier; other prominent families are omitted. The full benchmark consumed approximately 3 billion tokens across 138 configuration-environment pairs at five trials each, with an estimated API cost of \$7.5k for the two proprietary models (excluding infrastructure for the self-hosted open-weight model) for the main final runs shown in the main text (not including interventions and subtasks). \passat metrics account for stochastic variation, but scaling to additional models, trials, or environments carries a non-trivial cost.

\paragraph{Measurement caveats.}
Log-probability-based analyses of model confidence and calibration are restricted to \gptossfull, as proprietary API providers do not expose per-token log-probabilities for tool-calling or for sufficiently long completions. We also observed higher rates of malformed responses from \gptoss (41\% of traces affected, with 1.26 errors per trace), and to a lesser extent \gptfourO (0.28 errors per trial, affecting 6.2\% of traces), compared with \claudesonnet (0.5\% of traces affected); the most common failure modes were invalid JSON in tool arguments and missing XML closing tags in ReAct traces. We did not apply engineering mitigations such as constrained decoding, output grammars, or retry-with-repair loops. The reported numbers thus reflect the behavior of each system under the same minimal scaffold, not their best-case performance with model-specific hardening.

\clearpage
\printbibliography
\end{refsection}

\end{document}